%% file: main.tex
\RequirePackage{amsmath} 
\documentclass[runningheads]{llncs}

\usepackage[final,year=2024,ID=9297]{eccv}
\usepackage{times}
\usepackage{epsfig}
\usepackage{graphicx}
\usepackage{amsmath}
\usepackage{amssymb}
\usepackage{subcaption} 
\usepackage{multirow} 
\usepackage{algorithm}
\usepackage{algpseudocode}
\usepackage{multicol}
\usepackage{xcolor}

\algrenewcommand\algorithmicrequire{\textbf{Input:}}
\algrenewcommand\algorithmicensure{\textbf{Output:}}

\usepackage{eccv}

\usepackage{eccvabbrv}

\usepackage{graphicx}
\usepackage{booktabs}

\usepackage[accsupp]{axessibility}  

\usepackage{orcidlink}

\begin{document}


\title{Möbius Transform for Mitigating Perspective Distortions in
Representation Learning}


\titlerunning{MPD}

\author{
Prakash Chandra Chhipa\inst{1}\orcidlink{0000-0002-6903-7552} \and
Meenakshi Subhash Chippa\inst{1}\orcidlink{0009-0000-2770-6271} \and
Kanjar De\inst{2}\orcidlink{0000-0003-0221-8268} \and
Rajkumar Saini\inst{1}\orcidlink{0000-0001-8532-0895} \and
Marcus Liwicki\inst{1}\orcidlink{0000-0003-4029-6574} \and
Mubarak Shah\inst{3}\orcidlink{0000-0001-6172-5572}
}

\authorrunning{Chhipa et al.}

\institute{Luleå Tekniska Universitet, Sweden \\
\email{\{prakash.chandra.chhipa, rajkumar.saini, marcus.liwicki\}@ltu.se} \\
\email{meechi-2@student.ltu.se} \and
 Fraunhofer Heinrich-Hertz-Institut, Berlin, Germany \\
\email{kanjar.de@hhi.fraunhofer.de}\\ \and
University of Central Florida, USA\\
\email{shah@crcv.ucf.edu}}

\maketitle

\begin{abstract}
Perspective distortion (PD) causes unprecedented changes in shape, size, orientation, angles, and other spatial relationships of visual concepts in images. Precisely estimating camera intrinsic and extrinsic parameters is a challenging task that prevents synthesizing perspective distortion. Non-availability of dedicated training data poses a critical barrier to developing robust computer vision methods. Additionally, distortion correction methods make other computer vision tasks a multi-step approach and lack performance.
In this work, we propose mitigating perspective distortion (MPD) by employing a fine-grained parameter control on a specific family of Möbius transform to model real-world distortion without estimating camera intrinsic and extrinsic parameters and without the need for actual distorted data. Also, we present
a dedicated perspectively distorted benchmark dataset, ImageNet-PD, to benchmark the robustness of deep learning models against this new dataset. The proposed method outperforms existing benchmarks, ImageNet-E and ImageNet-X.
Additionally, it significantly improves performance on ImageNet-PD while consistently performing on standard data distribution. 
Notably, our method shows improved performance on three PD-affected real-world applications—crowd counting, fisheye image recognition, and person re-identification—and one PD-affected challenging CV task: object detection.
The source code, dataset, and models are available on the project webpage at \url{https://prakashchhipa.github.io/projects/mpd}.

\end{abstract}

\input{sec/1_intro}
\bibliographystyle{splncs04}
\bibliography{main}

\newpage
\input{sec/suppl}

\end{document}

%% file: sec/1_intro.tex
\section{Introduction}
\label{sec:intro}

Perspective distortion (PD) in real-world imagery is omnipresent and poses challenges in developing computer vision applications. PD arises from various factors, including camera positioning, depth, intrinsic parameters such as focal length and lens distortion, and extrinsic parameters such as rotation and translation. These factors collectively influence the projection of 3D scenes onto 2D planes \cite{rahman2011efficient}, altering semantic interpretation and local geometry. Precisely estimating these parameters for perspective distortion correction is challenging, presenting a critical barrier to developing robust computer vision (CV) methods.

 Previous studies in distortion correction predominantly concentrated on the correction of PD, with limited emphasis on enhancing the robustness of CV applications. Consequently, these correction methods transform CV tasks, such as image recognition and scene understanding, into a two-stage process: firstly, rectifying the distortion, and secondly, engaging in task-specific learning. GeoNet\cite{li2019blind} employs  CNN to predict distortion flow in images without prior distortion-type knowledge. PCL\cite{yu2021pcls} uses perspective crop layers in CNNs to correct perspective distortions for 3D pose estimation. A cascaded deep structure network\cite{tan2021practical} corrects wide-angle portrait distortions without calibrated camera parameters. Another study\cite{zhao2019learning} predicts per-pixel displacement for face portrait undistortion. PerspectiveNet\cite{jin2023perspective} and ParamNet\cite{jin2023perspective} predict perspective fields and derive camera calibration parameters, respectively. Methods for fisheye image distortion correction are also developed\cite{yang2023innovating,yin2018fisheyerecnet}. 
 Zolly\cite{wang2023zolly} corrects perspective distortion by adjusting focal lengths in human mesh reconstruction tasks.

Since the correction methods are fundamentally not focused on robust representation learning for CV tasks, they essentially follow an inefficient two-step process. 
One possible approach is to focus on developing task-specific methods where PD is noticeable, such as crowd counting\cite{yang2020weakly, idrees2018composition}, object re-identification\cite{ayala2019vehicle}, autonomous vehicle\cite{wang2020pillar}, and recognizing texts in scene\cite{phan2013recognizing}. 
Further, \cite{cho2021camera} meta-learning based approach to adapt 3D human pose estimation to varying degrees of camera distortion. SPEC\cite{kocabas2021spec} tackled PD by estimating the camera parameters directly from in-the-wild images to improve the accuracy of 3D human pose predictions. 
These task-specific approaches fall short in providing generalizable and robust representation learning for PD mitigation, a crucial step for advancing CV research towards distortion robustness.

 The core reason behind the lack of generalizable and robust representation learning CV methods to mitigate PD is the absence of a real-world PD-affected training dataset. Both capturing real-world PD data and synthesizing PD-affected image data using parameter estimation are far from reality. Although existing benchmarks (ImageNet-X \cite{idrissi2022imagenet}, ImageNet-E \cite{li2023imagenet}) are partially affected by PD, a dedicated benchmark is needed.  
Therefore, there is a need to artificially synthesize PD in existing data without estimating camera intrinsic and extrinsic parameters. 
Since PD is non-linear in nature, the focus should be shifted from affine to potential non-linear transformation.

 \begin{figure}[!t]
  \centering
  \includegraphics[width=.8\columnwidth]{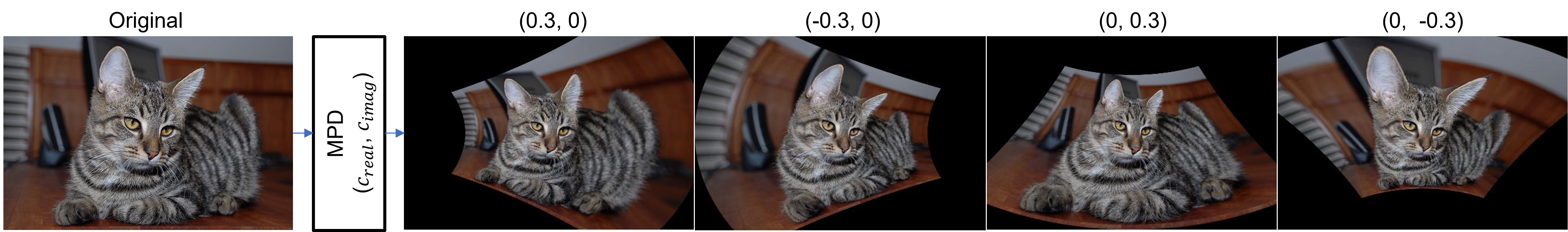}
  \vspace{-3mm}
  \caption{MPD synthesizing perspective distortion with different orientations corresponding to parameter $c: (c_{real}, c_{imag})$.}
  \label{fig:shematic}
  \vspace{-6mm}
\end{figure}

One potential non-linear family of transformation is Möbius transform \cite{arnold2008mobius,olsen2010geometry} in complex theory. Due to its conformal nature (preserving angles during transformation), \cite{zhou2021data} utilized the Möbius transform for eight specific transformations, including various twists and spreads, as a data augmentation strategy to enhance model performance on standard data distributions. In other works, Möbius convolution\cite{mitchel2022mobius}, the authors design a Möbius-equivalent convolution operator in spherical CNN for shape classification and image segmentation tasks. OmniZoomer\cite{cao2023omnizoomer} integrates Möbius transformations into a neural network, enabling movement and variable zoom on omnidirectional images while addressing the challenges of blur and aliasing associated with such transformations. MöbiusGCN\cite{azizi20223d} addresses the limitation of existing graph convolution network {GCNs} in explicitly encoding transformations between joints in 3D human pose estimation. None of these works focus on perspective distortion.

Due to the non-linear characteristics of Möbius transforming geometry in complex space while ensuring conformality, it is a potential approach to imitate perspective distortion when adapted to image data. Therefore, in the proposed mitigating perspective distortion (MPD) method,  we mathematically model PD in the images by employing fine-grained parameter control on Möbius. It is governed by four parameters $a,b,c$ and $d$ (Eq. \ref{eq_mobius}). MPD focuses on a specific family of transforms corresponding to the parameter $c$ to synthesize PD artificially. The core contribution of our method, MPD, is controlled parametrization for orientation (Fig. \ref{fig:shematic}) and varying intensity (Fig. \ref{fig:intensity}), which is capable of representing real-world perspective distortion. 

MPD transforms input image coordinates into complex vectors and performs a PD-specific family of Möbius transforms. Transformed complex vectors are then remapped to real-valued pixels, later discretized to obtain images mimicking perspective distortion while maintaining the integrity of visual concepts. Optionally, padding is applied to replace the black background with boundary pixel values to achieve another variant called integrated padding background variant (refer Fig. \ref{fig:padding} in suppl. material). Notably, MPD achieves this without estimating intrinsic and extrinsic parameters, thus bypassing the need for actual distorted data. 

Incorporating PD in computer vision is beneficial for robustness since it can be used as an augmentation, resulting in improved performance on different computer vision tasks. To address the non-availability of a suitable benchmark, we have developed a dedicated perspectively distorted benchmark dataset, {\bf ImageNet-PD},  derived from the ImageNet\cite{deng2009imagenet} validation set to evaluate the robustness of deep learning models against perspective distortion. We showcase a lack of robustness of standard deep learning models against perspective distortion by benchmarking them on ImageNet-PD.

We employ ImageNet-PD in our proposed MPD method and extensively investigate it across supervised and self-supervised learning methods (SimCLR \cite{chen2020simple}, DINO \cite{caron2021emerging}). MPD demonstrates improved performance on existing benchmarks, ImagNet-E \cite{li2023imagenet}, and ImageNet-X \cite{idrissi2022imagenet}. MPD improves $10\%$ on ImageNet-PD while consistently maintaining the performance on the original ImageNet validation set. MPD shows performance improvements when adapted to multiple real-world CV applications. Specifically, MPD advances crowd counting with a novel method, \textit{MPD-AutoCrowd}, achieving the highest performance (to the best of our knowledge) with a mean absolute error of 50.81 on ShanghaiTech-Part-A\cite{7780439} and 96.80 on UCFCC50\cite{idrees2013multi}. It demonstrates effective knowledge transfer in fish-eye image recognition, improving performance by $3\%$ on the VOC-360\cite{fu2019datasets} dataset. MPD combined with a transformer-based self-supervised method, Clip-ReIdent\cite{habel2022clip}, excels in person re-identification, achieving mean average precision of 97.02 and 98.30 with and without re-ranking, respectively, on the DeepSportradar dataset\cite{van2022deepsportradar}. Also, MPD adapted for object detection with FasterRCNN \cite{ren2015faster} and FCOS \cite{9010746} methods gain up to 3\% performance improvement on COCO \cite{lin2014microsoft} dataset. The main contributions of this paper are as follows: 
\begin{enumerate}
    \item We mathematically model perspective distortion within deep learning by developing MPD, a method that successfully mimics perspective distortion without estimating the intrinsic and extrinsic camera parameters.
    \item We develop a dedicated perspectively distorted dataset (ImageNet-PD) derived from the ImageNet\cite{deng2009imagenet} validation set to benchmark the robustness of computer vision models against PD to address the non-availability of a suitable benchmark.  
    \item We demonstrate MPD's generalizability, transferability, and adaptability across different learning paradigms, architectures, and real-world applications. 
    \item Additionally, we develop the MPD-AutoCrowd method in crowd counting to augment crowd and corresponding ground truth in scene images. 
\end{enumerate}

\section{Methodology}
\vspace{-2mm}
Möbius from complex theory \cite{arnold2008mobius, olsen2010geometry} is a family of non-linear and conformal transformations. It is defined on complex number $z$ where transformation is governed by four parameters, namely $a$, $b$, $c$, and $d$. Möbius transformation, $\Phi(z)$ is given in Eq.\ref{eq_mobius}:

\begin{equation}
\scriptsize
\Phi(z) = \frac{az + b}{cz + d}, ~~ad - bc \neq 0
\label{eq_mobius}
\end{equation}

\textbf{Non-linearity:}
Given Möbius transformations for variables \( z_1 \) and \( z_2 \), Möbius transformation of their sum \( z_1 + z_2 \):
\vspace{-2mm}
\begin{equation}
\scriptsize
    \Phi(z_1 + z_2) = \frac{a (z_1 + z_2) + b}{c (z_1 + z_2) + d} = 
    \frac{a z_1 + a z_2 + b}{c z_1 + c z_2 + d}, \rightarrow
    \Phi(z_1 + z_2) \neq \Phi(z_1) + \Phi(z_2)
    \label{eq:non-linear1}
\end{equation}
However, Eq. (\ref{eq:non-linear1}) is not equal to the sum of the individual transformations ($\Phi(z_1) + \Phi(z_2)$); therefore, \textit{non-linear}.


\textbf{Conformality:}
Conformal transformations preserve angles between curves. The angle between two curves at a point is given by the \textit{argument of the complex ratio}\cite{howie2003complex} of their derivatives. Let $f(z)$ and $g(z)$ represent curves in complex space. Their Möbius transforms are $\Phi(f(z))$, and $\Phi(g(z))$. If transformation preserves this ratio, it preserves angles. 
 Eq. \ref{eq:conform1} shows that the angles are preserved, hence conformal \cite{olsen2010geometry}. Theoretical details and theorem are in suppl. material ( refer sec. \ref{sec:sup_math} and \ref{sec:sup_theorem}).
\vspace{-2mm}
\begin{equation}
\scriptsize
\text{arg}((f(z))/ g(z))) = \text{arg}(\frac{d}{dz}(\Phi(f(z)))/ \frac{d}{dz}(\Phi(g(z)))
\label{eq:conform1}
\end{equation}
Möbius transform has been used in \cite{zhou2021data}, which defined parameter estimation for eight specific transformed appearances such as clockwise twist, clockwise half-twist, spread, spread twist, counterclockwise twist, counterclockwise half-twist, inverse, and inverse spread. These appearances are used as data augmentation in image classification on CIFAR and Tiny ImageNet datasets. However, their parameter estimation does not focus on exploring a specific family of Möbius transform to synthesize perspective distortion due to a lack of fine-grained parameter control. 

\subsection{Mitigating perspective distortion (MPD)}
\begin{figure}[h]
    \vspace{-6mm}
    \centering
    \includegraphics[width=1\columnwidth]{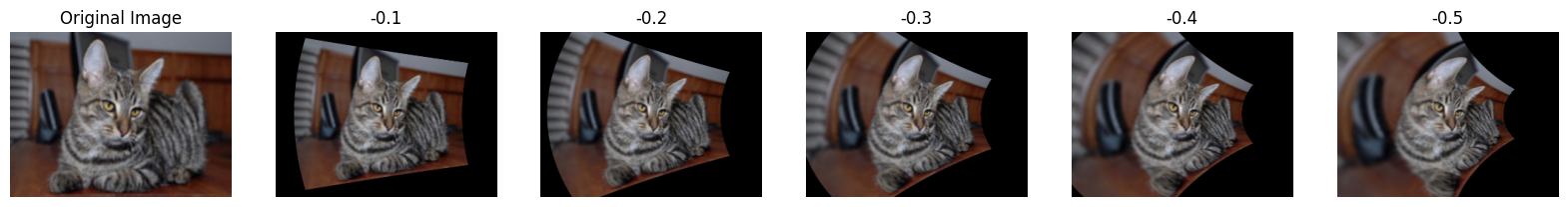}
    \vspace{-6mm}
    \caption{Demonstrate controlled scaling of distortion in MPD-transformed image. distortion is controlled by the intensity of the real component of complex parameter $c_{real}$ ranging from -0.1 to -0.5 to synthesize perspectively distorted left-views on example image of 'cat'.} 
    \label{fig:intensity}
    \vspace{-6mm}
\end{figure}

\vspace{-1.5mm}
The proposed MPD method mathematically models perspective distortion in the image using a family of Möbius transform. First, it maps the discrete coordinates $(x,y)$ of the input image $I$ from pixel space to complex vector $z = x+iy$ in complex space. Second, it applies Möbius transform to $z$ and achieves corresponding non-linear transformed representation $z_m =  x_m+iy_m$ by notably controlling real and imaginary parts of parameter $c$ in Eq. \ref{eq_mobius}. Next, the transformed complex vector $z_m$ is mapped to real-valued pixel space $(x_m,y_m)$, which is discretized to get discrete pixel coordinates  $(x_d,y_d)$. Obtained discrete pixel coordinates provide transformed image $I_{MPD}$ mimicking perspective distortion. Optionally, padding is applied to replace black background by boundary pixel to obtain integrated padding variant of transformed image. In this manner, MPD can synthesize perspective distortion artificially without estimating intrinsic and extrinsic parameters of perspective distortion. Noticeably, It eliminates the need for perspectively distorted real image data.
\begin{figure}[h]
\label{fig:algo1}
\vspace{-5mm}
\centering
\includegraphics[width=1\linewidth]{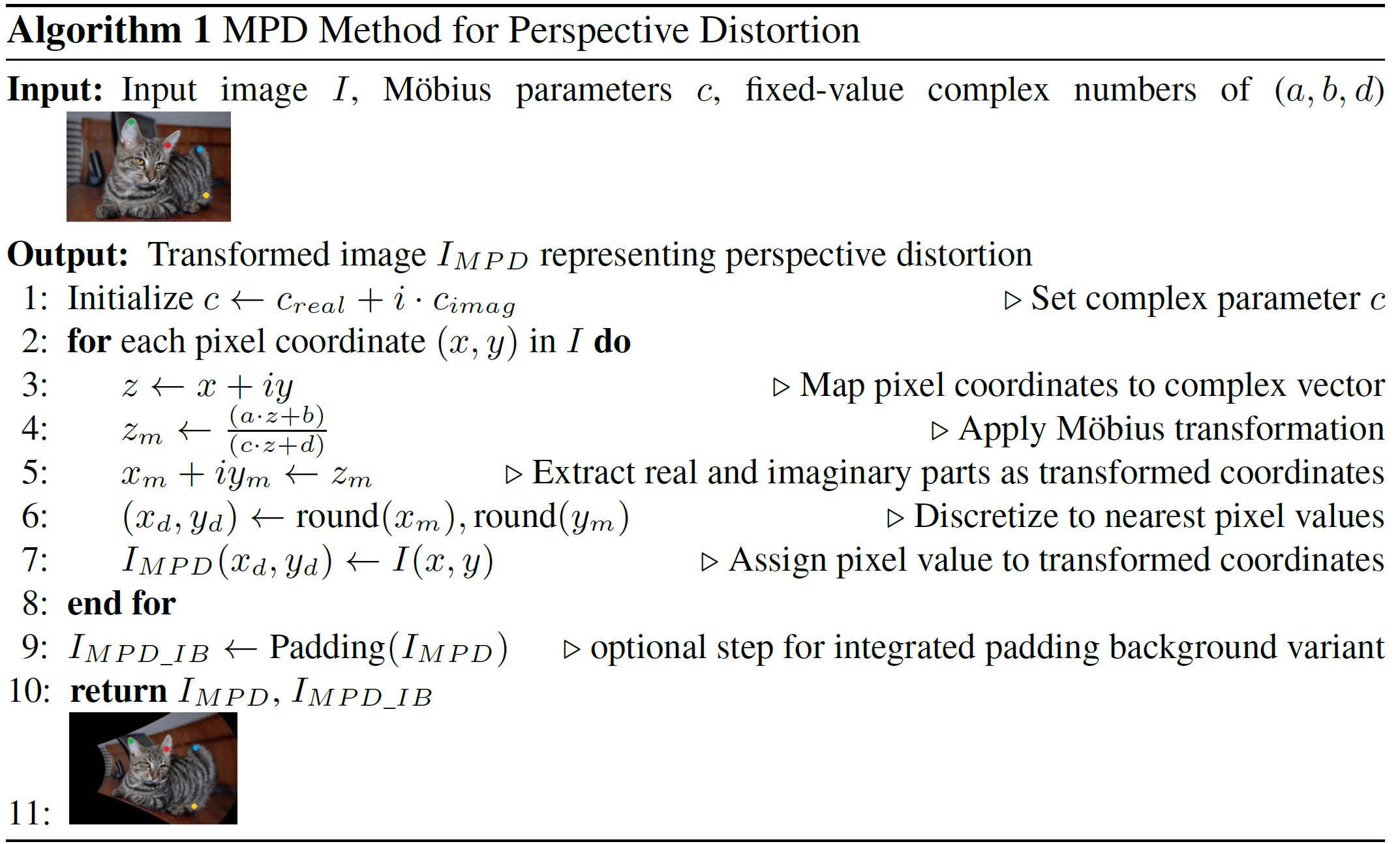}
\vspace{-9mm}
\end{figure}
The core part of MPD in modeling the PD lies in controlling the transformation through parameter $c$. Parameter $c$, being a complex number has real ($c_{real}$) and imaginary ($c_{imag}$) components; it is noteworthy that these components introduce real-world varieties of PD in the input image. $c_{real}$ perspectively distorts the in input image horizontally whereas $c_{imag}$ perspectively distorts the in input image vertically. 
The sign of $c_{real}$ dictates the direction ($+ve$ for left view and $-ve$ for right view) of distortion, whereas magnitude defines the intensity of applied distortion. Similarly, the sign of $c_{imag}$ dictates the direction ($+ve$ for top view and $-ve$ for bottom view) of distortion, whereas magnitude defines the intensity of applied distortion. The impact of signs of $c_{real}$ and imaginary $c_{imag}$ is depicted in Fig. \ref{fig:shematic}. The synthesis of the increasingly distorted left view of the input image is demonstrated in Fig. \ref{fig:intensity} by controlling the intensity of $c_{real}$ component of MPD. Demonstrations of increasingly distorted right, top, bottom, and other views are provided in Fig. \ref{fig:intensity_suppl} in the suppl. material. The main contribution of our MPD method is its ability to control parameterization for varying intensity and orientation, effectively representing real-world perspective distortions.
One of the ways to incorporate MPD is augmentation, which is investigated in this work. MPD is effortlessly employable to incorporate into the learning process to make networks robust against perspective distortion, materializing the proposed approach to shifting the focus from exclusion to inclusion of PD. The above-described method MPD is presented in Algorithm 1. MPD returned a transformed image of a 'cat' as shown in step 11. Four regions are marked with colors (green, red, blue, and yellow) to observe the correspondence between input and transformed images.
Although the current work does not focus on other parameters ($a$, $b$, and $d$), their impact on transformation is worth mentioning. The real component of parameters $a$ and $d$ controls scaling, and the imaginary components control rotations. The real component of parameter $b$ controls horizontal translations, and the imaginary component controls vertical translations. Their visualizations are provided in Fig. \ref{fig:a_b_d_suppl} in the suppl. material. 
The transformation parameters are set as: \( a = 1.0 + 0.0i \), \( b = 0.0 + 0.0i \), \( d = 1.0 + 0.0i \). The real and imaginary components of \( c \) range from \( 0.2 \) to \( 0.3 \) to mimic perspective distortion. These values are uniformly applied in all experiments unless otherwise stated. The influence of MPD to control perspective distortion is carried out via parameter $c$ and given probability \( P \) to which MPD is applied during training. 
We establish the following objectives to thoroughly assess the proposed methodology in the context of perspective distortion.
\begin{itemize}
    \item \textbf{Objective 1- Robustness evaluation of existing models:} Investigation of existing trained models for robustness against perspective distortion by evaluating them on the PD-dedicated newly developed benchmark ImageNet-PD.
    \item \textbf{Objective 2- MPD's effects on supervised learning:} Investigation of the effect of MPD in making supervised learning approach robust against perspective distortion.
    \item \textbf{Objective 3- MPD's effects on self-supervised learning:} Investigation of the effect of MPD in learning robust self-supervised representations against PD.
    \item \textbf{Objective 4- MPD's generalizability:} Investigating the effectiveness of MPD on cross-domain adaptability on diverse PD-affected real-world CV applications.
\end{itemize}
\section{Experiments \& Results}
\begin{figure}[!h!tb]
  \vspace{-6mm}
  \centering
  \includegraphics[width=.8\columnwidth]{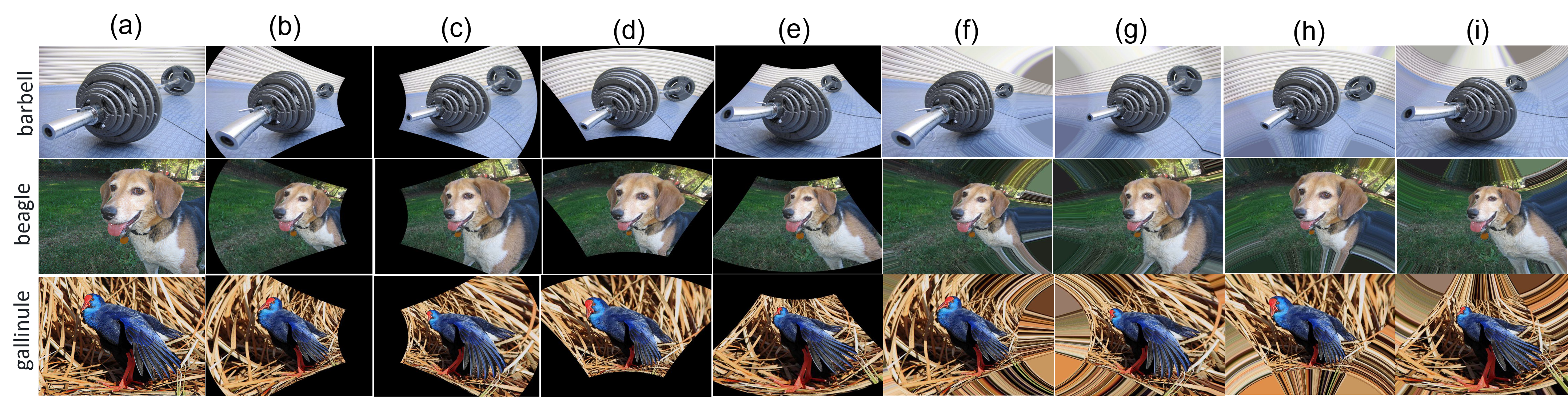}
  \vspace{-3mm}
  \caption{Perspectively distorted image examples from ImageNet-PD benchmark dataset. (a) Original image, (b) Left view (PD-L), (c) Right view (PD-R), (d) Top view (PD-T), (e) Bottom view (PD-B), (f) Left view with integrated padding background (PD-LI), (g) Right view with integrated padding background (PD-RI), (h) Top view with integrated padding background (PD-TI), (i) Bottom view with integrated padding background (PD-BI).}
  \label{fig:pd_datasets}
  \vspace{-5mm}
\end{figure}

\vspace{-3mm}
\textbf{Naming convention and experimental details: } \textit{MPD} defines the proposed transformation (shown in Fig. \ref{fig:shematic}). In addition, we also have another variant, \textit{MPD IB (integrated padding background)}, which replaces the black background with boundary foreground pixel values of the transformed image (shown in Fig. \ref{fig:mpd_ib_suppl} in suppl. material).
\textbf{ImageNet-PD} benchmark dataset has eight subsets (Fig. \ref{fig:pd_datasets}); four corresponding to four orientations (left, right, top, bottom) with black background (PD-L, PD-R, PD-T, PD-B). The other four subsets have the same orientations with integrated padding background, namely PD-LI (PD \textbf{L}eft -\textbf{I}ntegrated padding background), PD-RI, PD-TI, and PD-BI.

\textit{Standard ResNet50} denotes ResNet50 model \cite{he2016deep} with ImageNet-1k weights from torchvision library\cite{pytorch_vision}.
\textit{Supervised:MPD} and \textit{Supervised:MPD IB} indicate ResNet50 models trained on ImageNet\cite{deng2009imagenet} with MPD and MPD IB as augmentation respectively with defined probability \( P\), following training protocol in torchvision guide\cite{pytorch_vision}. 
\textit{ssl:MPD} and \textit{ssl:MPD IB} refer to ResNet50 encoders trained on ImageNet\cite{deng2009imagenet} incorporating MPD and MPD IB additional augmentation in a self-supervised approach using contrastive learning method, SimCLR\cite{chen2020simple} with \( P = 0.8 \), batch size of 512, and linear learning rate. After self-supervised pretraining (on a relatively smaller batch size of 512), models are fine-tuned for classification following the torchvision \cite{pytorch_vision} protocol.
\textit{ssl:DINO-MPD} refer to ViT small encoder trained on ImageNet\cite{deng2009imagenet} incorporating MPD transformation in a self-supervised approach using DINO \cite{caron2021emerging} method, with \( P = 0.8 \) and batch size of 512, following linear evaluation protocol described in \cite{caron2021emerging}. Details on MPD adaptations in CV applications are described in respective sections.
\subsection{Existing models' robustness evaluation} \label{ob1}

\subsubsection{New benchmark dataset- ImageNet-PD:}
To introduce perspective distortion by mimicking through the proposed transform MPD and its variant MPD-IB, we developed a new benchmarking dataset, ImageNet-PD (Fig. \ref{fig:pd_datasets}), derived from the ImageNet validation set. ImageNet-PD has eight subsets (Fig. \ref{fig:pd_datasets}), four corresponding to four orientations (left, right, top, bottom) with black background (PD-L, PD-R, PD-T, PD-B). The other four subsets have the same orientations but integrated padding backgrounds using boundary pixels (PD-LI, PD-RI, PD-TI, PD-BI).  
\begin{figure}[h]
  \centering
  \includegraphics[width=.8\columnwidth]{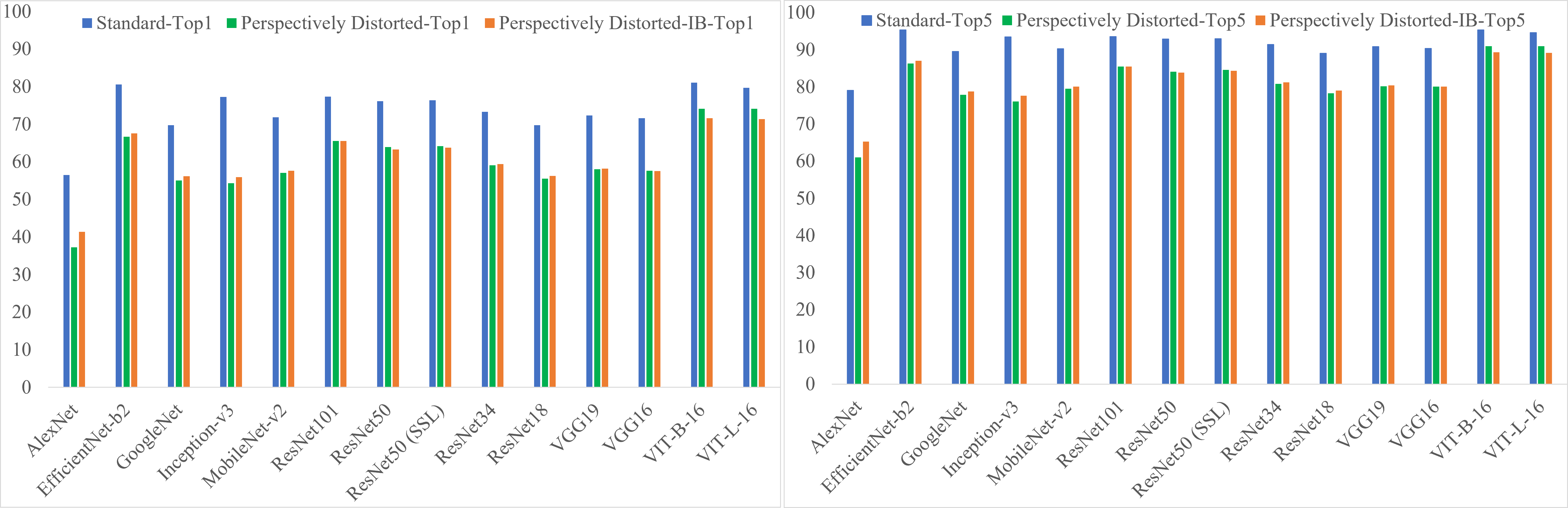}\\
  \vspace{-1mm}
  \caption{Top1 and Top5 accuracies of ImageNet trained models with standard architectures on ImageNet-PD subsets. Blue bars shows performance on original ImageNet validation set, green bars shows mean performance on ImageNet-PD subsets with black background, and orange bars shows mean performance on ImageNet-PD subsets with integrated padding background.}
  \label{fig:H1-top1-top5}
  \vspace{-3mm}
\end{figure}
\begin{figure*}[!t]
  \centering
  \includegraphics[width=0.7\columnwidth]{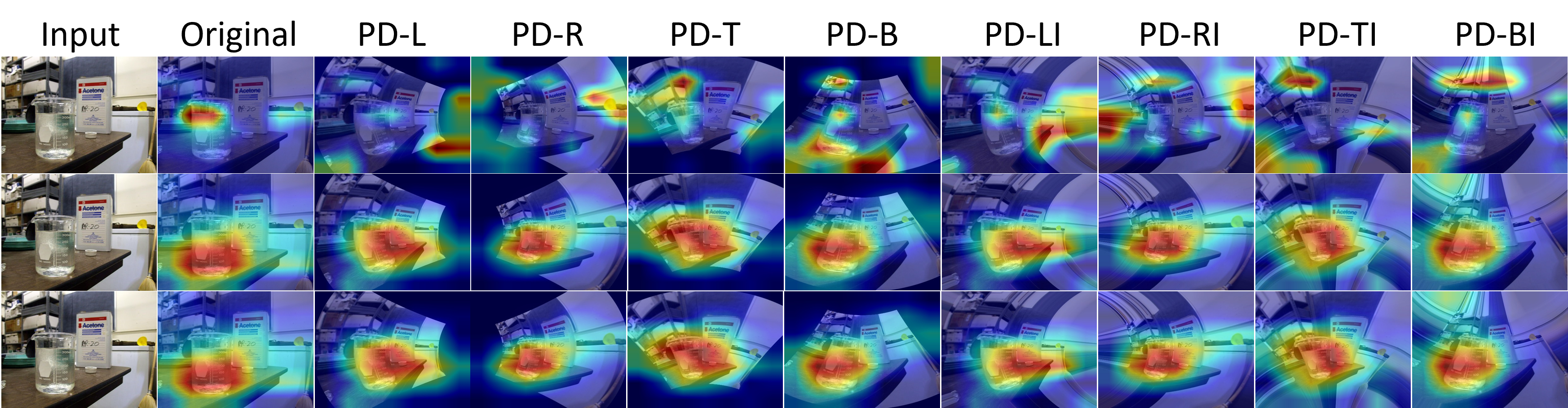}
  \vspace{-3mm}
  \caption{Activation maps of the 'beaker' example in ImageNet-PD subsets. Standard ResNet50 model (row 1), supervised:MPD (row 2), and ssl: MPD (row 3). } 
  \label{fig:cam}
  \vspace{-3mm}
\end{figure*}
In our evaluation across various deep learning architectures, from classical CNNs like ResNet, VGG, and AlexNet to modern structures like EfficientNet and Vision Transformers (ViT), we consistently observe a significant drop in model accuracy when faced with images exhibiting perspective distortions, refer Fig. \ref{fig:H1-top1-top5} (also in Tables \ref{tab:hype1A}, \ref{tab:hype1A_IB} in suppl. material). 
AlexNet shows a drop from 56.52\% to 37.25\% in Top-1 and from 79.06\% to 61.04\% in Top-5; EfficientNet-b2 falls from 80.60\% to 66.71\% in Top-1 and from 95.31\% to 86.30\% in Top-5; and even Vision Transformer models like VIT-B-16 decline from 81.07\% to 74.09\% in Top-1 and from 95.32\% to 90.98\% in Top-5, as demonstrated by activation maps for a standard ResNet50 (Fig.~\ref{fig:cam}, row 1). The uniform decrease in Top-1 and Top-5 accuracies across various models on ImageNet-PD vs. original ImageNet shows perspective distortion as a real challenge in computer vision.

\subsection{MPD's effects on supervised learning} \label{ob2}
\subsubsection{Existing Benchmarks:}
Both \textit{supervised:MPD} and \textit{supervised:MPD IB} models are trained on ImageNet dataset and evaluated on existing benchmarks and ImageNet-PD. ImageNet-X \cite{idrissi2022imagenet} and ImageNet-E \cite{li2023imagenet}) considered as they are partially affected by perspective distortion. \textbf{ImageNet-E}\cite{li2023imagenet} is a recent benchmark dataset specifically designed to evaluate the robustness of image classifiers concerning object attributes, including background settings, object sizes, spatial positioning, and orientation. \textbf{ImageNet-X}\cite{idrissi2022imagenet} introduces a set of sixteen human annotations focusing on attributes like pose, background, and lighting derived from the ImageNet. 
\begin{figure}[!t]
  \centering
  \includegraphics[width=0.6\columnwidth]{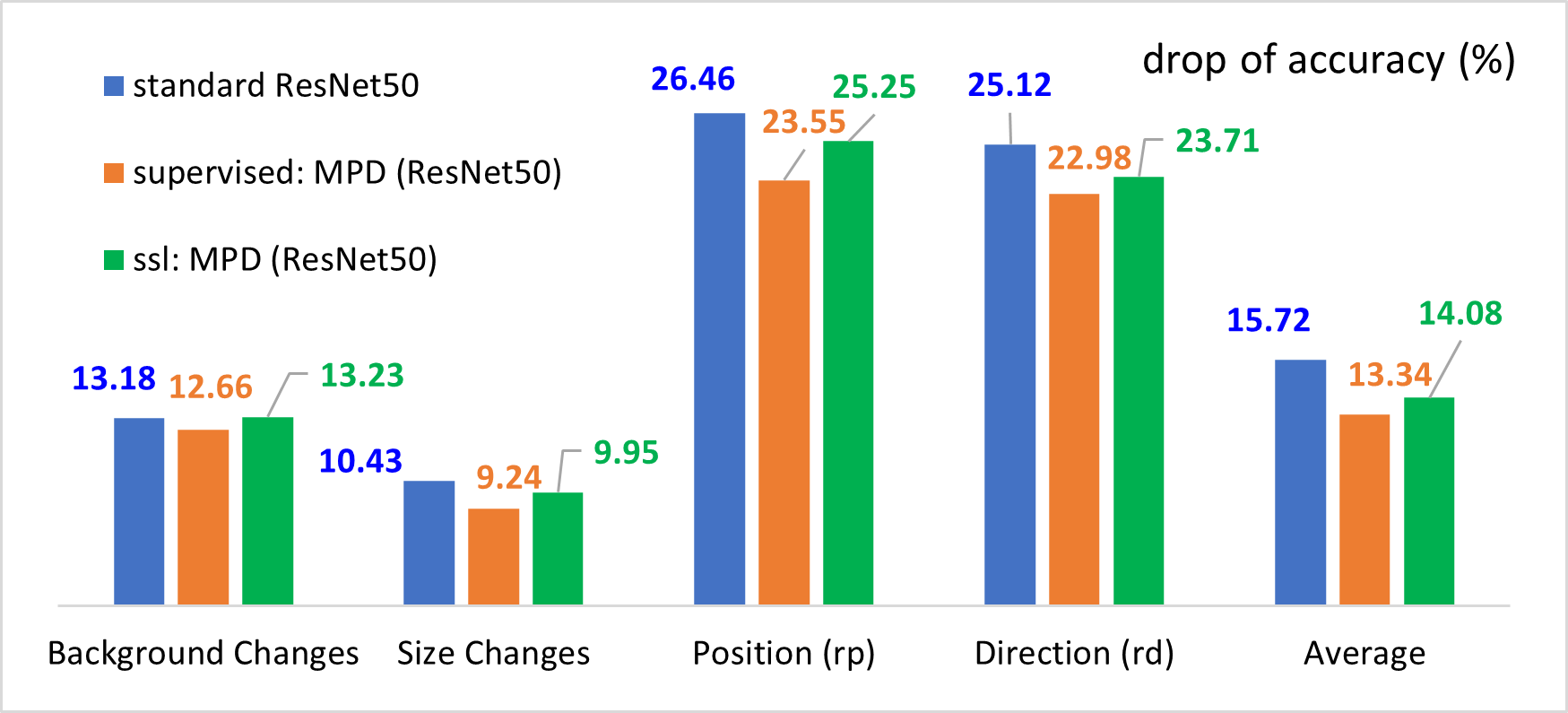}
  \vspace{-3mm}
  \caption{ImageNet-E\cite{li2023imagenet}: Drop of Top1 accuracy under background changes, size changes, random position (rp), random direction (rd), and average over 11 subsets. Lower is better. The mean is reported for size and background-related subsets. \textit{Average} reports the mean of all subsets.} 
  \label{fig:hypo2_ImageNet-E}
  \vspace{-6mm}
\end{figure}
\begin{figure}[!t]
  \centering
  \includegraphics[width=0.8\columnwidth]{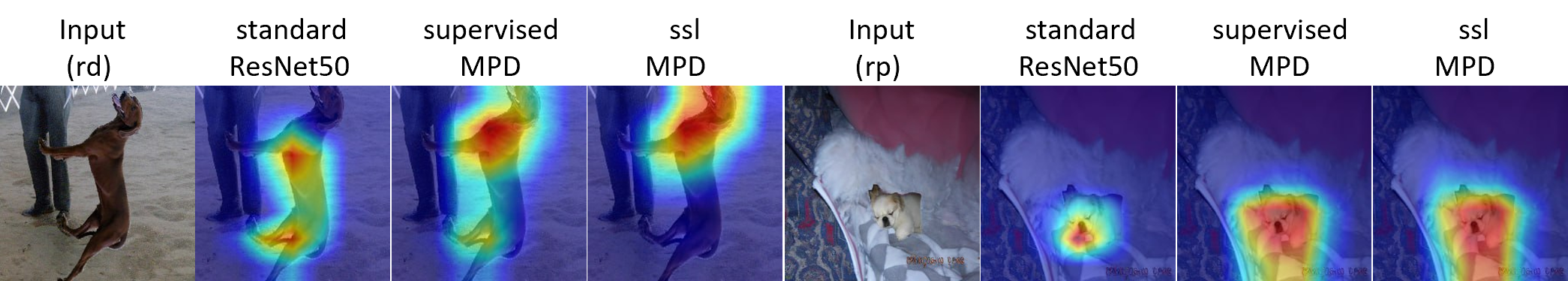}
  \vspace{-3mm}
  \caption{Activation maps (ImageNet-E). '\textit{redbone}' from rd and '\textit{peke}' from rp subset.} 
  \label{fig:hypo2_imagenet_e_cam}
  \vspace{-5mm}
\end{figure}
Fig. \ref{fig:hypo2_ImageNet-E} demonstrates that both models improved performance on ImageNet-E by and reduced the accuracy drop (improving absolute accuracy) in \textit{position} (2.91\%), \textit{direction} (2.14\%), and \textit{size} (1.19\%) changes; supported by activation maps in Fig. \ref{fig:hypo2_imagenet_e_cam}. Other variants of MPD in supervised and self-supervised approaches consistently follow the robustness trend. Results are in Table \ref{tab:tab:imagenet-e_detailed} and Fig. \ref{fig:imagnet_e_extended} in suppl. material. 
\begin{figure}[!t]
  \centering
  \includegraphics[width=0.75\columnwidth]{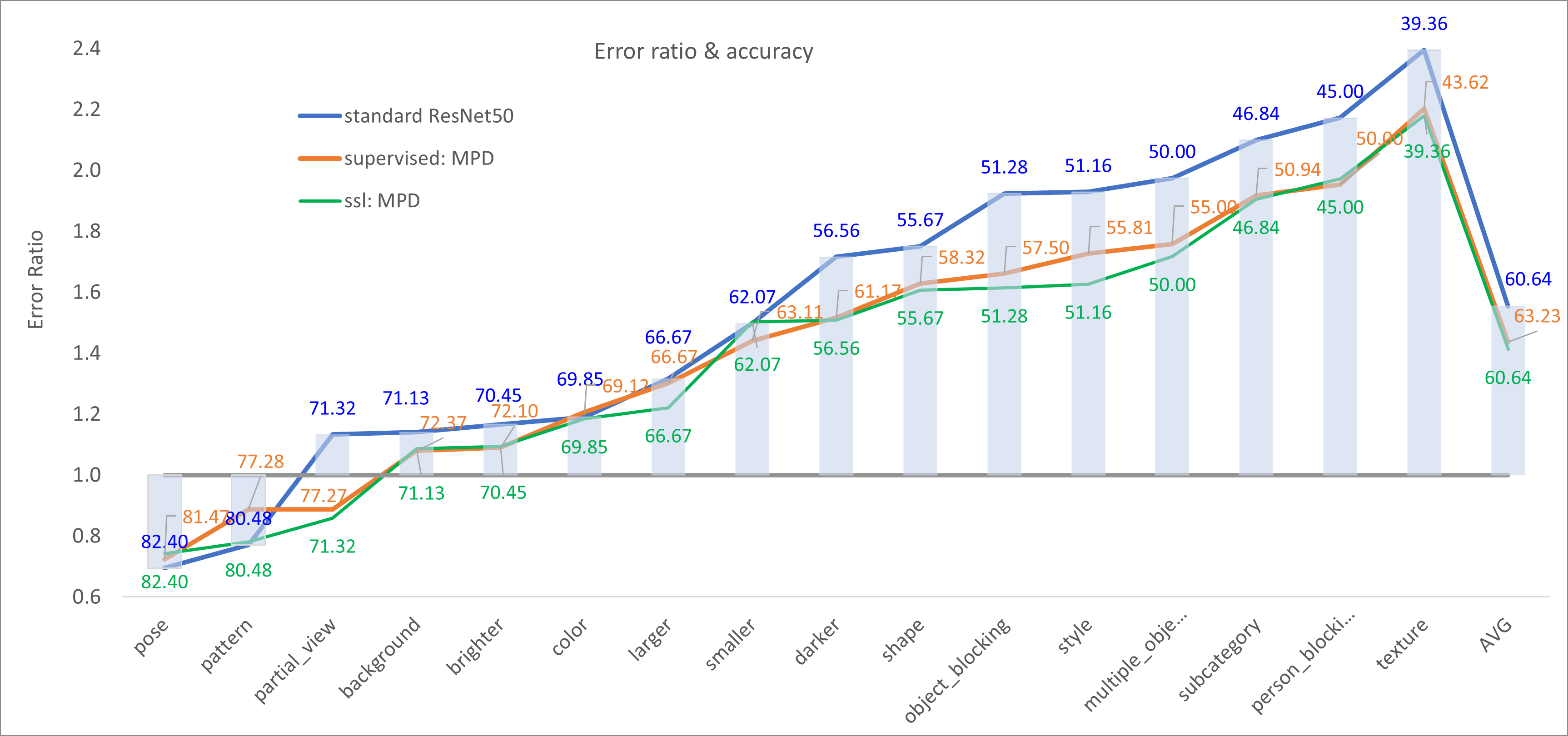}
  \vspace{-0mm}
  \caption{ImageNet-X\cite{idrissi2022imagenet}: Error ratio (close to 1.0 is best) of 16 factors in the plot. Accuracy is reported in numbers for each subset. \textit{Average} reports the mean accuracy of all factors. Both supervised:MPD and ssl:MPD models improved the error ratio with higher accuracy.} 
  \label{fig:hypo2_ImageNet-X}
  \vspace{-6mm}
\end{figure}
Both \textit{supervised:MPD} and \textit{supervised:MPD IB} models outperform standard ResNet50 on the ImageNet-X dataset\cite{idrissi2022imagenet}, achieving 63.23\% accuracy compared to ResNet50's 60.64\% and a lower error ratio of 1.44 versus 1.55 (refer Fig. \ref{fig:hypo2_ImageNet-X}). This improvement is particularly evident in handling perspective-related factors like size and shape variations. Detailed results are in Table \ref{tab:imagenet_x_detailed} in suppl. material.

\subsubsection{Benchmarking on ImageNet-PD:}
\begin{figure}[!t]
  \centering
  \includegraphics[width=0.8\columnwidth]{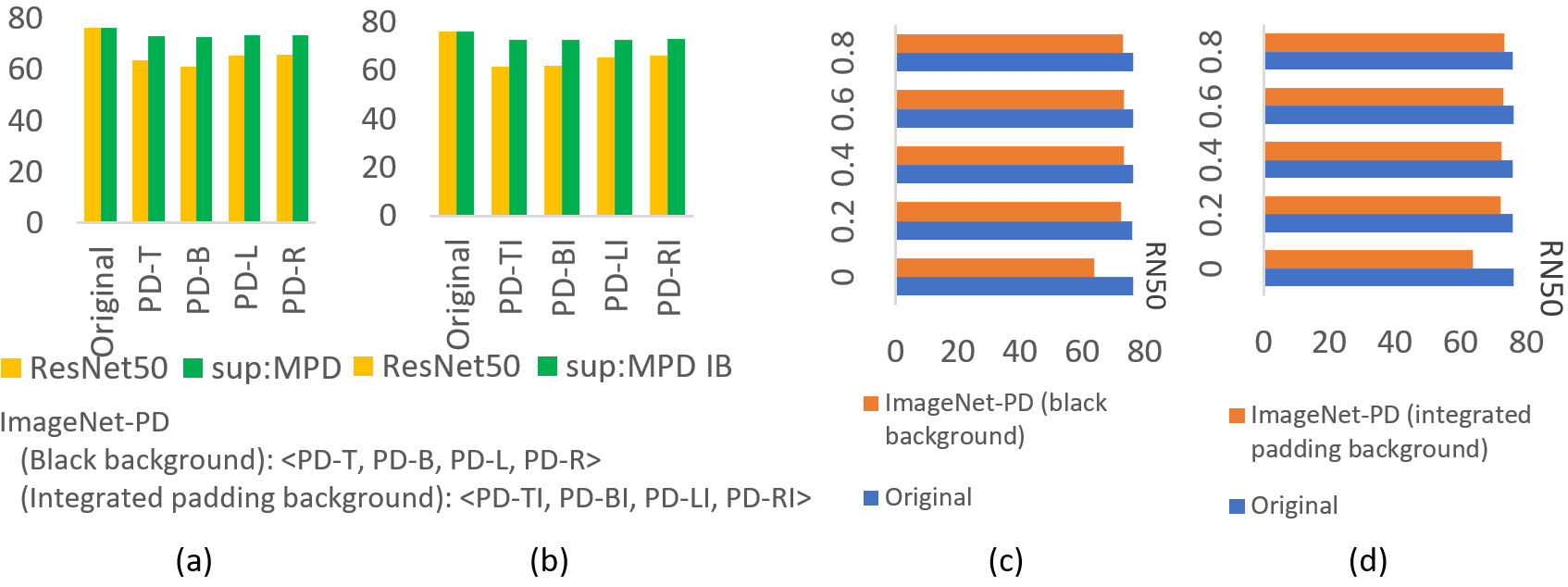}
  \vspace{-1mm}
  \caption{\textbf{(a)-(b)}[x-axis: ImageNet-PD subsets, y-axis: accuracy] $\rightarrow$ Top-1 accuracy comparison of standard ResNet50 \& \textit{supervised:MPD} and \textit{supervised:MPD IB} models on ImageNet-PD subsets. \textbf{(c)-(d)}[x-axis: accuracy, y-axis: probability]$\rightarrow$  Mean top-1 accuracy across ImageNet-PD subsets for \textit{supervised:MPD} and \textit{supervised:MPD IB} models with varied probability. 0.0 refers standard ResNet50. Original-original ImageNet validation set.} 
  \label{fig:hypo2_sup}
  \vspace{-5mm}
\end{figure}

Both \textit{supervised:MPD} and \textit{supervised:MPD IB} demonstrate notable improvements (Fig. \ref{fig:hypo2_sup} (a)-(b)) on ImageNet-PD subsets, achieving average Top-1 accuracy enhancements of 9.42\% and 9.54\% respectively, over the standard ResNet50 model.
\begin{table}[ht]
\vspace{-2mm}
\caption{Comparative analysis of MPD trained on supervised (\textit{supervised:MPD model}) and self-supervised (\textit{ssl:MPD model}) approaches and evaluated on original ImageNet and \textbf{ImageNet-PD subsets} (black background). The probability \(P\) of applying MPD in model fine-tuning is in the first column. More results on probabilities 0.0 and 1.0 in Table \ref{tab:Hype2-3main-standard_extended} in suppl. material.}
\vspace{-3mm}
\label{tab:Hype2-3main-standard}
\scriptsize
\begin{tabular}{ccccccccccc}
\hline
\multicolumn{1}{c|}{P} & \multicolumn{2}{c|}{Original Validation Set} & \multicolumn{2}{c|}{\begin{tabular}[c]{@{}c@{}}Perspectively Distorted\\ Top-view (PD-T)\end{tabular}} & \multicolumn{2}{c|}{\begin{tabular}[c]{@{}c@{}}Perspectively Distorted\\ Bottom-view (PD-B)\end{tabular}} & \multicolumn{2}{c|}{\begin{tabular}[c]{@{}c@{}}Perspectively Distorted \\ Left-view (PD-L)\end{tabular}} & \multicolumn{2}{c}{\begin{tabular}[c]{@{}c@{}}Perspectively Distorted\\ Right-view (PD-R)\end{tabular}} \\ \hline
\multicolumn{1}{c|}{} & Top1 & \multicolumn{1}{c|}{Top 5} & Top1 & \multicolumn{1}{c|}{Top 5} & Top1 & \multicolumn{1}{c|}{Top 5} & Top1 & \multicolumn{1}{c|}{Top 5} & Top1 & Top 5 \\ \hline
\multicolumn{11}{c}{\textbf{Supervised training from scratch}} \\ \hline
\multicolumn{1}{c|}{0.0} & 76.13±0.04 & \multicolumn{1}{c|}{92.86±0.01} & 63.37±0.06 & \multicolumn{1}{c|}{83.61±0.02} & 61.15±0.04 & \multicolumn{1}{c|}{81.86±0.01} & 65.20±0.03 & \multicolumn{1}{c|}{85.13±0.03} & 65.84±0.06 & 85.64±0.02 \\ \hline
\multicolumn{1}{c|}{0.2} & 76.05±0.03 & \multicolumn{1}{c|}{92.99±0.02} & 72.48±0.02 & \multicolumn{1}{c|}{91.02±0.01} & 72.12±0.02 & \multicolumn{1}{c|}{91.08±0.02} & 72.57±0.03 & \multicolumn{1}{c|}{91.13±0.03} & 72.80±0.05 & 91.35±0.02 \\
\multicolumn{1}{c|}{0.4} & 76.17±0.04 & \multicolumn{1}{c|}{93.03±0.02} & 73.13±0.03 & \multicolumn{1}{c|}{91.33±0.01} & 72.94±0.02 & \multicolumn{1}{c|}{91.42±0.01} & 73.19±0.06 & \multicolumn{1}{c|}{91.48±0.01} & 73.38±0.01 & 91.69±0.01 \\
\multicolumn{1}{c|}{0.6} & 76.19±0.05 & \multicolumn{1}{c|}{93.14±0.03} & 73.23±0.02 & \multicolumn{1}{c|}{91.46±0.03} & 73.01±0.05 & \multicolumn{1}{c|}{91.34±0.02} & 73.47±0.02 & \multicolumn{1}{c|}{91.66±0.01} & 73.54±0.06 & 91.61±0.02 \\
\multicolumn{1}{c|}{0.8} & 76.34±0.02 & \multicolumn{1}{c|}{93.03±0.02} & 73.00±0.02 & \multicolumn{1}{c|}{90.69±0.02} & 72.31±0.03 & \multicolumn{1}{c|}{90.81±0.03} & 73.50±0.04 & \multicolumn{1}{c|}{91.33±0.03} & 72.91±0.02 & 91.29±0.02 \\ \hline
\multicolumn{11}{c}{\textbf{Self-supervised  pre-taining on contrastive learning integrating MPD (probability for pre-training=0.8)}} \\ \hline
\multicolumn{1}{c|}{0.2} & 76.37±0.05 & \multicolumn{1}{c|}{93.60±0.02} & 72.34±0.02 & \multicolumn{1}{c|}{90.81±0.02} & 72.24±0.03 & \multicolumn{1}{c|}{90.95±0.02} & 72.48±0.04 & \multicolumn{1}{c|}{91.08±0.01} & 72.81±0.03 & 91.16±0.01 \\
\multicolumn{1}{c|}{0.4} & 76.77±0.02 & \multicolumn{1}{c|}{93.40±0.01} & 73.23±0.03 & \multicolumn{1}{c|}{91.39±0.01} & 73.06±0.04 & \multicolumn{1}{c|}{91.54±0.02} & 73.55±0.03 & \multicolumn{1}{c|}{91.53±0.03} & 73.54±0.05 & 91.65±0.03 \\
\multicolumn{1}{c|}{0.6} & 76.14±0.04 & \multicolumn{1}{c|}{93.58±0.02} & 73.39±0.02 & \multicolumn{1}{c|}{91.50±0.02} & 73.29±0.03 & \multicolumn{1}{c|}{91.57±0.01} & 73.57±0.03 & \multicolumn{1}{c|}{91.66±0.02} & 73.73±0.04 & 91.58±0.02 \\
\multicolumn{1}{c|}{0.8} & 76.29±0.03 & \multicolumn{1}{c|}{92.78±0.02} & 73.61±0.04 & \multicolumn{1}{c|}{91.49±0.03} & 73.27±0.05 & \multicolumn{1}{c|}{91.48±0.03} & 73.60±0.06 & \multicolumn{1}{c|}{91.69±0.01} & 73.81±0.02 & 91.71±0.02 \\ \hline
\end{tabular}
\vspace{-3mm}
\end{table}
Please refer to Table \ref{tab:Hype2-3main-standard} and Table \ref{tab:Hype2-3main-IB-extended} for detailed comparison; Table \ref{tab:Hype2-3main-IB-extended} is in suppl. material.
Concurrently, these models maintain performance on the original ImageNet validation set, highlighting their efficacy and adaptability amidst perspective distortion. The activation maps shown in Fig.~\ref{fig:cam} (row 2) further corroborate this outperforming trend. Ablation on the probability of applying MPD and MPD IB as a data augmentation (Fig. \ref{fig:hypo2_sup} (c)-(d)) exhibits consistent performance. MPD also compared with other popular augmentation methods (Mixup \cite{zhang2018mixup}, Cutout \cite{devries2017improved}, AugMix \cite{hendrycksaugmix}, and Pixmix \cite{hendrycks2022pixmix}) on ImageNet-PD and ImageNet-E, consistently outperforming them. Detailed results are in the suppl. material (refer to Table \ref{tab:imagenetaugs}).

\subsection{MPD's effects on self-supervised learning} \label{ob3}
MPD as data augmentation integrated in SimCLR \cite{chen2020simple} and DINO \cite{caron2021emerging} (small ViT \cite{dosovitskiy2020image} backbone), as in Fig. \ref{fig:simclr_MPD} and \ref{fig:DINO_MPD} in suppl. material. Models are pre-trained and finetuned / liner-probed on ImageNet and evaluated on existing benchmarks and ImageNet-PD.
\subsubsection{Existing Benchmarks:}
\textit{ssl:MPD} and \textit{ssl:MPD IB} models demonstrate efficacy in handling perspective-related variations in \textbf{ImageNet-E} as shown in Fig. \ref{fig:hypo2_ImageNet-E}. Specifically, it shows a reduction in the accuracy drop for position (1.21\%), direction (1.41\%), and size (0.48\%) changes compared to the standard ResNet50, indicating its robustness. On the \textbf{ImageNet-X} benchmark, \textit{ssl:MPD} and \textit{ssl:MPD IB} models outperform with an average accuracy of 62.99\%, compared to the standard ResNet50's 60.64\%, and also show a reduced error ratio of 1.41 against the ResNet50's 1.55 (refer Fig. \ref{fig:hypo2_ImageNet-X}). Notably, performance improvement is evident in managing variations in size and shape, showcasing its robustness against PD. Detailed results are in Table \ref{tab:imagenet_x_detailed} in suppl. material.
\subsubsection{Benchmarking on ImageNet-PD:}
\begin{figure}[!t]
  \centering
  \includegraphics[width=0.8\linewidth]{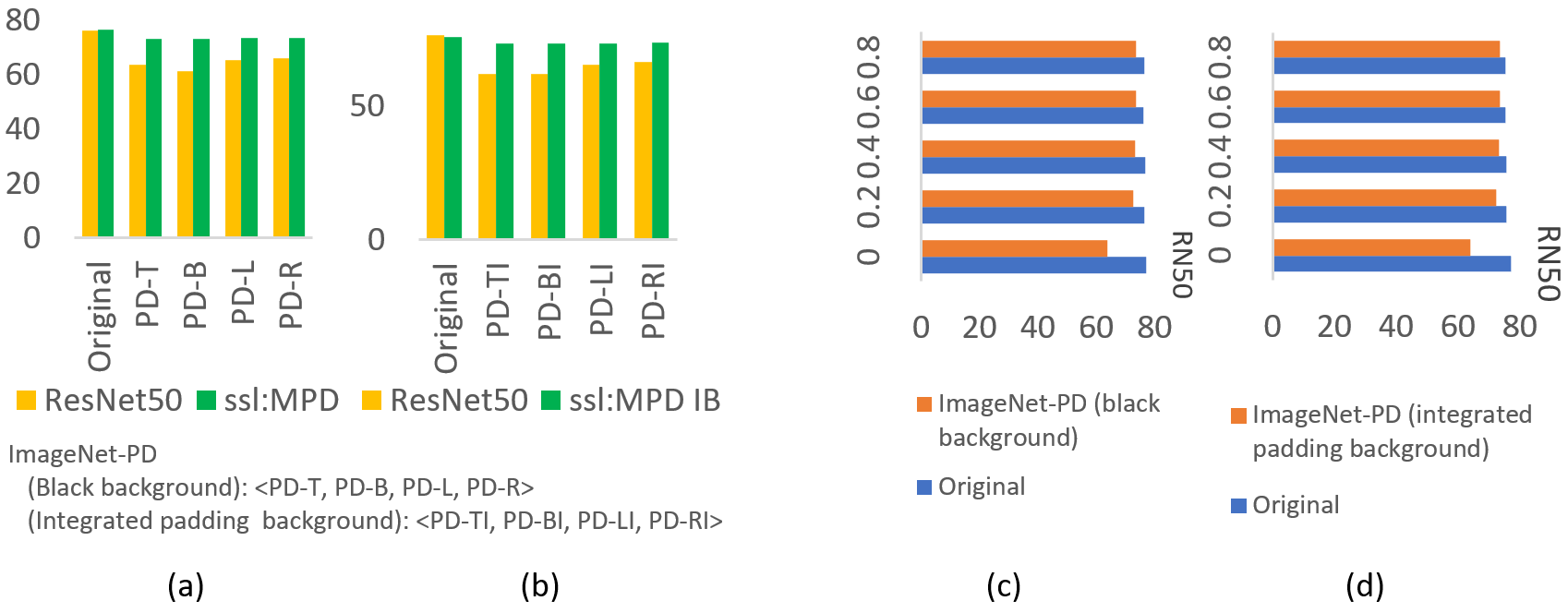}
  \vspace{-3mm}
  \caption{\textbf{(a)-(b)}[x-axis: ImageNet-PD subsets, y-axis: accuracy] $\rightarrow$ Top-1 accuracy comparison of standard ResNet50 and \textit{ssl:MPD} \& \textit{ssl:MPD IB} models on ImageNet-PD subsets. \textbf{(c)-(d)}[x-axis: accuracy, y-axis: probability]$\rightarrow$ Mean top-1 accuracy across ImageNet-PD subsets for \textit{ssl:MPD} model with varied probability. 0.0 refers standard ResNet50. } 
  \label{fig:hypo3_ssl}
  \vspace{-7.5mm}
\end{figure}
\begin{figure}[h]
   \vspace{-7mm}
  \centering
  \includegraphics[width=.8\columnwidth]{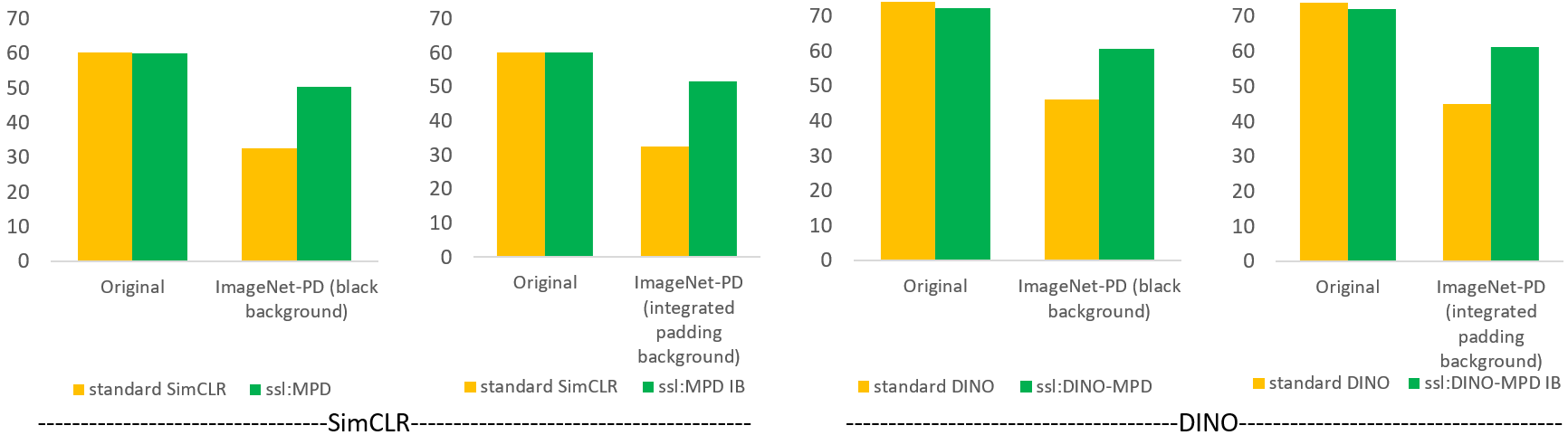}
  \vspace{-3mm}
  \caption{Linear evaluation of SimCLR \cite{chen2020simple} and DINO \cite{caron2021emerging} with MPD and MPD IB augmentations.} 
  \label{fig:simclr_MPD_MPD_ib_1}
  \vspace{-4mm}
\end{figure}

\textit{ssl:MPD} model (Fig. \ref{fig:hypo3_ssl} (a)) shows an average improvement of approximately 10.32\% across the PD subsets, while the \textit{ssl:MPD IB} (Fig. \ref{fig:hypo3_ssl} (b)) demonstrates an average increase of around 10.14\%. In addition, both variants display robust performance on the original ImageNet validation set. Please refer to Table \ref{tab:Hype2-3main-standard} and Table \ref{tab:Hype2-3main-IB-extended} for detailed comparison in suppl. material. The activation maps depicted in Fig.~\ref{fig:cam} (row 3) further validate the improved performance. Moreover, ablation on the probability of applying MPD and MPD IB as data augmentation (Fig. \ref{fig:hypo3_ssl} (c)-(d)) demonstrates a consistent increase in performance. We have also performed linear evaluations on self-supervised method SimCLR \cite{chen2020simple} and DINO \cite{caron2021emerging} (refer Table \ref{tab:Hype3-DINO-MPD-LE}, \ref{tab:Hype3-DINO-MPD-IB-LE}; Fig. \ref{fig:dino_attn_example} for detailed results in suppl. material). Similarly, significant improvement is noticed on ImageNet-PD subsets while maintaining the performance on standard ImageNet as shown in Fig. \ref{fig:simclr_MPD_MPD_ib_1} (refer Table \ref{tab:Hype3-MPD-LE}, \ref{tab:Hype3-MPDIB-LE} for detailed results in suppl. material).
\subsection{ MPD's generalizability} \label{ob4}
This section discusses MPD's adaptability in diverse applications, namely crowd counting (CC), fisheye, and person re-identification, and challenging CV tasks such as object detection, where perspective distortion is evident.
\subsubsection{Crowd counting:} \label{subsec:cc}
Numerous human heads within a single image present challenges due to the variations in size, orientation, and shape, naturally including perspective distortion. We investigate the potential of MPD in crowd counting on several publicly available datasets, including ShanghaiTech (SHTech) Part A and Part B\cite{7780439} and UCF-CC50\cite{idrees2013multi}. The ShanghaiTech Part A features highly congested scenes, and Part B consists of sparser scenes captured on a busy street. The UCF-CC50 dataset comprises 50 diverse scenes featuring a considerable variation in the crowd.  

\textbf{MPD-CC}: MPD is adapted to CC by extending its capability to transform scene images and their CC labels, as mentioned in steps 4-7 of algorithm 2.  

\begin{figure}[h]
   \vspace{-5mm}
  \centering
\includegraphics[scale=0.35]{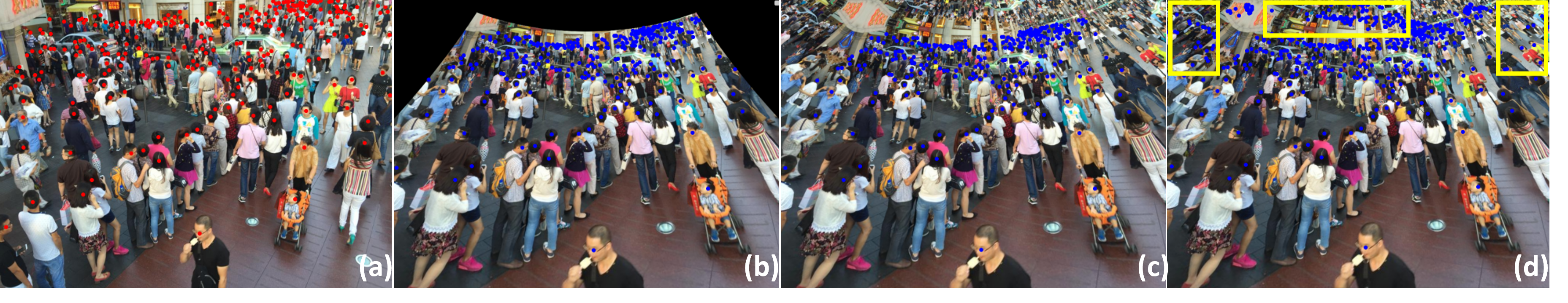}
  \vspace{-4mm}
  \caption{MPD-AutoCrowd: Artificially augments crowd \& labels. (a) Original image with labels, (b) transformed image \& labels, (c) Transformed image with augmented crowd, (d) Transformed image with augmented crowd \& labels (yellow box).} 
    \vspace{-6mm}
\label{fig:crowd}
\end{figure}

\textbf{MPD-AutoCrowd}:  We introduce MPD-AutoCrowd (Algorithm 2), a novel approach for the automated augmentation of the crowd in scene images by upgrading MPD-CC. It seamlessly integrates synthetic crowds into the background regions of transformed images, complete with corresponding labels. This approach facilitates scene-level crowd augmentation without human supervision (Fig. \ref{fig:crowd}).
MPD-CC and MPD-AutoCrowd are integrated with P2P-Net\cite{song2021rethinking} to evaluate the effectiveness of MPD in crowd counting. Training parameters, CNN architecture (VGG16), and evaluation configuration are used as mentioned in P2P-Net\cite{song2021rethinking}. The probability of applying MPD-CC and MPD AutoCrowd is 0.1, and real and imaginary components of \( c \) range from \( 0.2 \) to \( 0.6 \). This method has yielded improved results, surpassing previous state-of-the-art performance. Refer to Table \ref{tab:hype5A} for a detailed comparison with other CC methods. MPD-CC and MPD-AutoCrowd outperform on multiple benchmarks. 
  \begin{figure}[h!t]
  \vspace{-2mm}
    \centering
    \vspace{-3mm}
    \includegraphics[width=1\columnwidth]{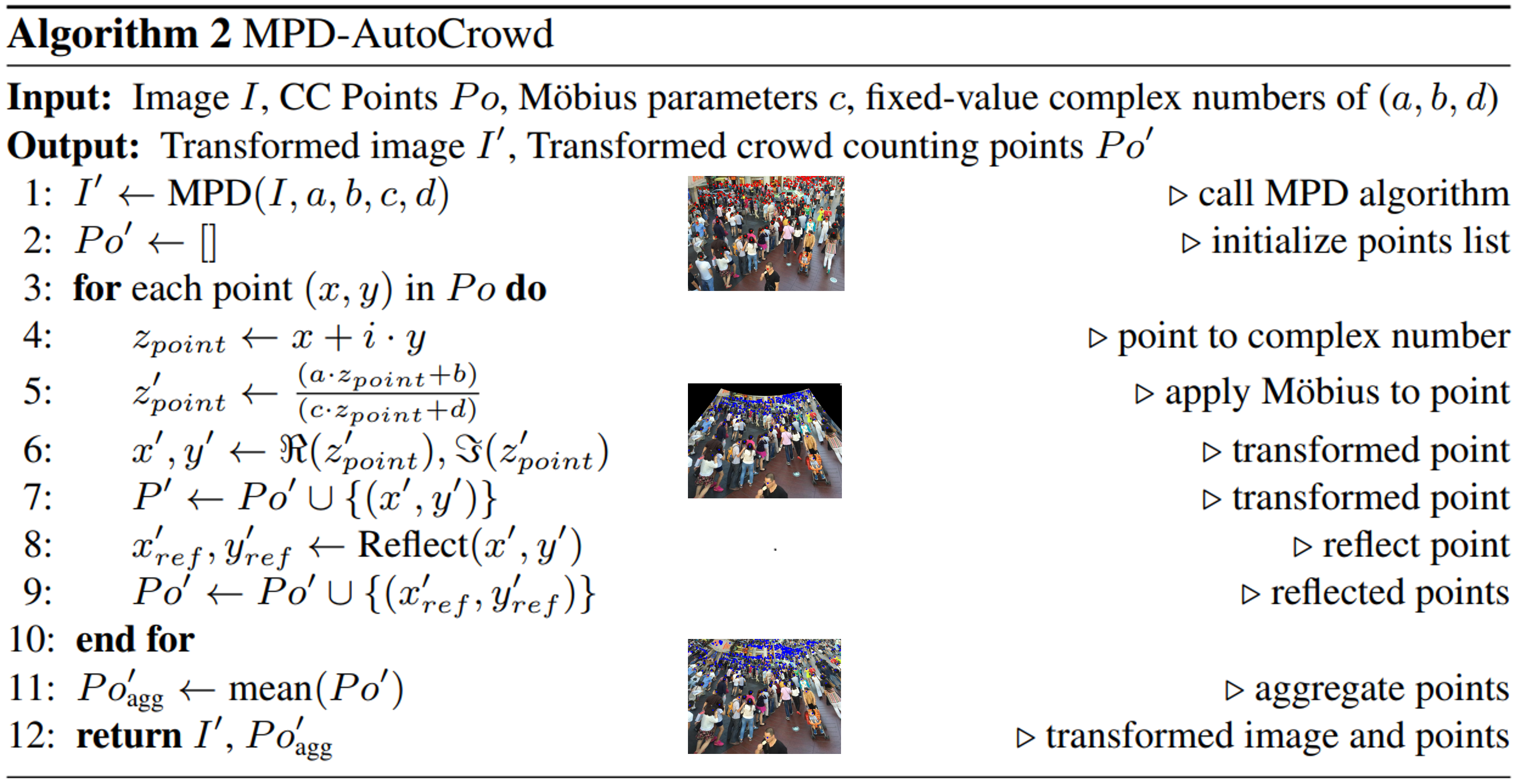}
    \vspace{-8mm}
\end{figure}
MPD-AutoCrowd has a MAE of 50.81 on SHTech Part A and 6.73 on SHTech Part B. The performance on the UCF-CC50 ( a highly dense crowds) dataset, where MPD-AutoCrowd has an MAE of 96.80 and a Mean Squared Error (MSE) of 139.50, significantly outperforms other models. We further improved results on UCF-CC50 with ResNet50 encoder as shown in Table \ref{tab:MPD_auto_crowd_extended} in suppl. material.
We also adapted other perspective transforms\cite{zhang2020perspective}\cite{papakipos2022augly} and RandAug\cite{cubuk2020randaugment} augmentation for crowd counting and compared with MPD-CC and MPD-AutoCrowd which shows that our proposed methods outperform. Details are in suppl. material (refer Table \ref{tab:cc_data_augs} and Fig. \ref{fig:augs_cc} in sec. \ref{sec:supp_cc}).
\begin{table}[]
\vspace{-2mm}
\caption{Comparison of different crowd-counting methods with the proposed work. MPD-CC \& MPD-AutoCrowd outperform on multiple benchmarks}
\vspace{-1mm}
\label{tab:hype5A}
\scriptsize
\centering
\begin{tabular}{@{\hspace{-1.5pt}}c@{\hspace{-1.5pt}}c@{\hspace{0.5pt}}c@{\hspace{-1.5pt}}c@{\hspace{-1.5pt}}c@{\hspace{-1.5pt}}c@{\hspace{-1.5pt}}c@{\hspace{-1.5pt}}c@{\hspace{-1.5pt}}}
\hline
\textbf{Method} & \textbf{Venue} & \multicolumn{2}{c}{\textbf{SHTech PartA}} & \multicolumn{2}{c}{\textbf{SHTech PartB}} & \multicolumn{2}{c}{\textbf{UCF\_CC-50}} \\ \hline
 &  & MAE & MSE & MAE & MSE & MAE & MSE \\
CAN\cite{liu2019context} & CVPR'19 & 62.3 & 100 & 7.8 & 12.2 & 212.2 & 243.7 \\
Bayesian+\cite{ma2019bayesian} & ICCV'19 & 62.8 & 101.8 & 7.7 & 12.7 & 229.3 & 308.2 \\
S-DCNet\cite{xiong2019open} & ICCV'19 & 58.3 & 95 & 6.7 & 10.7 & 204.2 & 301.3 \\
SA+SPANet\cite{cheng2019learning} & ICCV'19 & 59.4 & 100 & 6.5 & 9.9 & 232.6 & 311.7 \\
SDANet\cite{miao2020shallow} & AAAI'20 & 63.6 & 101.8 & 7.8 & 10.2 & 227.6 & 316.4 \\
ADSCNet\cite{bai2020adaptive} & CVPR'20 & 55.4 & 97.7 & 6.4 & 11.3 & 198.4 & 267.3 \\
ASNet\cite{jiang2020attention} & CVPR'20 & 57.78 & 90.13 & - & - & 174.84 & 251.63 \\
AMRNet\cite{liu2020adaptive} & ECCV'20 & 61.59 & 98.36 & 7.02 & 11 & 184 & 265.8 \\
AMSNet\cite{hu2020count} & ECCV'20 & 56.7 & 93.4 & 6.7 & 10.2 & 208.4 & 297.3 \\
DM\cite{wang2020distribution} & NeurIPS'20 & 59.7 & 95.7 & 7.4 & 11.8 & 211 & 291.5 \\
P2P-Net\cite{song2021rethinking} & ICCV'21 & 52.74 & 85.06 & \textbf{6.25} & \textbf{9.9} & 172.72 & 256.18 \\
SDA(best)\cite{ma2021towards} & ICCV'21 & 52.90 & 87.30 & - & - & 159.1 & 239.4 \\
ChfL\cite{shu2022crowd} & CVPR'22 & 57.50 & 94.30 & 6.90 & 11.0 & - & - \\
DMCNet\cite{wang2023dynamic} & WACV'23 & 58.46 & 84.55 & 8.64 & 13.67 & - & - \\ \hline
MPD-CC & ours & \textbf{51.93} & \textbf{84.3} & {6.73} & {9.82} & \textbf{101.30} & \textbf{140.65} \\
\begin{tabular}[c]{@{}c@{}}MPD-\\ AutoCrowd\end{tabular} & ours & \textbf{50.81} & \textbf{85.01} & {\textbf{6.61}} & {\textbf{9.58}} & \textbf{96.80} & \textbf{139.50} \\ \hline
\end{tabular}
\vspace{-7mm}
\end{table}
\subsubsection{Transfer learning on fisheye images} \label{subsec:fisheye}
\begin{figure}[ht]
    \vspace{-4mm}
    \centering
    \begin{minipage}[t]{0.45\linewidth}
                \centering
          \includegraphics[width=1\columnwidth]{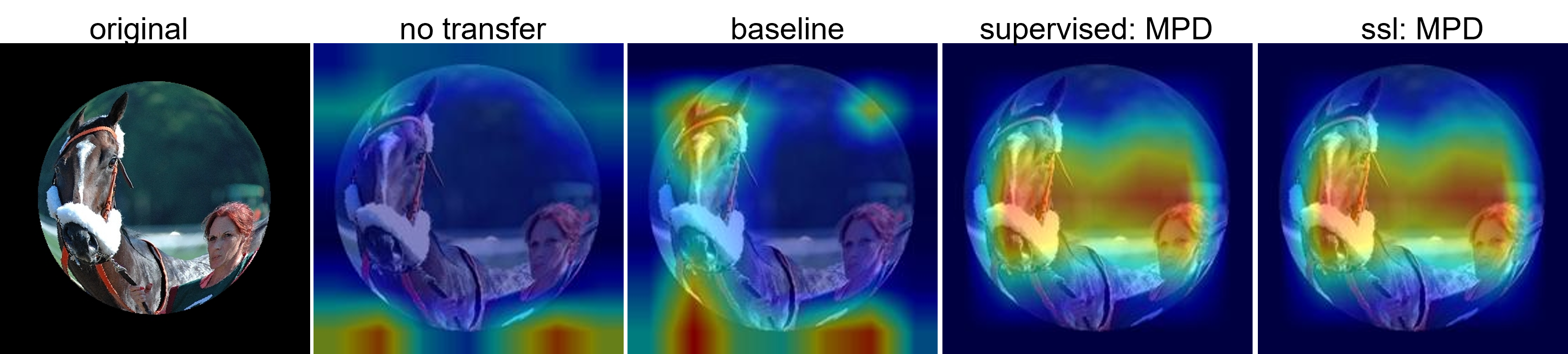}
          \caption{Activation maps of 'horse' and 'person' on VOC-360 dataset.} 
          \label{fig:voc_cam}
            \end{minipage}
            \hfill 
    \begin{minipage}[t]{0.45\linewidth}
        \centering
        \vspace{-16mm}
        \scriptsize
        \captionof{table}{MPD performance on VOC-360.}
        \label{tab:VOC-main}
        \begin{tabular}{cccc}
            \hline
            \textbf{Transfer   Learning} & \textbf{Accuracy} & \textbf{Precision} & \textbf{Recall} \\ \hline
            No transfer   learning & 82.02 & 94.90 & 89.96 \\
            Baseline & 91.32 & 97.24 & 95.67 \\ \hline
            supervised:   MPD IB & 91.87 & 97.00 & 96.22 \\
            supervised:   MPD & 92.69 & 97.76 & 96.23 \\
            ssl:   MPD IB & 92.43 & 97.16 & 96.50 \\
            ssl: MPD & \textbf{94.79} & \textbf{98.17} & \textbf{97.44} \\ \hline
        \end{tabular}
    \end{minipage}
\vspace{-4mm}
\end{figure}

Fisheye image distortions, caused by a wide field of view, are handled by spherical perspective models in \cite{Zhang2021-kv,kumar2023surround}. MPD simulates spherical views in pixel space and improves representation learning for fisheye images. We showcase the transferable capability of models trained with MPD and MPD IB through supervised and self-supervised methods on the ImageNet dataset, applying them to the VOC-360 fisheye dataset\cite{fu2019datasets}. This dataset has 39,575 fisheye images with multi-label classification labels. We fine-tune these models on the VOC360, running for 100 epochs with a batch size 16. We show the performance of supervised and self-supervised models for the multi-label classification task in Table \ref{tab:VOC-main}. ResNet50 is used as backbone. Ablations on label efficiency (Table \ref{tab:MPD_voc_prob_labels}, \ref{tab:MPD_ib_voc_prob_labels}, and Fig. \ref{fig:voc_label_eff}) are in suppl. material.
\textit{supervised:MPD} and \textit{supervised:MPD IB} models fine-tuned on VOC-360 dataset achieve accuracies 92.69\% and 91.87\% respectively. \textit{ssl:MPD} and \textit{ssl:MPD IB} models fine-tuned on VOC-360 achieve 92.43\% and 94.79\% respectively. \textit{ssl:MPD} performs best among all models. Fig. \ref{fig:voc_cam} depicts the activation maps.

\subsubsection{Person Re-Identification:} \label{subsec:reident}
The DeepSportRadar re-identification dataset\cite{van2022deepsportradar}  includes video frames with varying poses and camera angles, offering training (436 queries, 8133 gallery images) and testing sets (50 queries, 910 gallery images) to address perspective variability in re-identification tasks. We incorporated the MPD within the state-of-the-art CLIP-ReIdent method\cite{habel2022clip}. The probability of applying MPD is 0.1, and real \& imaginary components of \( c \) range from \( 0.2 \) to \( 0.6 \). CLIP-ReIdent is CLIP\cite{radford2021learning}-adapted contrastive image-image pre-training for person re-identification, leveraging transformer architecture.
MPD-incorporated Clip-ReIdent models improve performance with both backbone architectures, ViT-L-14 and ResNet50x16 (ref. Table \ref{tab:hype6}). Ablations on MPD parameters are in suppl. material in Table \ref{tab:hype_6}. 
In addition, it boosts performance across different frame counts, achieving competitive results with limited frames and performance further enhanced by longer training (refer Table \ref{tab:frame-reident}). 
\begin{table}[h]
\vspace{-5mm}
\begin{minipage}[b]{0.45\linewidth}
\caption{MPD improves the performance of the person re-identification. RR:re-ranking.}
\vspace{-3mm}
\centering
\scriptsize
\label{tab:hype6}
\begin{tabular}{cccc}
\hline
\textbf{Method} &
  \textbf{Encoder} &
  \textbf{\begin{tabular}[c]{@{}c@{}}mAP\\ (w/o RR)\end{tabular}} &
  \textbf{\begin{tabular}[c]{@{}c@{}}mAP\\ (with RR)\end{tabular}} \\ \hline
Baseline            & \multirow{3}{*}{ViT-L-14}    & 72.70          & -              \\
Clip-ReIdent        &                              & 96.90          & 98.20          \\
MPD (Clip-ReIdent) &                              & \textbf{97.02} & \textbf{98.30} \\ \hline
Clip-ReIdent        & \multirow{2}{*}{ResNet50x16} & 88.50          & 94.90          \\
MPD (Clip-ReIdent) &                              & \textbf{91.95} & \textbf{97.50} \\ \hline
\end{tabular}
\end{minipage}
\hfill
\begin{minipage}[b]{0.45\linewidth}
    \caption{Label efficiency shown in terms of frame counts.}
\vspace{-3mm}
\centering
\label{tab:frame-reident}
\scriptsize
\begin{tabular}{ccccc}
\hline
\textbf{epoch}      & \textbf{}       & \textbf{6 frames} & \textbf{10 frames} & \textbf{20 frames} \\ \hline
\multirow{2}{*}{8}  & w/o re-ranking  & 95                & 95.19              & 97.02              \\
                    & with re-ranking & 98                & 98.05              & 98.3               \\
\multirow{2}{*}{30} & w/o re-ranking  & 95.68             & 96.42              & 97.02              \\
                    & with re-ranking & 98.24             & 98.27              & 98.3               \\ \hline
\end{tabular}
\end{minipage}
\vspace{-8mm}
\end{table}
\subsubsection{Object Detection:} \label{subsec:od}
\begin{figure}[h]
    \vspace{-4mm}
    \centering
    \begin{minipage}[b]{0.45\linewidth}
        \centering
        \scriptsize
        \setlength{\tabcolsep}{2pt}
        \begin{tabular}{c|c|c}
        \hline
        Method & Original & MPD-OD (IoU=0.50:0.95/IoU=0.50) \\ \hline
        FasterRCNN\cite{ren2015faster}  & 37.10/55.80 & \textbf{40.00/61.10} \\ \hline
        FCOS\cite{9010746} & 38.60/57.40 & \textbf{40.20/60.30} \\ \hline
        \end{tabular}
        \vspace{-3mm}
        \captionof{table}{MPD-OD on COCO dataset.}
        \label{tab:coco_od_results}
    \end{minipage}\hfill
    \begin{minipage}[b]{0.45\linewidth}
        \centering
        \includegraphics[width=.6\linewidth]{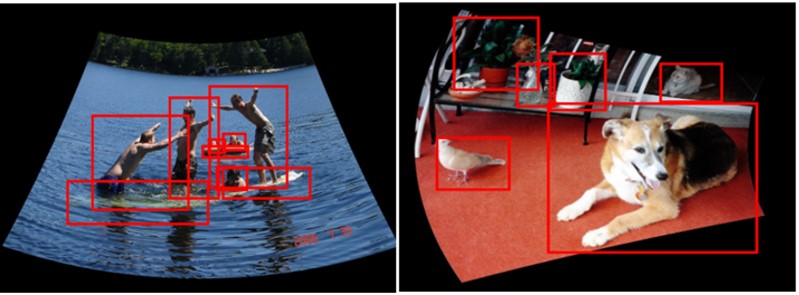} 
        \vspace{-3mm}
        \caption{MPD-OD examples}
        \label{fig:coco-pd}
    \end{minipage}
    \vspace{-8mm}
\end{figure}

We applied MPD method for the object detection (MPD-OD) by transforming image and bounding boxes as augmentation to mimic PD (fig. \ref{fig:coco-pd}). We have trained FasterRCNN \cite{ren2015faster} and FCOS \cite{9010746} with MPD-OD on COCO \cite{lin2014microsoft}. Applying MPD-OD to  50\%  probability (P=0.5), we achieved 3\% and 1.6\% gain over original models (Table \ref{tab:coco_od_results}). Ablations are in suppl. material (Table \ref{tab:coco_od_results_ablation} and Fig. \ref{fig:coco-pd-detailed} in sec. \ref{subsec:od}).
\section{Conclusion}
We address perspective distortion for robust computer vision applications. The proposed MPD method models real-world distortions by applying a family of Möbius transforms on existing data without estimating camera parameters or relying on actual distorted data.
We develop a perspectively-distorted ImageNet-PD dataset to benchmark the robustness of the computer vision models and reveal that existing models lack robustness. Further, MPD-incorporated models improve performance on existing benchmarks ImageNet-E and ImageNet-X and significantly enhance results on the ImageNet-PD while maintaining consistent performance on standard data. 
Also, MPD incorporated in practical applications such as crowd counting, fisheye image recognition, person re-identification, and object detection improved performance. We believe that our work on mitigating PD could inspire future work for other CV challenges.
\noindent\textbf{Acknowledgment} The authors thank Sumit Rakesh, Luleå University of Technology, for his support with the Lotty Bruzelius cluster. We also thank the National Supercomputer Centre at Linköping University for the Berzelius supercomputing, supported by the Knut and Alice Wallenberg Foundation.

\newpage

%% file: sec/suppl.tex
\begin{center}
\noindent\textbf{\Large Möbius Transform for Mitigating Perspective Distortions in
Representation Learning\\Supplementary Material}    
\end{center}

\section{Non-linearity and Conformality of Möbius} \label{sec:sup_math}
\noindent\textbf{Non-linearity}
Given Möbius transformations for variables \( z_1 \) and \( z_2 \), Möbius transformation of their sum \( z_1 + z_2 \):

\begin{equation}
    \Phi(z_1) = \frac{a z_1 + b}{c z_1 + d},~~~
    \Phi(z_2) = \frac{a z_2 + b}{c z_2 + d}
\end{equation}
\vspace{-2mm}
\begin{equation}
    \Phi(z_1 + z_2) = \frac{a (z_1 + z_2) + b}{c (z_1 + z_2) + d} = 
    \frac{a z_1 + a z_2 + b}{c z_1 + c z_2 + d}
\end{equation}
\vspace{-2mm}
\begin{equation}
    \Phi(z_1 + z_2) \neq \Phi(z_1) + \Phi(z_2)
    \label{eq:non-linear}
\end{equation}
However, Eq. (\ref{eq:non-linear}) is not equal to the sum of the individual transformations ($\Phi(z_1) + \Phi(z_2)$); therefore, \textit{non-linear}.


\noindent\textbf{Conformality}
Conformal transformations preserve angles between curves. The angle between two curves at a point is given by the \textit{argument of the complex ratio}\cite{howie2003complex} of their derivatives. Let $f(z)$ and $g(z)$ represent curves in complex space. If transformation preserves this ratio, it preserves angles.

Möbius transformation \(\Phi\) of $f(z)$ and $g(z)$ is given as:
\vspace{-2mm}
\begin{equation}
\Phi(f(z)) = \frac{af(z) + b}{cf(z) + d},~~~
\Phi(g(z)) = \frac{ag(z) + b}{cg(z) + d}
\end{equation}

To find the derivatives of the transformed functions, we apply the chain rule:
\vspace{-2mm}
\begin{equation}
\begin{split}
\Phi'(f(z))=\frac{d}{dz}(\Phi(f(z))) = \frac{d}{dz}\left(\frac{af(z) + b}{cf(z) + d}\right)\\
\Phi'(g(z))=\frac{d}{dz}(\Phi(g(z))) = \frac{d}{dz}\left(\frac{ag(z) + b}{cg(z) + d}\right)
\end{split}
\end{equation}

The angle between the original and transformed curves is given by the argument of the ratio\cite{howie2003complex}:
\vspace{-2mm}
\begin{equation}
\begin{split}
\text{arg}((f(z))/ g(z)))\equiv \newline
\text{arg}\left(\frac{d}{dz}(f(z))\right) 
- \text{arg}\left(\frac{d}{dz}(g(z))\right)
\end{split}
\end{equation}
\vspace{-3mm}
\begin{equation}
\begin{split}
\text{arg}(\Phi'(f(z))/ \Phi'(g(z)))\equiv \\
\vspace{-2mm}
\text{arg}\left(\frac{d}{dz}(\Phi(f(z)))\right) 
- \text{arg}\left(\frac{d}{dz}(\Phi(g(z)))\right)
\end{split}
\end{equation}

Eq. \ref{eq:conform} indicates conformality\cite{olsen2010geometry}.
\vspace{-2mm}
\begin{equation}
\text{arg}((f(z))/ g(z))) = \text{arg}(\Phi'(f(z))/ \Phi'(g(z)))
\label{eq:conform}
\end{equation}

\section{Theorem: MPD complies semantic-preserving property in pixel space by approximating conformality} \label{sec:sup_theorem}

\noindent\textbf{Proof:} Consider \( I: D \rightarrow \mathbb{C} \) as the discrete representation of an image over the domain \( D \), corresponding to the pixel grid, and let \( \Phi: \mathbb{C} \rightarrow \mathbb{C} \) be the continuous Möbius transformation given by \( \Phi(z) = \frac{az + b}{cz + d} \). MPD applies discrete transformation \( T \) to each point \(p\) in \( D \), which approximates \( \Phi \). The transformation \( T \) is constructed to preserve the angles between points in \( D \) as follows:

\begin{equation}
T(p) = \begin{cases}
\frac{a \cdot I(p) + b}{c \cdot I(p) + d} & \text{, }  c \cdot I(p) + d \neq 0, \\
\end{cases}
\end{equation}

The preservation of angles in the discrete domain is approximated by considering the discrete differential \( \Delta I \) at a point \( p \) in \( D \), defined as the vector of differences between \( I(p) \) and \( I(q) \) evaluated at neighboring points. For any two adjacent points \( p, q \in D \), the discrete differential \( \Delta I(p) \) and \( \Delta I(q) \) form vectors that subtend an angle \( \theta \) in the original pixel space. The transformation \( T \) aims to approximate the preservation of \( \theta \) such that the angle between \( \Delta T(p) \) and \( \Delta T(q) \), denoted as \( \tilde{\theta} \), is close to \( \theta \) in the transformed image space. This is expressed as \( \Delta T(p) \) and \( \Delta T(q) \) approximating the original vectors in a manner that the cosine of the angles is nearly preserved:
\begin{equation}
\cos(\theta) = \frac{\Delta I(p) \cdot \Delta I(q)}{\|\Delta I(p)\|\|\Delta I(q)\|} \approx \frac{\Delta T(p) \cdot \Delta T(q)}{\|\Delta T(p)\|\|\Delta T(q)\|}.    
\end{equation}
Since \( T \) approximates \( \Phi \), and \( \Phi \) is conformal, \( T \) aims to preserve the cosines of angles between points in \( D \), thus achieving an approximation of conformal mapping in the discrete pixel space.
MPD utilizes this property of \( T \) to maintain the visual integrity of the original image \( I \) in the transformed image \( I' \), where \( I'(p) = T(p) \) for all \( p \in D \).

\section{Perspectively distorted views synthesis through MPD}

\begin{figure}[h]
  \centering
  \includegraphics[width=.8\textwidth]{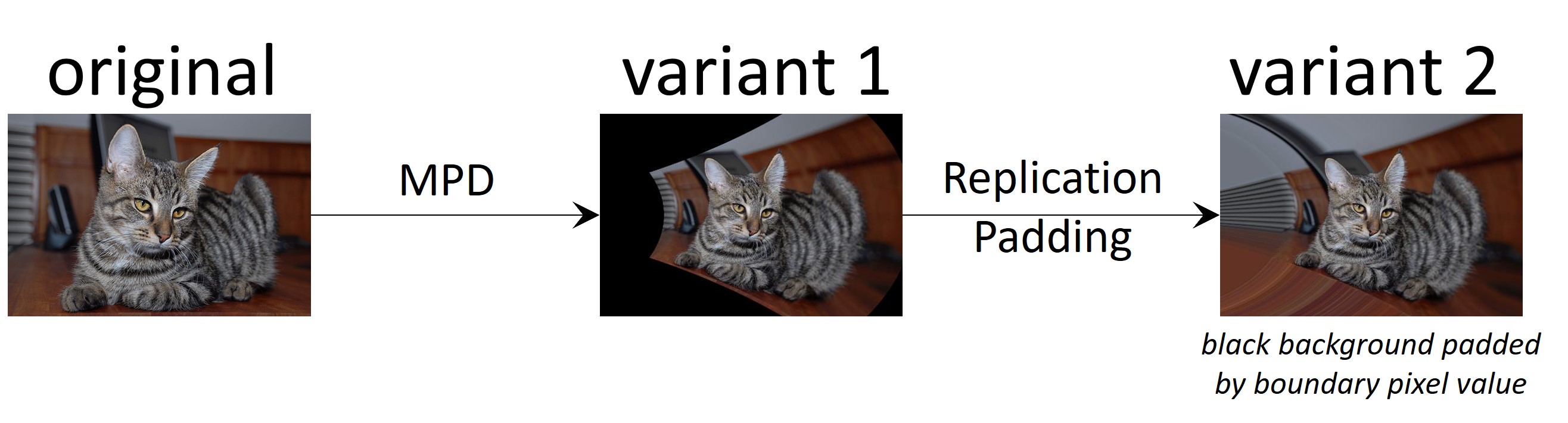}
  \vspace{-4mm}
  \caption{MPD transformed images - (black background) variant 1 and (integrated padding background) variant 2}
  \label{fig:padding}
\end{figure}

MPD distorts the original image to synthesize perspective distortion based on steps shown in Algorithm 1 in main paper, resulting in a distorted image with a black background (\textit{variant 1}, Fig. \ref{fig:padding}). Optionally, to remove the black background, replication padding is used to obtain integrated padding variant (variant 2, Fig. \ref{fig:padding}) of transformed image in step 9 of Algorithm 1.

\begin{figure}
    \centering
    \includegraphics[width=1\columnwidth]{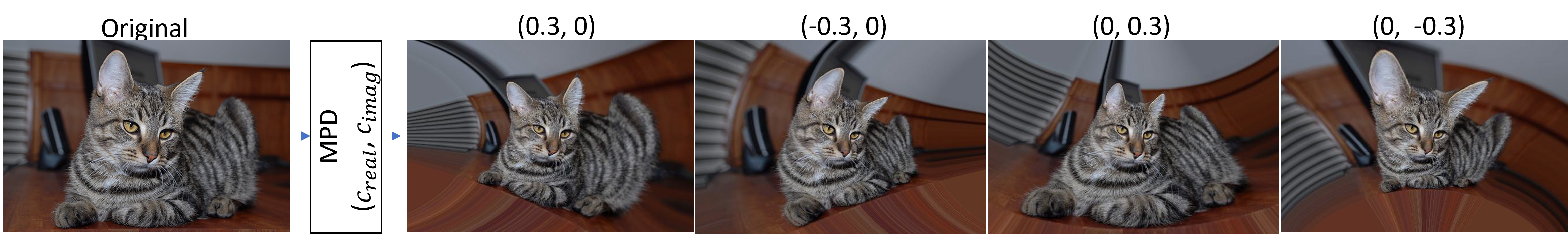}
    \caption{MPD IB (integrated padding background) synthesizes perspective distortion with different orientations corresponding to the parameters. It replaces the black background with boundary foreground pixel values of the transformed image.} 
    \label{fig:mpd_ib_suppl}
\end{figure}

In connection to Fig. \ref{fig:shematic}, we additionally show the integrated padding background variants of different orientations views synthesizing PD in Fig. \ref{fig:mpd_ib_suppl}. 

\begin{figure}
    \centering
    (a) \includegraphics[width=1\columnwidth]{figs/left_view.png}
    (b) \includegraphics[width=1\columnwidth]{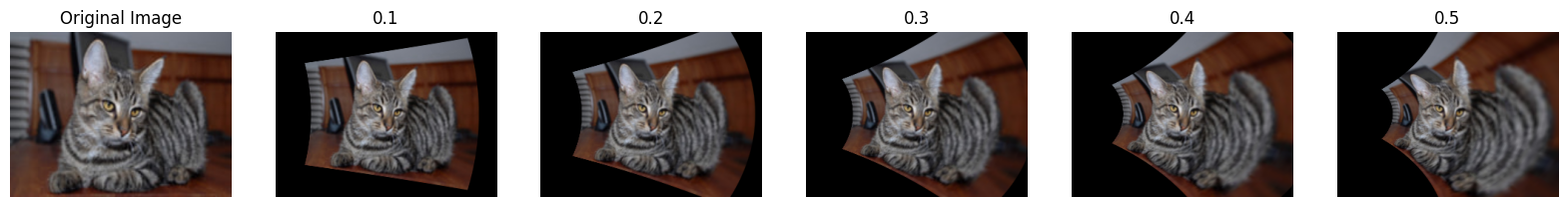}
    (c) \includegraphics[width=1\columnwidth]{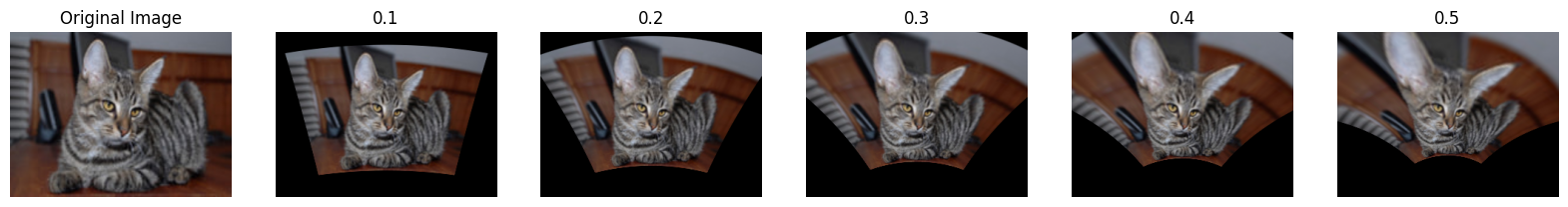}
    (d) \includegraphics[width=1\columnwidth]{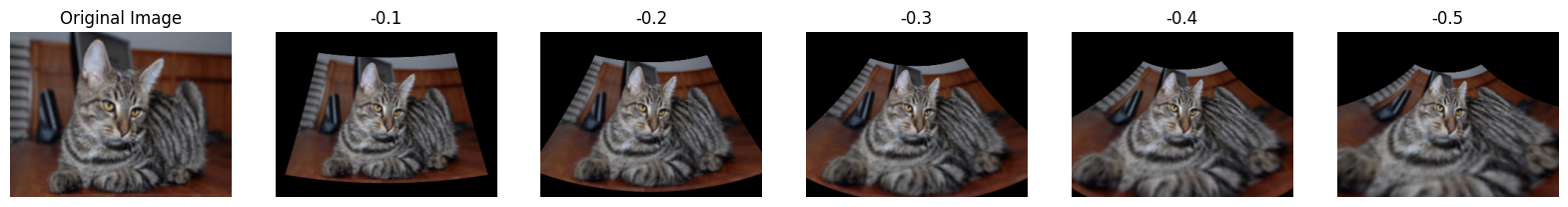}
    (e) \includegraphics[width=1\columnwidth]{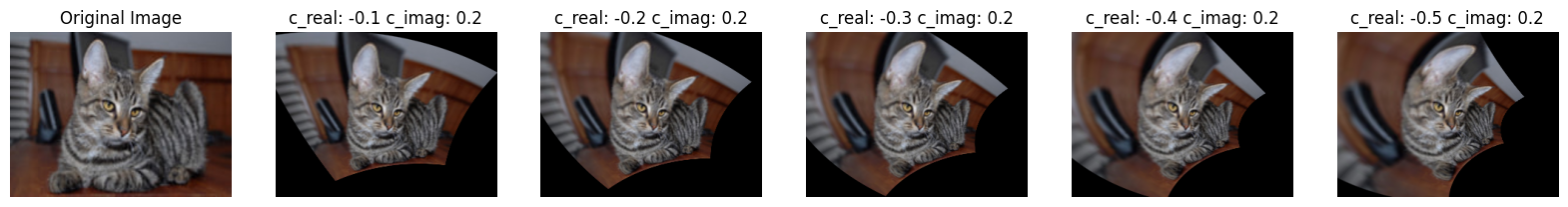}
    (f) \includegraphics[width=1\columnwidth]{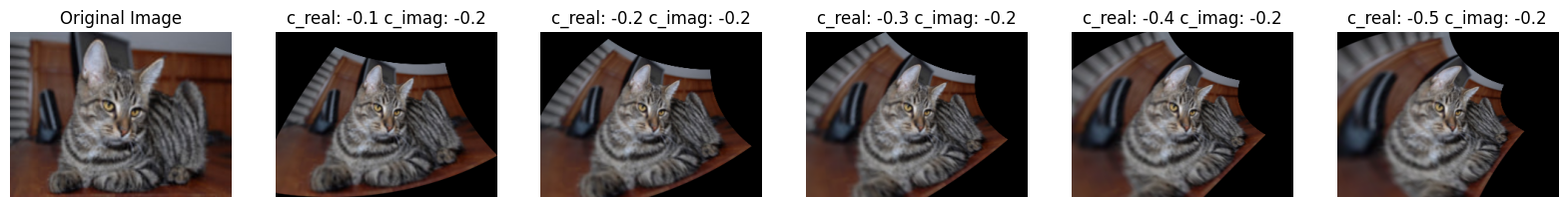}
    (g) \includegraphics[width=1\columnwidth]{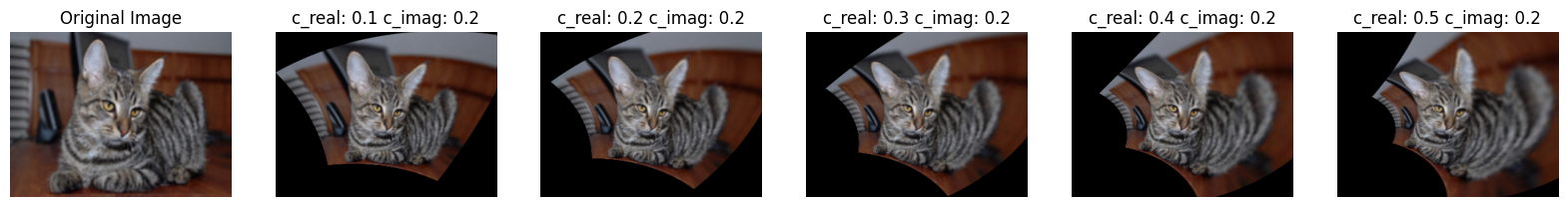}
    (h) \includegraphics[width=1\columnwidth]{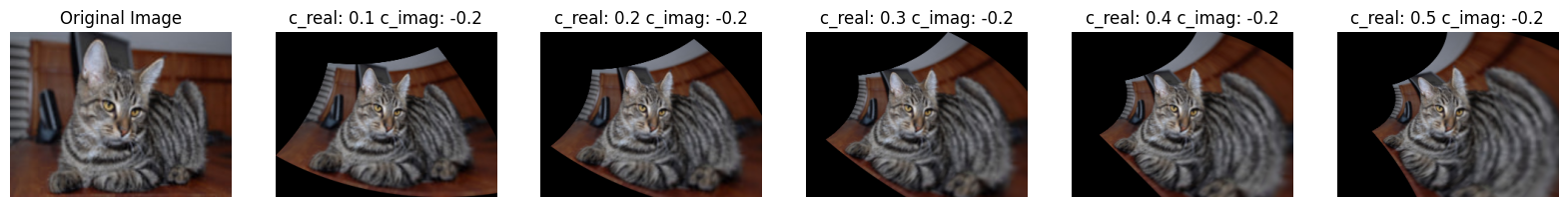}    
    \caption{Demonstrate controlled scaling of distortion in MPD-transformed images for different views. a) distorted left view by varying c\_real from $-0.1$ to $-0.5$ b) distorted right view by varying c\_real from $+0.1$ to $+0.5$ c) distorted top view by varying c\_imag from $+0.1$ to $+0.5$ d) distorted bottom view by varying c\_imag from $-0.1$ to $-0.5$ e) distorted left-top view by varying c\_real from $-0.1$ to $-0.5$ with fixed c\_imag to $+0.2$ f) distorted left-bottom view by varying c\_real from $-0.1$ to $-0.5$ with fixed c\_imag to $-0.2$ g) distorted right-top view by varying c\_real from $+0.1$ to $+0.5$ with fixed c\_imag to $+0.2$ h) distorted right-bottom view by varying c\_real from $+0.1$ to $+0.5$ with fixed c\_imag to $-0.2$. Current work experiments only use left, right, top, and bottom view synthesis.} 
    \label{fig:intensity_suppl}
\end{figure}

Fig. \ref{fig:intensity_suppl} Demonstrate controlled scaling of distortion in MPD-transformed image for different views in detail. It showcases the capabilities of MPD to mimic real-world distortions synthetically, which are otherwise challenging to capture and train the models.

\section{Exploration of parameters $a$, $b$, and $d$}
\begin{figure}
    \centering
    (a) \includegraphics[width=1\columnwidth]{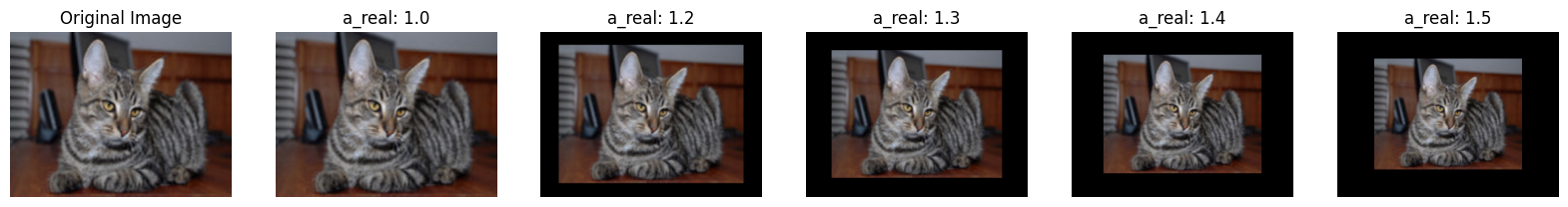}
    (b) \includegraphics[width=1\columnwidth]{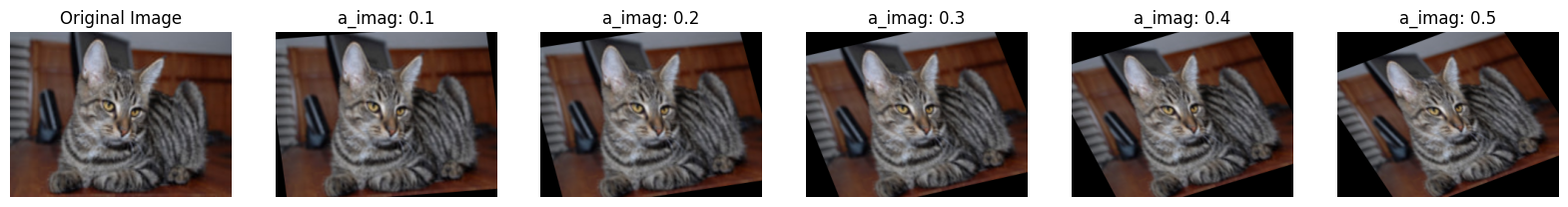}
    (c) \includegraphics[width=1\columnwidth]{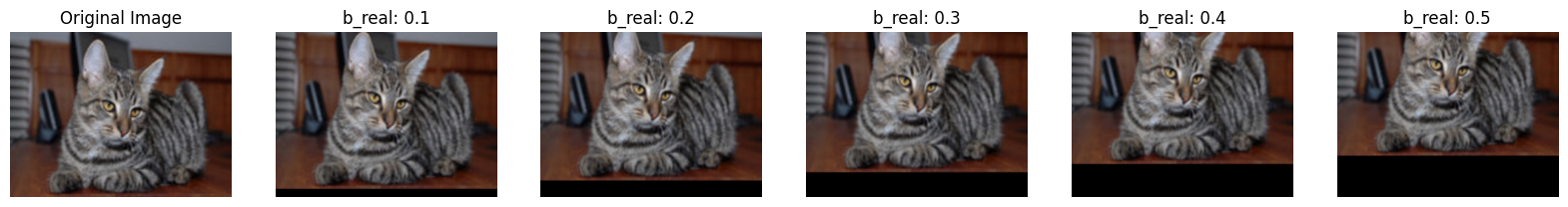}
    (d) \includegraphics[width=1\columnwidth]{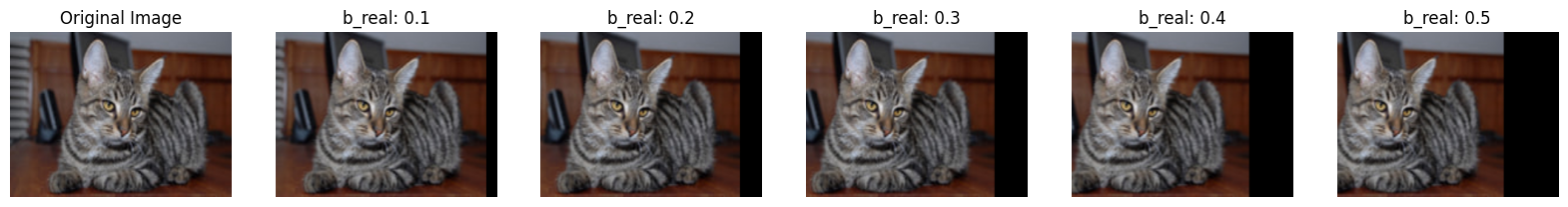}
    (e) \includegraphics[width=1\columnwidth]{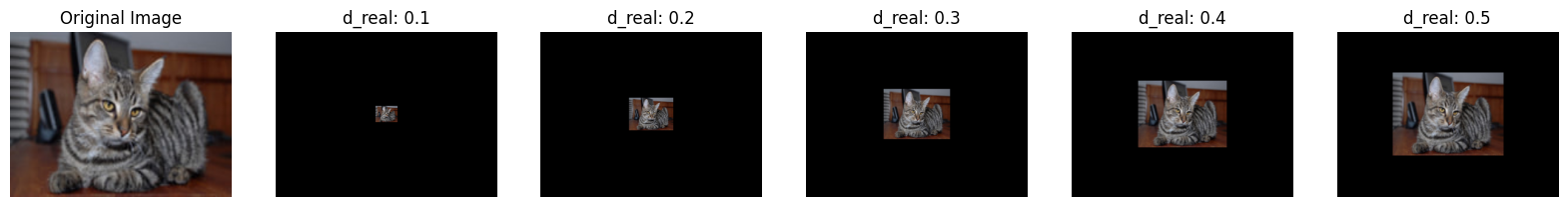}
    (f) \includegraphics[width=1\columnwidth]{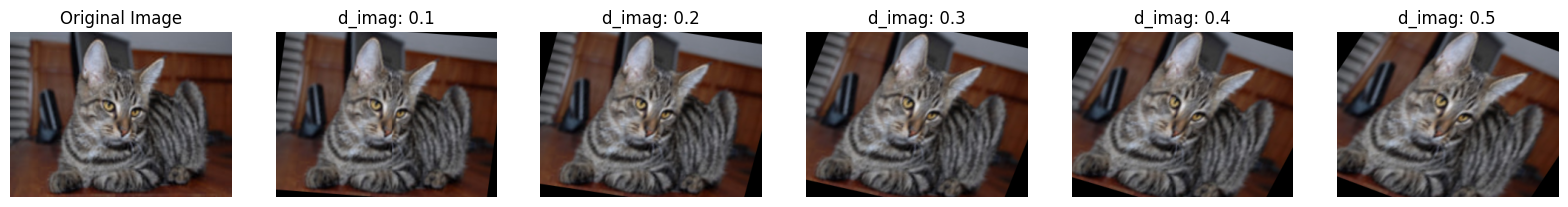}
    \caption{Visualizing the effect of parameters $a, b,$ and $d$, representing subsets of affine transformations. a) imitating resizing by controlling parameter (increasing value decreases size) $a_{real}$ $1.0$ to $1.5$ b) rotating (anti-clockwise) by controlling parameter $a_{imag}$ $0.1$ to $0.5$ c) imitating vertical translation by controlling parameter $b_{real}$ $0.1$ to $0.5$ d) imitating horizontal translation by controlling parameter $b_{imag}$ $0.1$ to $0.5$ e) imitating resizing  by controlling parameter (increasing value increase size) $d_{real}$ $0.1$ to $0.5$ f) imitating rotation  by controlling parameter (clockwise) $d_{real}$ $0.1$ to $0.5$} 
    \label{fig:a_b_d_suppl}
\end{figure}

Fig. \ref{fig:a_b_d_suppl} shows the effect of parameters $a, b,$ and $d$, representing subsets of affine transformations. 

\section{MPD in self-supervised learning methods}
\begin{figure}[h]
  \centering
  \includegraphics[width=\columnwidth]{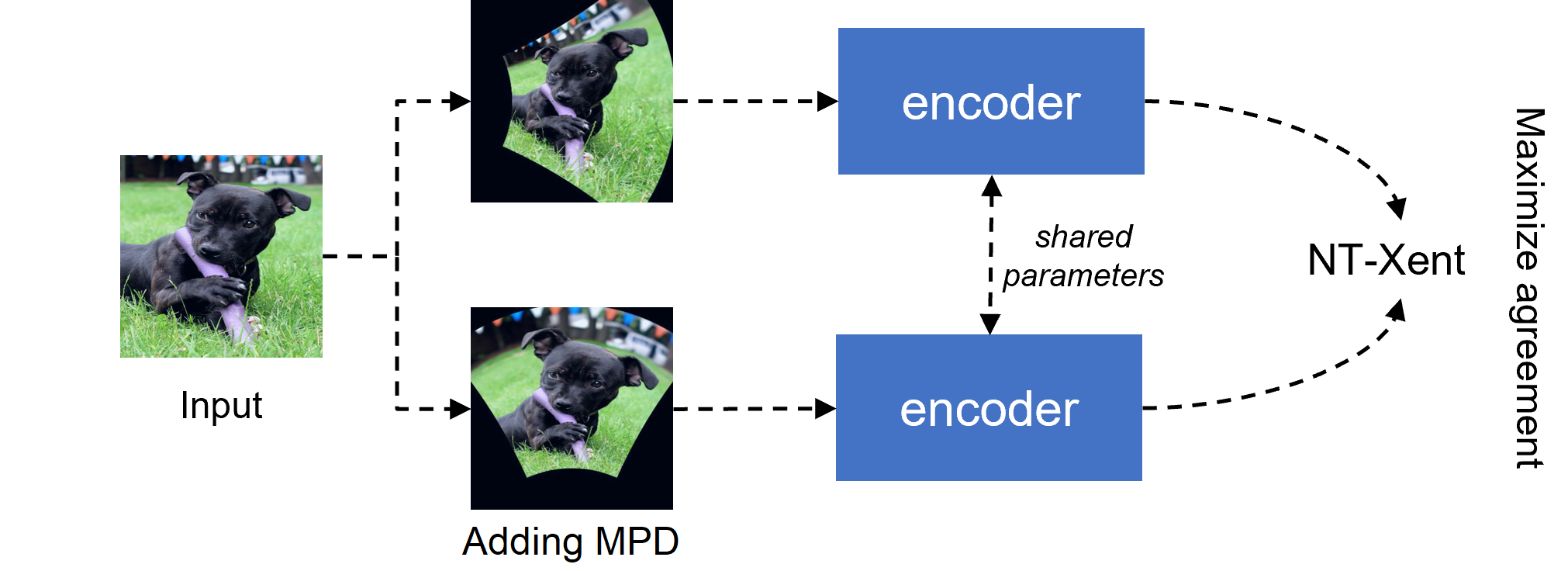}
  \caption{ssl:MPD - MPD integrated in SimCLR \cite{chen2020simple} contrastive learning self-supervised method. MPD introduces perspective distortion by transforming  the views of the input image. Standard SimCLR and ssl:DINO-MPD both are pretrained for 100 epochs with a batch size of 512 and linear evaluation for 100 epochs with a batch size of 256. ResNet50 is backbone.} 
  \label{fig:simclr_MPD}
\end{figure}

\begin{figure}[h]
  \centering
  \includegraphics[width=\columnwidth]{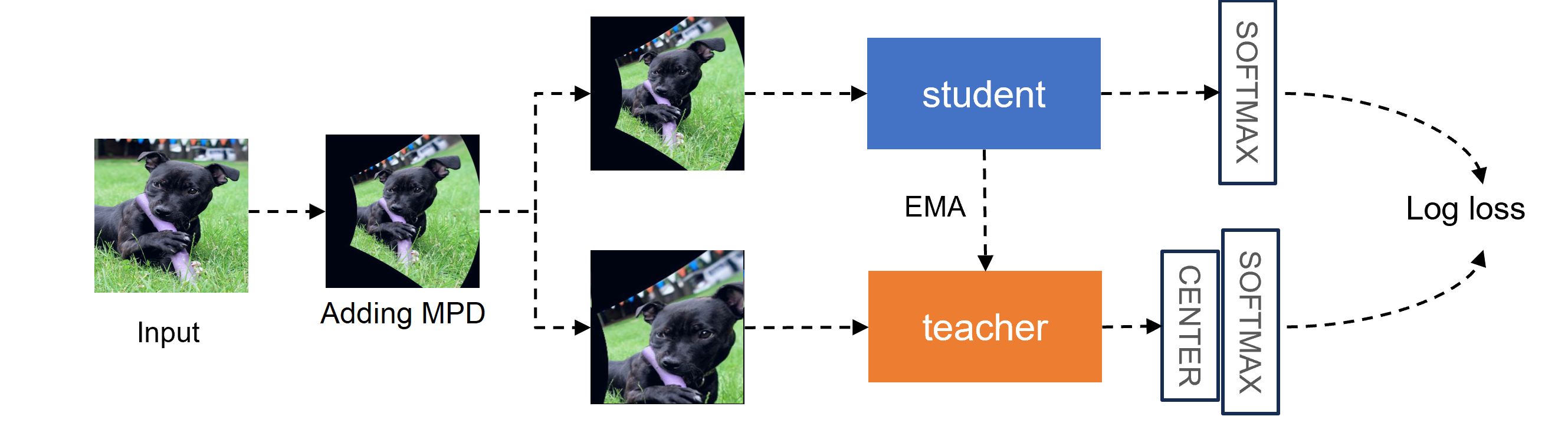}
  \caption{ssl:DINO-MPD - MPD integrated in DINO \cite{caron2021emerging} knowledge distillation based self-supervised method. MPD introduces perspective distortion by transforming the input image before the creation of views. Standard DINO and ssl:DINO-MPD both are pretrained for 100 epochs with a batch size of 512 and linear evaluation for 100 epochs with a batch size of 256. ViT-small is backbone.} 
  \label{fig:DINO_MPD}
\end{figure}

We present a comprehensive investigation into the application of MPD (Geometry-Exploring Encoded Transformation) within self-supervised learning (SSL) approaches, with a primary focus on its integration with the contrastive learning method SimCLR \cite{chen2020simple}, shown in Fig. \ref{fig:simclr_MPD}. Our core objective is to assess MPD's effectiveness in enhancing SSL's representation learning capacities by adapting to perspective variations. Section \ref{ob3} in the main paper explains the detailed results of MPD incorporation with simCLR contrastive learning to mitigate perspective distortion. Further Table linear evaluation results are included in Section \ref{ob3_c}. Furthermore, we extend our exploration to evaluate MPD's compatibility and performance in conjunction with another SSL method, DINO \cite{caron2021emerging}, which is grounded in knowledge distillation-based self-supervised approaches. Results on MPD incorporation in DINO (Fig. \ref{fig:DINO_MPD}) are presented in Table \ref{tab:Hype3-DINO-MPD-LE} and Fig. \ref{fig:dino_attn_example} in Section \ref{ob3_c}.
\noindent\textbf{Probability of 0.8}: Incorporating  MPD with a probability of 0.8 within the SimCLR contrastive learning shows a well-balanced strategy. This allocation ensures that each of the four perspective views—left, right, top, and bottom—receives an equitable opportunity approximately 20\% of the time, fostering diversity and divergence within the training data. This design navigates the intricate terrain of the diversity-consistency trade-off with precision, enabling our model to gain insights from multiple angles while maintaining the stability and consistency garnered from the original, unaltered views. The empirical validation of our approach underscores its efficacy, as it not only enhances the model's ability to recognize objects and scenes from varying perspectives but also fortifies its robustness against real-world perspective distortions. The DINO method applies MPD to the original input image before creating local and global views (as defined in the DINO method), so the MPD probability is kept at 0.4. All the results described with standard deviation are repeated three times.
\section{Continuation (Sec. \ref{ob1}): Existing models' robustness evaluation } \label{ob1_c}
Extended evaluations on the ImageNet-PD benchmark dataset, detailed in Tables \ref{tab:hype1A} and \ref{tab:hype1A_IB}, reveal how different architectures respond to perspective distortions, extending the main paper's insights (Fig. \ref{fig:H1-top1-top5}). Models like ResNet50 and VIT-L-16 show varied robustness to distortions such as Left-view (PD-L and PD-LI), with ResNet50 maintaining 63.37\% Top-1 accuracy for PD-L compared to 61.15\% for PD-B, and VIT-L-16 at 74.34\% for PD-LI versus 71.73\% for PD-BI. With their self-attention mechanisms, Vision Transformers demonstrate a smaller performance drop than traditional CNNs under perspective distortions, indicating their potential to maintain representational robustness. As discussed in our main paper, this analysis underscores the need for adaptive, perspective-aware models like MPD in real-world applications.
\begin{table*}[h]
\caption{Detailed results of ImageNet trained models with standard CNN and transformer architectures; evaluated on original validation set and ImageNet-PD subsets with black background. Results in continuation of Fig. \ref{fig:H1-top1-top5} (first plot).}
\centering
\scriptsize
\label{tab:hype1A}
\tiny
\begin{tabular}{c|cccccccccc|cc}
\hline
\multirow{2}{*}{Model} & \multicolumn{2}{c}{Original   Validation Set} & \multicolumn{2}{c}{\begin{tabular}[c]{@{}c@{}}Perspectively   Distorted\\      Top-view (PD-T)\end{tabular}} & \multicolumn{2}{c}{\begin{tabular}[c]{@{}c@{}}Perspectively   Distorted\\      Bottom-view (PD-B)\end{tabular}} & \multicolumn{2}{c}{\begin{tabular}[c]{@{}c@{}}Perspectively   Distorted\\      Left-view (PD-L)\end{tabular}} & \multicolumn{2}{c|}{\begin{tabular}[c]{@{}c@{}}Perspectively   Distorted\\      Right-view (PD-R)\end{tabular}} & \multicolumn{2}{c}{PD Average} \\ \cline{2-13} 
                       & Top1                  & Top5                  & Top1                                                  & Top5                                                 & Top1                                                   & Top5                                                   & Top1                                                  & Top5                                                  & Top1                                                   & Top5                                                   & Top1         & Top5         \\ \hline
AlexNet                & 56.52                 & 79.07                 & 36.74                                                 & 60.02                                                & 37.08                                                  & 61.21                                                  & 37.87                                                 & 61.87                                                 & 37.33                                                  & 61.06                                                  & 37.26        & 61.04        \\
EfficientNet-b2        & 80.61                 & 95.31                 & 67.14                                                 & 86.49                                                & 64.97                                                  & 84.84                                                  & 67.60                                                 & 87.11                                                 & 67.15                                                  & 86.77                                                  & 66.71        & 86.30        \\
GoogleNet              & 69.78                 & 89.53                 & 54.63                                                 & 76.97                                                & 55.29                                                  & 78.45                                                  & 55.39                                                 & 78.24                                                 & 54.68                                                  & 77.72                                                  & 55.00        & 77.85        \\
Inception-v3           & 77.29                 & 93.45                 & 55.09                                                 & 76.76                                                & 52.79                                                  & 74.83                                                  & 54.45                                                 & 76.19                                                 & 54.86                                                  & 76.58                                                  & 54.30        & 76.09        \\
MobileNet-v2           & 71.88                 & 90.29                 & 57.60                                                 & 79.78                                                & 56.26                                                  & 78.73                                                  & 57.40                                                 & 79.91                                                 & 56.87                                                  & 79.54                                                  & 57.03        & 79.49        \\
ResNet101              & 77.37                 & 93.55                 & 65.86                                                 & 85.58                                                & 65.17                                                  & 85.29                                                  & 65.62                                                 & 85.46                                                 & 65.40                                                  & 85.49                                                  & 65.51        & 85.46        \\
ResNet50               & 76.13                 & 92.86                 & 63.37                                                 & 83.61                                                & 61.15                                                  & 81.86                                                  & 65.20                                                 & 85.13                                                 & 65.84                                                  & 85.64                                                  & 63.89        & 84.06        \\
ResNet50 (SSL)         & 76.36                 & 92.99                 & 63.76                                                 & 84.12                                                & 61.71                                                  & 82.91                                                  & 65.20                                                 & 85.13                                                 & 65.96                                                  & 86.11                                                  & 64.16        & 84.57        \\
ResNet34               & 73.31                 & 91.42                 & 59.23                                                 & 80.84                                                & 58.65                                                  & 80.60                                                  & 59.72                                                 & 81.25                                                 & 58.79                                                  & 80.76                                                  & 59.10        & 80.86        \\
ResNet18               & 69.76                 & 89.08                 & 56.00                                                 & 78.52                                                & 55.07                                                  & 77.90                                                  & 55.68                                                 & 78.73                                                 & 55.27                                                  & 78.01                                                  & 55.50        & 78.29        \\
VGG19                  & 72.38                 & 90.88                 & 58.50                                                 & 80.36                                                & 58.09                                                  & 80.36                                                  & 58.00                                                 & 80.13                                                 & 57.64                                                  & 79.67                                                  & 58.06        & 80.13        \\
VGG16                  & 71.59                 & 90.38                 & 57.94                                                 & 80.54                                                & 57.32                                                  & 79.89                                                  & 57.47                                                 & 79.91                                                 & 57.75                                                  & 80.15                                                  & 57.62        & 80.12        \\
VIT-B-16               & 81.07                 & 95.32                 & 73.94                                                 & 90.87                                                & 73.63                                                  & 90.73                                                  & 74.50                                                 & 91.31                                                 & 74.30                                                  & 91.06                                                  & 74.09        & 90.99        \\
VIT-L-16               & 79.66                 & 94.64                 & 74.05                                                 & 91.06                                                & 74.30                                                  & 90.97                                                  & 74.34                                                 & 91.08                                                 & 73.93                                                  & 90.91                                                  & 74.15        & 91.00        \\ \hline
\end{tabular}
\end{table*}
\begin{table*}[]
\caption{Detailed results of ImageNet trained models with standard CNN and transformer architectures; evaluated on original validation set and ImageNet-PD subsets with integrated padding background. Results in continuation of Fig. \ref{fig:H1-top1-top5} (second plot).}
\scriptsize
\centering
\label{tab:hype1A_IB}
\tiny
\begin{tabular}{c|cccccccccc|cc}
\hline
\multirow{2}{*}{Model} & \multicolumn{2}{c}{Original   Validation Set} & \multicolumn{2}{c}{\begin{tabular}[c]{@{}c@{}}Perspectively   Distorted\\      Top-view (PD-TI)\end{tabular}} & \multicolumn{2}{c}{\begin{tabular}[c]{@{}c@{}}Perspectively   Distorted\\      Bottom-view (PD-BI)\end{tabular}} & \multicolumn{2}{c}{\begin{tabular}[c]{@{}c@{}}Perspectively   Distorted\\      Left-view (PD-LI)\end{tabular}} & \multicolumn{2}{c|}{\begin{tabular}[c]{@{}c@{}}Perspectively   Distorted\\      Right-view (PD-RI)\end{tabular}} & \multicolumn{2}{c}{PD Average} \\ \cline{2-13} 
                       & Top1                  & Top5                  & Top1                                                  & Top5                                                  & Top1                                                    & Top5                                                   & Top1                                                   & Top5                                                  & Top1                                                    & Top5                                                   & Top1           & Top5          \\ \hline
AlexNet                & 56.52                 & 79.07                 & 40.73                                                 & 64.47                                                 & 41.09                                                   & 65.36                                                  & 42.53                                                  & 66.32                                                 & 41.13                                                   & 65.08                                                  & 41.37          & 65.31         \\
EfficientNet-b2        & 80.61                 & 95.31                 & 67.83                                                 & 87.28                                                 & 66.29                                                   & 85.95                                                  & 68.16                                                  & 87.43                                                 & 68.12                                                   & 87.48                                                  & 67.60          & 87.04         \\
GoogleNet              & 69.78                 & 89.53                 & 54.68                                                 & 77.22                                                 & 56.84                                                   & 79.51                                                  & 57.24                                                  & 79.68                                                 & 56.01                                                   & 78.82                                                  & 56.19          & 78.81         \\
Inception-v3           & 77.29                 & 93.45                 & 56.77                                                 & 78.36                                                 & 54.32                                                   & 76.23                                                  & 55.94                                                  & 77.56                                                 & 56.81                                                   & 78.55                                                  & 55.96          & 77.68         \\
MobileNet-v2           & 71.88                 & 90.29                 & 57.34                                                 & 79.68                                                 & 57.84                                                   & 80.20                                                  & 58.18                                                  & 80.42                                                 & 57.26                                                   & 79.96                                                  & 57.66          & 80.07         \\
ResNet101              & 77.37                 & 93.55                 & 64.30                                                 & 84.47                                                 & 65.57                                                   & 85.64                                                  & 66.55                                                  & 86.14                                                 & 65.69                                                   & 85.70                                                  & 65.53          & 85.49         \\
ResNet50               & 76.13                 & 92.86                 & 62.26                                                 & 83.07                                                 & 63.39                                                   & 83.91                                                  & 63.81                                                  & 84.17                                                 & 63.57                                                   & 84.16                                                  & 63.26          & 83.83         \\
ResNet50 (SSL)         & 76.24                 & 92.90                 & 61.73                                                 & 82.50                                                 & 61.91                                                   & 83.07                                                  & 65.58                                                  & 85.74                                                 & 65.79                                                   & 86.11                                                  & 63.75          & 84.36         \\
ResNet34               & 73.31                 & 91.42                 & 58.29                                                 & 80.11                                                 & 59.61                                                   & 81.38                                                  & 60.65                                                  & 82.22                                                 & 59.20                                                   & 81.06                                                  & 59.44          & 81.19         \\
ResNet18               & 69.76                 & 89.08                 & 55.85                                                 & 78.58                                                 & 56.27                                                   & 79.13                                                  & 56.87                                                  & 79.83                                                 & 56.05                                                   & 78.57                                                  & 56.26          & 79.03         \\
VGG19                  & 72.38                 & 90.88                 & 57.68                                                 & 80.06                                                 & 59.28                                                   & 81.43                                                  & 58.05                                                  & 80.41                                                 & 57.58                                                   & 79.73                                                  & 58.15          & 80.41         \\
VGG16                  & 71.59                 & 90.38                 & 56.26                                                 & 78.97                                                 & 58.58                                                   & 80.94                                                  & 57.96                                                  & 80.37                                                 & 57.41                                                   & 80.07                                                  & 57.55          & 80.09         \\
VIT-B-16               & 81.07                 & 95.32                 & 70.49                                                 & 88.58                                                 & 71.89                                                   & 89.49                                                  & 72.45                                                  & 89.93                                                 & 71.51                                                   & 89.27                                                  & 71.59          & 89.32         \\
VIT-L-16               & 79.66                 & 94.64                 & 70.71                                                 & 88.73                                                 & 71.73                                                   & 89.41                                                  & 71.87                                                  & 89.49                                                 & 71.25                                                   & 89.16                                                  & 71.39          & 89.20         \\ \hline
\end{tabular}
\end{table*}
\section{Continuation (section \ref{ob2}): MPD's effects on supervised learning} \label{ob2_c}
In the extended analysis presented in Tables \ref{tab:Hype2-3main-standard_extended} and \ref{tab:Hype2-3main-IB-extended}, we observe a shift in the performance of MPD and MPD IB when applied at a probability of 1.0, approximately following the trends reported in the main paper up to a probability of 0.8 (Fig. \ref{fig:hypo2_sup}). For MPD at P=1.0, a marginal decrease in Top-1 accuracy on the original ImageNet validation set (74.10\%\,$\pm$\,0.03) compared to P=0.8 (76.34\%\,$\pm$\,0.02), the robustness against perspective distortions on ImageNet-PD subsets is maintained, with accuracies between 71.57\%\,$\pm$\,0.04 to 72.05\%\,$\pm$\,0.04. This highlights a nuanced balance between handling distortions and achieving peak performance on standard datasets. Similarly, MPD IB at P=1.0 follows a comparable pattern. The Top-1 accuracy on the standard ImageNet set slightly reduces to 74.62\%\,$\pm$\,0.03 from 75.63\%\,$\pm$\,0.05 at P=0.8. However, its performance on the ImageNet-PD subsets, though marginally lower than MPD, remains consistent with Top-1 accuracies in the range of 68.23\%\,$\pm$\,0.04 to 70.30\%\,$\pm$\,0.03. These results emphasize the trade-offs involved in fully applying MPD and MPD IB and the importance of finding a balanced approach to optimize performance across both standard data and perspective distortion present in data.

\noindent\textbf{ImageNet-E:} In our extended analysis of MPD and MPD IB within a supervised approach (Table \ref{tab:tab:imagenet-e_detailed} and Fig. \ref{fig:hypo2_ImageNet-E}), we observe notable performance improvements in size change scenarios of ImageNet-E, indicative of enhanced robustness against perspective distortion. These models, specifically in size-related subsets (full, 0.10, 0.08, 0.05), show a reduced drop in Top-1 accuracy and higher absolute accuracy than standard ResNet50. However, in background change subsets, MPD and MPD IB's performance occasionally falls behind ResNet50, reflecting their focus on perspective distortions over background variations. This further aligns with the findings on MPD's capabilities in perspective-centric scenarios.

\noindent \textbf{ImageNet-X} Detailed analysis in Table \ref{tab:imagenet_x_detailed} extends the findings from the main paper (Fig. \ref{fig:hypo2_ImageNet-X}), showcasing the superior performance of supervised MPD models on ImageNet-X, especially in managing perspective-related factors. These models demonstrate a marked improvement in handling size variations ('larger' and 'smaller') and complex scenarios involving 'object blocking' and 'person blocking' compared to the standard ResNet50. The error ratio and absolute accuracy across these factors highlight the robustness of MPD in dealing with perspective challenges, reflecting the conclusion about MPD's enhanced learning capabilities in diverse visual conditions.
\begin{table*}[t]
\caption{Extended comparative analysis of MPD trained on supervised (\textit{supervised:MPD model}) and self-supervised (\textit{ssl:MPD model}) approaches and evaluated on original ImageNet and \textbf{ImageNet-PD subsets} (black background). The probability \(P\) of applying MPD in model fine-tuning is in the first column. Added results for probability values 0.0 and 1.0 in continuation to Table \ref{tab:Hype2-3main-standard}.}
\centering
\scriptsize
\label{tab:Hype2-3main-standard_extended}
\begin{tabular}{ccccccccccc}
\hline
\multicolumn{1}{c|}{\multirow{2}{*}{P}} & \multicolumn{2}{c|}{Original Validation Set} & \multicolumn{2}{c|}{\begin{tabular}[c]{@{}c@{}}Perspectively Distorted\\ Top-view (PD-T)\end{tabular}} & \multicolumn{2}{c|}{\begin{tabular}[c]{@{}c@{}}Perspectively Distorted\\ Bottom-view (PD-B)\end{tabular}} & \multicolumn{2}{c|}{\begin{tabular}[c]{@{}c@{}}Perspectively Distorted\\ Left-view (PD-L)\end{tabular}} & \multicolumn{2}{c}{\begin{tabular}[c]{@{}c@{}}Perspectively Distorted\\ Right-view (PD-R)\end{tabular}} \\ \cline{2-11} 
\multicolumn{1}{c|}{}                   & Top1       & \multicolumn{1}{c|}{Top 5}      & Top1                                    & \multicolumn{1}{c|}{Top 5}                                   & Top1                                      & \multicolumn{1}{c|}{Top 5}                                    & Top1                                     & \multicolumn{1}{c|}{Top 5}                                   & Top1                                               & Top 5                                              \\ \hline
\multicolumn{11}{c}{\textbf{Supervised training from scratch}}                                                                                                                                                                                                                                                                                                                                                                                                                                                                  \\ \hline
\multicolumn{1}{c|}{0.0}                & 76.13±0.04 & \multicolumn{1}{c|}{92.86±0.01} & 63.37±0.06                              & \multicolumn{1}{c|}{83.61±0.02}                              & 61.15±0.04                                & \multicolumn{1}{c|}{81.86±0.01}                               & 65.20±0.03                               & \multicolumn{1}{c|}{85.13±0.03}                              & 65.84±0.06                                         & 85.64±0.02                                         \\ \hline
\multicolumn{1}{c|}{0.2}                & 76.05±0.03 & \multicolumn{1}{c|}{92.99±0.02} & 72.48±0.02                              & \multicolumn{1}{c|}{91.02±0.01}                              & 72.12±0.02                                & \multicolumn{1}{c|}{91.08±0.02}                               & 72.57±0.03                               & \multicolumn{1}{c|}{91.13±0.03}                              & 72.80±0.05                                         & 91.35±0.02                                         \\
\multicolumn{1}{c|}{0.4}                & 76.17±0.04 & \multicolumn{1}{c|}{93.03±0.02} & 73.13±0.03                              & \multicolumn{1}{c|}{91.33±0.01}                              & 72.94±0.02                                & \multicolumn{1}{c|}{91.42±0.01}                               & 73.19±0.06                               & \multicolumn{1}{c|}{91.48±0.01}                              & 73.38±0.01                                         & 91.69±0.01                                         \\
\multicolumn{1}{c|}{0.6}                & 76.19±0.05 & \multicolumn{1}{c|}{93.14±0.03} & 73.23±0.02                              & \multicolumn{1}{c|}{91.46±0.03}                              & 73.01±0.05                                & \multicolumn{1}{c|}{91.34±0.02}                               & 73.47±0.02                               & \multicolumn{1}{c|}{91.66±0.01}                              & 73.54±0.06                                         & 91.61±0.02                                         \\
\multicolumn{1}{c|}{0.8}                & 76.34±0.02 & \multicolumn{1}{c|}{93.03±0.02} & 73.00±0.02                              & \multicolumn{1}{c|}{90.69±0.02}                              & 72.31±0.03                                & \multicolumn{1}{c|}{90.81±0.03}                               & 73.50±0.04                               & \multicolumn{1}{c|}{91.33±0.03}                              & 72.91±0.02                                         & 91.29±0.02                                         \\
\multicolumn{1}{l|}{1.0}                & 74.10±0.03 & \multicolumn{1}{c|}{91.50±0.03} & 71.57±0.04                              & \multicolumn{1}{c|}{90.31±0.03}                              & 71.65±0.03                                & \multicolumn{1}{c|}{90.44±0.02}                               & 71.88±0.04                               & \multicolumn{1}{c|}{90.62±0.03}                              & 72.05±0.04                                         & 90.60±0.03                                         \\ \hline
\multicolumn{11}{c}{\textbf{Self-supervised  pre-taining on contrastive learning integrating MPD (probability for pre-training=0.8)}}                                                                                                                                                                                                                                                                                                                                                                                                           \\ \hline
\multicolumn{1}{c|}{0.2}                & 76.37±0.05 & \multicolumn{1}{c|}{93.60±0.02} & 72.34±0.02                              & \multicolumn{1}{c|}{90.81±0.02}                              & 72.24±0.03                                & \multicolumn{1}{c|}{90.95±0.02}                               & 72.48±0.04                               & \multicolumn{1}{c|}{91.08±0.01}                              & 72.81±0.03                                         & 91.16±0.01                                         \\
\multicolumn{1}{c|}{0.4}                & 76.77±0.02 & \multicolumn{1}{c|}{93.40±0.01} & 73.23±0.03                              & \multicolumn{1}{c|}{91.39±0.01}                              & 73.06±0.04                                & \multicolumn{1}{c|}{91.54±0.02}                               & 73.55±0.03                               & \multicolumn{1}{c|}{91.53±0.03}                              & 73.54±0.05                                         & 91.65±0.03                                         \\
\multicolumn{1}{c|}{0.6}                & 76.14±0.04 & \multicolumn{1}{c|}{93.58±0.02} & 73.39±0.02                              & \multicolumn{1}{c|}{91.50±0.02}                              & 73.29±0.03                                & \multicolumn{1}{c|}{91.57±0.01}                               & 73.57±0.03                               & \multicolumn{1}{c|}{91.66±0.02}                              & 73.73±0.04                                         & 91.58±0.02                                         \\
\multicolumn{1}{c|}{0.8}                & 76.29±0.03 & \multicolumn{1}{c|}{92.78±0.02} & 73.61±0.04                              & \multicolumn{1}{c|}{91.49±0.03}                              & 73.27±0.05                                & \multicolumn{1}{c|}{91.48±0.03}                               & 73.60±0.06                               & \multicolumn{1}{c|}{91.69±0.01}                              & 73.81±0.02                                         & 91.71±0.02                                         \\
\multicolumn{1}{l|}{1.0}                & 74.13±0.04 & \multicolumn{1}{c|}{91.89±0.03} & 71.70±0.03                              & \multicolumn{1}{c|}{90.42±0.03}                              & 71.79±0.04                                & \multicolumn{1}{c|}{90.65±0.02}                               & 71.96±0.04                               & \multicolumn{1}{c|}{90.70±0.02}                              & 72.20±0.03                                         & 90.88±0.02                                         \\ \hline
\end{tabular}
\end{table*}

\begin{table*}[]
\caption{Extended comparative analysis of MPD IB trained on supervised (\textit{supervised:MPD IB model}) and self-supervised (\textit{ssl:MPD IB  model}) approaches and evaluated on original ImageNet and \textbf{ImageNet-PD subsets} (black background). The probability \(P\) of applying MPD BI in model fine-tuning is in the first column.}
\label{tab:Hype2-3main-IB-extended}
\scriptsize
\begin{tabular}{ccccccccccc}
\hline
\multicolumn{1}{c|}{\multirow{2}{*}{P}} & \multicolumn{2}{c|}{Original Validation Set} & \multicolumn{2}{c|}{\begin{tabular}[c]{@{}c@{}}Perspectively Distorted\\ Top-view (PD-TI)\end{tabular}} & \multicolumn{2}{c|}{\begin{tabular}[c]{@{}c@{}}Perspectively Distorted\\ Bottom-view (PD-BI)\end{tabular}} & \multicolumn{2}{c|}{\begin{tabular}[c]{@{}c@{}}Perspectively Distorted\\ Left-view (PD-LI)\end{tabular}} & \multicolumn{2}{c}{\begin{tabular}[c]{@{}c@{}}Perspectively Distorted\\ Right-view (PD-RI)\end{tabular}} \\ \cline{2-11} 
\multicolumn{1}{c|}{} & Top1 & \multicolumn{1}{c|}{Top 5} & Top1 & \multicolumn{1}{c|}{Top 5} & Top1 & \multicolumn{1}{c|}{Top 5} & Top1 & \multicolumn{1}{c|}{Top 5} & Top1 & Top 5 \\ \hline
\multicolumn{11}{c}{\textbf{Supervised training from scratch}} \\ \hline
\multicolumn{1}{c|}{0.0} & 76.13±0.04 & \multicolumn{1}{c|}{92.86±0.01} & 61.59±0.04 & \multicolumn{1}{c|}{82.39±0.03} & 61.80±0.05 & \multicolumn{1}{c|}{82.40±0.02} & 65.30±0.05 & \multicolumn{1}{c|}{85.09±0.02} & 66.11±0.05 & 85.92±0.02 \\ \hline
\multicolumn{1}{c|}{0.2} & 75.78±0.04 & \multicolumn{1}{c|}{92.76±0.02} & 71.68±0.01 & \multicolumn{1}{c|}{90.48±0.03} & 71.83±0.04 & \multicolumn{1}{c|}{90.48±0.02} & 72.09±0.04 & \multicolumn{1}{c|}{90.73±0.01} & 72.26±0.05 & 90.90±0.01 \\
\multicolumn{1}{c|}{0.4} & 75.88±0.03 & \multicolumn{1}{c|}{92.92±0.03} & 72.37±0.04 & \multicolumn{1}{c|}{90.35±0.02} & 72.31±0.04 & \multicolumn{1}{c|}{90.78±0.03} & 72.43±0.05 & \multicolumn{1}{c|}{91.12±0.03} & 72.77±0.05 & 91.09±0.03 \\
\multicolumn{1}{c|}{0.6} & 76.14±0.06 & \multicolumn{1}{c|}{93.05±0.03} & 72.99±0.03 & \multicolumn{1}{c|}{90.78±0.03} & 72.59±0.04 & \multicolumn{1}{c|}{91.24±0.03} & 73.11±0.03 & \multicolumn{1}{c|}{91.41±0.02} & 73.42±0.01 & 91.52±0.01 \\
\multicolumn{1}{c|}{0.8} & 75.63±0.05 & \multicolumn{1}{c|}{92.73±0.02} & 73.11±0.06 & \multicolumn{1}{c|}{91.23±0.02} & 73.02±0.02 & \multicolumn{1}{c|}{91.43±0.02} & 73.33±0.02 & \multicolumn{1}{c|}{91.62±0.02} & 73.51±0.06 & 91.50±0.01 \\ \hline

\multicolumn{1}{c|}{1.0} & 74.62±0.03 & \multicolumn{1}{c|}{91.88±0.02} & 68.23±0.04 & \multicolumn{1}{c|}{87.80±0.02} & 69.02±0.02 & \multicolumn{1}{c|}{88.48±0.03} & 69.83±0.03 & \multicolumn{1}{c|}{89.19±0.03} & 70.30±0.03 & 89.39±0.02 \\ \hline

\multicolumn{11}{c}{\textbf{Self-supervised  pre-taining on contrastive learning integrating MPD (probability for pre-training=0.8)}} \\ \hline
\multicolumn{1}{c|}{0.2} & 75.43±0.02 & \multicolumn{1}{c|}{92.59±0.01} & 72.38±0.06 & \multicolumn{1}{c|}{90.71±0.02} & 72.26±0.04 & \multicolumn{1}{c|}{90.83±0.03} & 72.40±0.05 & \multicolumn{1}{c|}{90.88±0.01} & 72.61±0.04 & 90.91±0.02 \\
\multicolumn{1}{c|}{0.4} & 75.58±0.03 & \multicolumn{1}{c|}{92.67±0.01} & 72.87±0.01 & \multicolumn{1}{c|}{91.28±0.01} & 73.17±0.04 & \multicolumn{1}{c|}{91.39±0.02} & 73.34±0.03 & \multicolumn{1}{c|}{91.58±0.03} & 73.41±0.02 & 91.49±0.03 \\
\multicolumn{1}{c|}{0.6} & 75.41±0.04 & \multicolumn{1}{c|}{92.64±0.03} & 73.43±0.06 & \multicolumn{1}{c|}{91.41±0.01} & 73.01±0.05 & \multicolumn{1}{c|}{91.50±0.02} & 73.37±0.02 & \multicolumn{1}{c|}{91.64±0.01} & 73.61±0.05 & 91.60±0.02 \\
\multicolumn{1}{c|}{0.8} & 75.23±0.01 & \multicolumn{1}{c|}{92.46±0.01} & 73.54±0.03 & \multicolumn{1}{c|}{91.59±0.02} & 73.41±0.02 & \multicolumn{1}{c|}{91.57±0.01} & 73.60±0.03 & \multicolumn{1}{c|}{91.73±0.02} & 73.79±0.04 & 91.84±0.02 \\ \hline

\multicolumn{1}{c|}{1.0} & 74.98±0.02 & \multicolumn{1}{c|}{91.35±0.02} & 68.44±0.03 & \multicolumn{1}{c|}{88.10±0.03} & 69.46±0.01 & \multicolumn{1}{c|}{89.05±0.02} & 70.08±0.03 & \multicolumn{1}{c|}{89.88±0.03} & 70.88±0.02 & 90.24±0.02 \\ \hline

\end{tabular}
\end{table*}

\begin{table}[]
\scriptsize
\centering
\caption{Detailed results on ImageNet-E: (a) Drop of Top1 accuracy (Lower is better) and (b) Absolute Accuracy (Higher is better) under background changes, size changes, random position (rp), random direction (rd), and average over 11 subsets.}
\label{tab:tab:imagenet-e_detailed}
\tiny
\begin{tabular}{cccccccccccccc}
\multicolumn{14}{c}{{\color[HTML]{000000} \textbf{(a) Drop of Accuracy}}}                                                                                                                               \\ \hline
\multicolumn{1}{c|}{{\color[HTML]{000000} }}                            & \multicolumn{1}{c|}{{\color[HTML]{000000} Original}} & \multicolumn{5}{c|}{{\color[HTML]{000000} Background changes}}                                                                                                                         & \multicolumn{4}{c|}{{\color[HTML]{000000} Size changes}}                                                                                       & \multicolumn{1}{c|}{{\color[HTML]{000000} Position}} & \multicolumn{1}{c|}{{\color[HTML]{000000} Direction}} & {\color[HTML]{000000} }                       \\ \cline{2-13}
\multicolumn{1}{c|}{\multirow{-2}{*}{{\color[HTML]{000000} Models}}}    & \multicolumn{1}{c|}{{\color[HTML]{000000} }}         & {\color[HTML]{000000} Inver} & {\color[HTML]{000000} \( \lambda \) = -20} & {\color[HTML]{000000} \( \lambda \) = 20} & {\color[HTML]{000000} \( \lambda \) = 20-adv} & \multicolumn{1}{c|}{{\color[HTML]{000000} Random}} & {\color[HTML]{000000} Full}  & {\color[HTML]{000000} 0.10}  & {\color[HTML]{000000} 0.08}  & \multicolumn{1}{c|}{{\color[HTML]{000000} 0.05}}  & \multicolumn{1}{c|}{{\color[HTML]{000000} rp}}       & \multicolumn{1}{c|}{{\color[HTML]{000000} rd}}        & \multirow{-2}{*}{{\color[HTML]{000000} Avg.}} \\ \hline
\multicolumn{1}{c|}{{\color[HTML]{000000} standard ResNet50}}           & \multicolumn{1}{c|}{{\color[HTML]{000000} 92.69}}    & {\color[HTML]{000000} 1.97}  & {\color[HTML]{000000} 7.30}    & {\color[HTML]{000000} 13.35}  & {\color[HTML]{000000} 29.92}      & \multicolumn{1}{c|}{{\color[HTML]{000000} 13.34}}  & {\color[HTML]{000000} 2.71}  & {\color[HTML]{000000} 7.25}  & {\color[HTML]{000000} 10.51} & \multicolumn{1}{c|}{{\color[HTML]{000000} 21.26}} & \multicolumn{1}{c|}{{\color[HTML]{000000} 26.46}}    & \multicolumn{1}{c|}{{\color[HTML]{000000} 25.12}}     & {\color[HTML]{000000} 15.72}                  \\ \hline
\multicolumn{1}{c|}{{\color[HTML]{000000} supervised: MPD (ResNet50)}} & \multicolumn{1}{c|}{{\color[HTML]{000000} 92.51}}    & {\color[HTML]{000000} 1.75}  & {\color[HTML]{000000} 6.94}    & {\color[HTML]{000000} 11.53}  & {\color[HTML]{000000} 29.66}      & \multicolumn{1}{c|}{{\color[HTML]{000000} 13.40}}  & {\color[HTML]{000000} 3.22}  & {\color[HTML]{000000} 6.07}  & {\color[HTML]{000000} 9.11}  & \multicolumn{1}{c|}{{\color[HTML]{000000} 18.54}} & \multicolumn{1}{c|}{{\color[HTML]{000000} 23.55}}    & \multicolumn{1}{c|}{{\color[HTML]{000000} 22.98}}     & {\color[HTML]{000000} 13.34}                  \\
\multicolumn{1}{c|}{{\color[HTML]{000000} supervised: MPD IB}}         & \multicolumn{1}{c|}{{\color[HTML]{000000} 92.39}}    & {\color[HTML]{000000} 1.63}  & {\color[HTML]{000000} 7.42}    & {\color[HTML]{000000} 13.67}  & {\color[HTML]{000000} 31.16}      & \multicolumn{1}{c|}{{\color[HTML]{000000} 14.32}}  & {\color[HTML]{000000} 2.62}  & {\color[HTML]{000000} 6.23}  & {\color[HTML]{000000} 9.82}  & \multicolumn{1}{c|}{{\color[HTML]{000000} 19.09}} & \multicolumn{1}{c|}{{\color[HTML]{000000} 24.75}}    & \multicolumn{1}{c|}{{\color[HTML]{000000} 22.10}}     & {\color[HTML]{000000} 13.89}                  \\
\multicolumn{1}{c|}{{\color[HTML]{000000} ssl: MPD (ResNet50)}}        & \multicolumn{1}{c|}{{\color[HTML]{000000} 92.51}}    & {\color[HTML]{000000} 1.38}  & {\color[HTML]{000000} 7.26}    & {\color[HTML]{000000} 12.22}  & {\color[HTML]{000000} 31.53}      & \multicolumn{1}{c|}{{\color[HTML]{000000} 13.76}}  & {\color[HTML]{000000} 2.85}  & {\color[HTML]{000000} 6.85}  & {\color[HTML]{000000} 10.12} & \multicolumn{1}{c|}{{\color[HTML]{000000} 20.91}} & \multicolumn{1}{c|}{{\color[HTML]{000000} 26.31}}    & \multicolumn{1}{c|}{{\color[HTML]{000000} 24.43}}     & {\color[HTML]{000000} 14.33}                  \\
\multicolumn{1}{c|}{{\color[HTML]{000000} ssl:MPD IB}}                 & \multicolumn{1}{c|}{{\color[HTML]{000000} 92.30}}    & {\color[HTML]{000000} 1.82}  & {\color[HTML]{000000} 7.38}    & {\color[HTML]{000000} 12.18}  & {\color[HTML]{000000} 30.79}      & \multicolumn{1}{c|}{{\color[HTML]{000000} 13.97}}  & {\color[HTML]{000000} 2.40}  & {\color[HTML]{000000} 6.90}  & {\color[HTML]{000000} 10.21} & \multicolumn{1}{c|}{{\color[HTML]{000000} 20.29}} & \multicolumn{1}{c|}{{\color[HTML]{000000} 25.25}}    & \multicolumn{1}{c|}{{\color[HTML]{000000} 23.71}}     & {\color[HTML]{000000} 14.08}                  \\ \hline
\multicolumn{14}{c}{{\color[HTML]{000000} \textbf{(b) Absolute Accuracy}}}                                                                                                                                                                                                                                                                                                                                                                                                                                                                                                                                                              \\ \hline
\multicolumn{1}{c|}{{\color[HTML]{000000} }}                            & \multicolumn{1}{c|}{{\color[HTML]{000000} Original}} & \multicolumn{5}{c|}{{\color[HTML]{000000} Background changes}}                                                                                                                         & \multicolumn{4}{c|}{{\color[HTML]{000000} Size changes}}                                                                                       & \multicolumn{1}{c|}{{\color[HTML]{000000} Position}} & \multicolumn{1}{c|}{{\color[HTML]{000000} Direction}} & {\color[HTML]{000000} }                       \\ \cline{2-13}
\multicolumn{1}{c|}{\multirow{-2}{*}{{\color[HTML]{000000} Models}}}    & \multicolumn{1}{c|}{{\color[HTML]{000000} }}         & {\color[HTML]{000000} Inver} & {\color[HTML]{000000} \( \lambda \) = -20} & {\color[HTML]{000000} \( \lambda \) = 20} & {\color[HTML]{000000} \( \lambda \) = 20-adv} & \multicolumn{1}{c|}{{\color[HTML]{000000} Random}} & {\color[HTML]{000000} Full}  & {\color[HTML]{000000} 0.10}  & {\color[HTML]{000000} 0.08}  & \multicolumn{1}{c|}{{\color[HTML]{000000} 0.05}}  & \multicolumn{1}{c|}{{\color[HTML]{000000} rp}}       & \multicolumn{1}{c|}{{\color[HTML]{000000} rd}}        & \multirow{-2}{*}{{\color[HTML]{000000} Avg.}} \\ \hline
\multicolumn{1}{c|}{{\color[HTML]{000000} standard ResNet50}}           & \multicolumn{1}{c|}{{\color[HTML]{000000} 92.69}}    & {\color[HTML]{000000} 90.72} & {\color[HTML]{000000} 85.39}   & {\color[HTML]{000000} 79.34}  & {\color[HTML]{000000} 62.77}      & \multicolumn{1}{c|}{{\color[HTML]{000000} 79.35}}  & {\color[HTML]{000000} 89.98} & {\color[HTML]{000000} 85.44} & {\color[HTML]{000000} 82.18} & \multicolumn{1}{c|}{{\color[HTML]{000000} 71.43}} & \multicolumn{1}{c|}{{\color[HTML]{000000} 66.23}}    & \multicolumn{1}{c|}{{\color[HTML]{000000} 67.57}}     & {\color[HTML]{000000} 78.22}                  \\ \hline
\multicolumn{1}{c|}{{\color[HTML]{000000} supervised: MPD (ResNet50)}} & \multicolumn{1}{c|}{{\color[HTML]{000000} 92.51}}    & {\color[HTML]{000000} 90.76} & {\color[HTML]{000000} 85.57}   & {\color[HTML]{000000} 80.97}  & {\color[HTML]{000000} 62.84}      & \multicolumn{1}{c|}{{\color[HTML]{000000} 79.11}}  & {\color[HTML]{000000} 89.29} & {\color[HTML]{000000} 86.44} & {\color[HTML]{000000} 83.40} & \multicolumn{1}{c|}{{\color[HTML]{000000} 73.97}} & \multicolumn{1}{c|}{{\color[HTML]{000000} 68.96}}    & \multicolumn{1}{c|}{{\color[HTML]{000000} 69.53}}     & {\color[HTML]{000000} 79.17}                  \\
\multicolumn{1}{c|}{{\color[HTML]{000000} supervised: MPD IB}}         & \multicolumn{1}{c|}{{\color[HTML]{000000} 92.39}}    & {\color[HTML]{000000} 90.76} & {\color[HTML]{000000} 84.97}   & {\color[HTML]{000000} 78.72}  & {\color[HTML]{000000} 61.24}      & \multicolumn{1}{c|}{{\color[HTML]{000000} 78.08}}  & {\color[HTML]{000000} 89.77} & {\color[HTML]{000000} 86.16} & {\color[HTML]{000000} 82.57} & \multicolumn{1}{c|}{{\color[HTML]{000000} 73.30}} & \multicolumn{1}{c|}{{\color[HTML]{000000} 67.65}}    & \multicolumn{1}{c|}{{\color[HTML]{000000} 70.29}}     & {\color[HTML]{000000} 78.50}                  \\
\multicolumn{1}{c|}{{\color[HTML]{000000} ssl: MPD (ResNet50)}}        & \multicolumn{1}{c|}{{\color[HTML]{000000} 92.51}}    & {\color[HTML]{000000} 91.13} & {\color[HTML]{000000} 85.25}   & {\color[HTML]{000000} 80.28}  & {\color[HTML]{000000} 60.98}      & \multicolumn{1}{c|}{{\color[HTML]{000000} 78.75}}  & {\color[HTML]{000000} 89.66} & {\color[HTML]{000000} 85.66} & {\color[HTML]{000000} 82.39} & \multicolumn{1}{c|}{{\color[HTML]{000000} 71.60}} & \multicolumn{1}{c|}{{\color[HTML]{000000} 66.20}}    & \multicolumn{1}{c|}{{\color[HTML]{000000} 68.08}}     & {\color[HTML]{000000} 78.18}                  \\
\multicolumn{1}{c|}{{\color[HTML]{000000} ssl:MPD IB}}                 & \multicolumn{1}{c|}{{\color[HTML]{000000} 92.30}}    & {\color[HTML]{000000} 90.49} & {\color[HTML]{000000} 84.93}   & {\color[HTML]{000000} 80.12}  & {\color[HTML]{000000} 61.51}      & \multicolumn{1}{c|}{{\color[HTML]{000000} 78.33}}  & {\color[HTML]{000000} 89.90} & {\color[HTML]{000000} 85.40} & {\color[HTML]{000000} 82.09} & \multicolumn{1}{c|}{{\color[HTML]{000000} 72.01}} & \multicolumn{1}{c|}{{\color[HTML]{000000} 67.05}}    & \multicolumn{1}{c|}{{\color[HTML]{000000} 68.59}}     & {\color[HTML]{000000} 78.22}                  \\ \hline
\end{tabular}
\end{table}

\begin{figure*}[!t]
  \centering
  \includegraphics[width=1.\columnwidth]{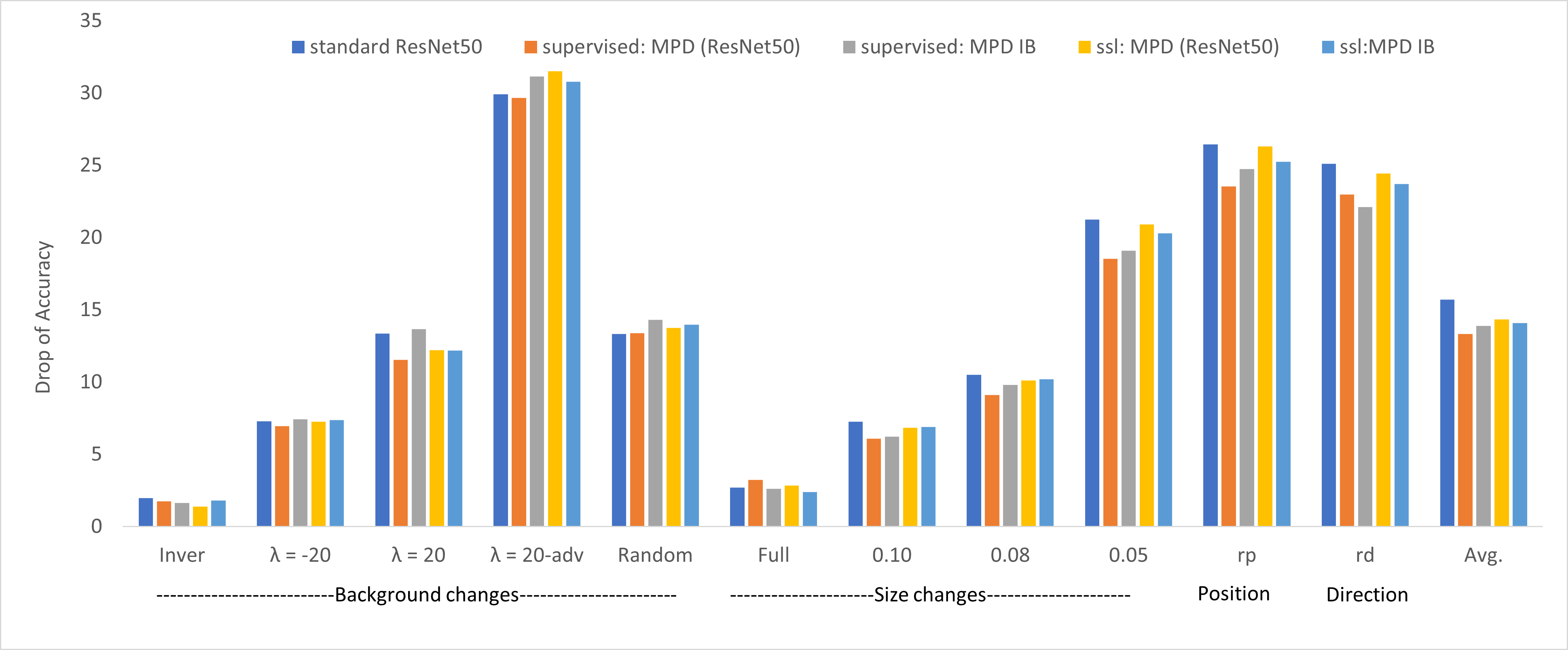}
  \caption{Detailed performance comparison on ImageNet-E \cite{li2023imagenet}: Detailed results on Drop of Top1 accuracy under background changes, size changes, random position (rp), random direction (rd), and average over 11 subsets. Lower is better. \textit{Average} reports the mean of all subsets. MPD variants outperforms standard ResNet50 on subsets which are affected with perspective distortion, e.g., size, position, and direction. Corresponding details on absolute accuracy measure are in Table \ref{tab:tab:imagenet-e_detailed}. Detailed analysis in continuation of Fig. \ref{fig:hypo2_ImageNet-E}; Fig. \ref{fig:hypo2_ImageNet-E} is given in main paper.} 
  \label{fig:imagnet_e_extended}
\end{figure*}

\begin{table*}[h]
\caption{Detailed results on ImageNet-X \cite{idrissi2022imagenet}: (a) Error ratio (close to 1.0 is best) and (b) Absolute accuracy (higher is better) of 16 factors. \textit{Average} reports the mean accuracy of all factors. Detailed results also includes \textit{supervised:MPD IB }and\textit{ ssl:MPD IB} models besides \textit{supervised:MPD} and \textit{ssl:MPD} models. All the models achieves improved error ratio with higher accuracy consistently across factors. Detailed results in continuation of Fig. \ref{fig:hypo2_ImageNet-X}; Fig. \ref{fig:hypo2_ImageNet-X} is given in main paper.}
\centering
\scriptsize
\label{tab:imagenet_x_detailed}
\begin{tabular}{cccccc}
\multicolumn{6}{c}{\textbf{(a) Error Ratio}}                                                                                                               \\ \hline
\multicolumn{1}{c|}{Factor}            & \multicolumn{1}{c|}{standard ResNet50} & supervised: MPD & ssl: MPD      & supervised: MPD IB & ssl: MPD IB   \\ \hline
\multicolumn{1}{c|}{pose}              & \multicolumn{1}{c|}{0.70}              & 0.72             & 0.74           & 0.73                & 0.72           \\
\multicolumn{1}{c|}{pattern}           & \multicolumn{1}{c|}{0.77}              & 0.89             & 0.78           & 0.87                & 0.87           \\
\multicolumn{1}{c|}{partial\_view}     & \multicolumn{1}{c|}{1.13}              & 0.89             & 0.86           & 1.07                & 0.90           \\
\multicolumn{1}{c|}{background}        & \multicolumn{1}{c|}{1.14}              & 1.08             & 1.09           & 1.09                & 1.07           \\
\multicolumn{1}{c|}{brighter}          & \multicolumn{1}{c|}{1.17}              & 1.09             & 1.09           & 1.19                & 1.13           \\
\multicolumn{1}{c|}{color}             & \multicolumn{1}{c|}{1.19}              & 1.21             & 1.18           & 1.23                & 1.20           \\
\multicolumn{1}{c|}{larger}            & \multicolumn{1}{c|}{1.32}              & 1.30             & 1.22           & 1.24                & 1.35           \\
\multicolumn{1}{c|}{smaller}           & \multicolumn{1}{c|}{1.50}              & 1.44             & 1.50           & 1.46                & 1.50           \\
\multicolumn{1}{c|}{darker}            & \multicolumn{1}{c|}{1.72}              & 1.52             & 1.51           & 1.46                & 1.53           \\
\multicolumn{1}{c|}{shape}             & \multicolumn{1}{c|}{1.75}              & 1.63             & 1.61           & 1.53                & 1.59           \\
\multicolumn{1}{c|}{object\_blocking}  & \multicolumn{1}{c|}{1.92}              & 1.66             & 1.61           & 1.75                & 1.62           \\
\multicolumn{1}{c|}{style}             & \multicolumn{1}{c|}{1.93}              & 1.73             & 1.63           & 1.80                & 1.71           \\
\multicolumn{1}{c|}{multiple\_objects} & \multicolumn{1}{c|}{1.97}              & 1.76             & 1.72           & 1.84                & 1.86           \\
\multicolumn{1}{c|}{subcategory}       & \multicolumn{1}{c|}{2.10}              & 1.92             & 1.90           & 1.88                & 1.98           \\
\multicolumn{1}{c|}{person\_blocking}  & \multicolumn{1}{c|}{2.17}              & 1.95             & 1.97           & 1.94                & 2.03           \\
\multicolumn{1}{c|}{texture}           & \multicolumn{1}{c|}{2.39}              & 2.20             & 2.18           & 2.21                & 2.16           \\ \hline
\multicolumn{1}{c|}{Avg}               & \multicolumn{1}{c|}{1.55}              & \textbf{1.44}    & \textbf{1.41}  & \textbf{1.45}       & \textbf{1.45}  \\ \hline
\multicolumn{6}{c}{\textbf{(b) Absolute Accuracy}}                                                                                                         \\ \hline
\multicolumn{1}{c|}{Factor}            & \multicolumn{1}{c|}{standard ResNet50} & supervised: MPD & ssl: MPD      & supervised: MPD IB & ssl: MPD IB   \\ \hline
\multicolumn{1}{c|}{pose}              & \multicolumn{1}{c|}{82.40}             & 81.47            & 80.55          & 81.12               & 81.00          \\
\multicolumn{1}{c|}{pattern}           & \multicolumn{1}{c|}{80.48}             & 77.28            & 79.55          & 77.41               & 77.27          \\
\multicolumn{1}{c|}{partial\_view}     & \multicolumn{1}{c|}{71.32}             & 77.27            & 77.49          & 72.41               & 76.24          \\
\multicolumn{1}{c|}{background}        & \multicolumn{1}{c|}{71.13}             & 72.37            & 71.53          & 71.87               & 71.85          \\
\multicolumn{1}{c|}{brighter}          & \multicolumn{1}{c|}{70.45}             & 72.10            & 71.32          & 69.30               & 70.38          \\
\multicolumn{1}{c|}{color}             & \multicolumn{1}{c|}{69.85}             & 69.12            & 68.96          & 68.18               & 68.44          \\
\multicolumn{1}{c|}{larger}            & \multicolumn{1}{c|}{66.67}             & 66.67            & 68.00          & 68.00               & 64.67          \\
\multicolumn{1}{c|}{smaller}           & \multicolumn{1}{c|}{62.07}             & 63.11            & 60.62          & 62.30               & 60.66          \\
\multicolumn{1}{c|}{darker}            & \multicolumn{1}{c|}{56.56}             & 61.17            & 60.47          & 62.28               & 59.93          \\
\multicolumn{1}{c|}{shape}             & \multicolumn{1}{c|}{55.67}             & 58.32            & 57.88          & 60.47               & 58.32          \\
\multicolumn{1}{c|}{object\_blocking}  & \multicolumn{1}{c|}{51.28}             & 57.50            & 57.69          & 54.64               & 57.50          \\
\multicolumn{1}{c|}{style}             & \multicolumn{1}{c|}{51.16}             & 55.81            & 57.38          & 53.33               & 55.13          \\
\multicolumn{1}{c|}{multiple\_objects} & \multicolumn{1}{c|}{50.00}             & 55.00            & 55.00          & 52.50               & 51.16          \\
\multicolumn{1}{c|}{subcategory}       & \multicolumn{1}{c|}{46.84}             & 50.94            & 50.09          & 51.28               & 48.03          \\
\multicolumn{1}{c|}{person\_blocking}  & \multicolumn{1}{c|}{45.00}             & 50.00            & 48.33          & 49.91               & 46.67          \\
\multicolumn{1}{c|}{texture}           & \multicolumn{1}{c|}{39.36}             & 43.62            & 42.91          & 42.91               & 43.26          \\ \hline
\multicolumn{1}{c|}{Avg}               & \multicolumn{1}{c|}{60.64}             & \textbf{63.23}   & \textbf{62.99} & \textbf{62.37}      & \textbf{61.91} \\ \hline
\end{tabular}
\end{table*}

\section{Continuation (section \ref{ob3}): MPD's effects on self-supervised learning} \label{ob3_c}
SimCLR \cite{chen2020simple} self-supervised MPD models demonstrate notable effectiveness on ImageNet-E and ImageNet-X, detailed in Tables \ref{tab:tab:imagenet-e_detailed} and \ref{tab:imagenet_x_detailed}. On ImageNet-E, self-supervised MPD variants show a marked reduction in error ratios in size-related subsets, with error ratios like 1.44 for 'larger' and 1.46 for 'smaller', indicating a strong grasp of perspective challenges. In ImageNet-X, these models excel in handling 'object blocking' (error ratio 1.61) and 'person blocking' (error ratio 1.97) factors, achieving higher absolute accuracy compared to standard ResNet50, as illustrated in Fig. \ref{fig:hypo2_ImageNet-X}. The detailed analysis in Fig. \ref{fig:imagnet_e_extended} further emphasizes the models' enhanced adaptability in self-supervised supervised learning, underlining their robustness in understanding complex visual scenarios.

\noindent\textbf{Linear evaluation} results of SimcLR \cite{chen2020simple} self-supervised MPD (ssl:MPD) and MPD IB (ssl:MPD IB) models, presented in Tables \ref{tab:Hype3-MPD-LE} and \ref{tab:Hype3-MPDIB-LE}, align remarkably with the trends observed in their fine-tuned counterparts (Tables \ref{tab:hype1A} and \ref{tab:hype1A_IB}). This complementary relationship underscores the robustness of MPD models in both linear and fine-tuned scenarios.
Specifically, ssl:MPD and ssl:MPD IB demonstrate impressive handling of perspective distortion in the ImageNet-PD subsets during linear evaluation, with ssl:MPD achieving an average Top-1 accuracy of 50.33±0.04, and ssl:MPD IB even more impressive at 51.62±0.03. These figures notably surpass the standard self-supervised model's performance, which scores 35.08±0.04 and 32.53±0.04, respectively. This substantial improvement in handling perspective distortions is consistent with the fine-tuned results, where both ssl:MPD and ssl:MPD IB models maintained high accuracy levels, showing their efficacy in learning robust representations. Furthermore, the linear evaluation results reveal that these improvements are not at the expense of general performance. Both ssl:MPD and ssl:MPD IB sustain competitive accuracies on the standard ImageNet validation set, mirroring the fine-tuning outcomes where they demonstrated broad applicability and versatility. These findings from the linear evaluation and fine-tuning phases collectively reinforce the conclusion in our main paper about the enhanced capabilities of MPD models in self-supervised learning environments. They affirm the models' ability to learn perspective-aware representations that are not only applicable to specialized scenarios involving perspective distortions but also effective across a wide range of visual recognition tasks. Linear evaluation results for ssl:MPD and ssl:MPD IB strongly reinforce the claim made in our main paper about their enhanced ability in self-supervised learning contexts. These models show not just incremental but significant improvements in handling complex visual distortions, establishing MPD as a powerful tool for self-supervised learning in real-world applications where perspective variations are prevalent.
\begin{table*}[h]
\caption{Linear evaluation performance comparison for self-supervised models, standard SimCLR \cite{chen2020simple} and  \textit{ssl:MPD} model on original ImageNet validation set and ImageNet-PD subsets \textit{(black background}). Self-supervised pretraining was performed on ImageNet dataset with batch size of 512 on linear learning rate for 100 epochs, refer SimCLR\cite{chen2020simple}. Top-1 accuracy reported. The probability of MPD is set to 0.8 for self-supervised pre-training and linear evaluation. ResNet50 is common backbone.  }
\centering
\scriptsize
\label{tab:Hype3-MPD-LE}
\tiny
\begin{tabular}{c|c|cccc}
\hline
Model          & Original Validation Set & \begin{tabular}[c]{@{}c@{}}Perspectively Distorted\\ Top-view (PD-T)\end{tabular} & \begin{tabular}[c]{@{}c@{}}Perspectively Distorted\\ Bottom-view (PD-B)\end{tabular} & \begin{tabular}[c]{@{}c@{}}Perspectively Distorted\\ Left-view (PD-L)\end{tabular} & \begin{tabular}[c]{@{}c@{}}Perspectively Distorted\\ Right-view (PD-R)\end{tabular} \\ \hline
standard SimCLR & 60.14              & 34.22                                                                        & 32.87                                                                           & 36.33                                                                         & 36.89                                                                         \\ \hline
ssl:MPD       & 60.02±0.03              & 50.63±0.03                                                                        & 49.65±0.03                                                                           & 50.41±0.03                                                                         & 50.64±0.05                                                                        \\ \hline
\end{tabular}
\end{table*}

\begin{table*}[h]
\caption{Linear evaluation performance comparison for self-supervised models, standard SimCLR \cite{chen2020simple} and  \textit{ssl:MPD IB} model on original ImageNet validation set and ImageNet-PD subsets (\textit{with integrated padding background}). Self-supervised pretraining was performed on batch size of 512 on linear learning rate for 100 epochs, refer SimCLR\cite{chen2020simple}. Top-1 accuracy reported. The probability of MPD IB is set to 0.8 for self-supervised pre-training and linear evaluation. ResNet50 is common backbone.}
\centering
\scriptsize
\label{tab:Hype3-MPDIB-LE}
\tiny
\begin{tabular}{c|c|cccc}
\hline
Model          & Original Validation Set & \begin{tabular}[c]{@{}c@{}}Perspectively Distorted\\ Top-view (PD-TI)\end{tabular} & \begin{tabular}[c]{@{}c@{}}Perspectively Distorted\\ Bottom-view (PD-BI)\end{tabular} & \begin{tabular}[c]{@{}c@{}}Perspectively Distorted\\ Left-view (PD-LI)\end{tabular} & \begin{tabular}[c]{@{}c@{}}Perspectively Distorted\\ Right-view (PD-RI)\end{tabular} \\ \hline
standard SimCLR & 60.14              & 30.41                                                                         & 30.18                                                                            & 34.81                                                                          & 34.72                                                                           \\ \hline
ssl:MPD IB    & 60.05±0.04              & 51.30±0.03                                                                         & 51.45±0.04                                                                            & 51.81±0.03                                                                          & 51.93±0.03                                                                           \\ \hline
\end{tabular}
\end{table*}

\noindent\textbf{MPD with DINO}: Incorporating MPD into the knowledge distillation-based self-supervised method DINO \cite{caron2021emerging} demonstrates MPD's adaptability and effectiveness across different SSL approaches. 
The linear evaluation results for ssl:DINO-MPD and ssl:DINO-MPD IB, as detailed results presented in Tables \ref{tab:Hype3-DINO-MPD-LE} and \ref{tab:Hype3-DINO-MPD-IB-LE}, demonstrate the significant impact of integrating MPD and MPD IB into the DINO method. Both ssl:DINO-MPD and ssl:DINO-MPD IB models substantially improves robustness against perspective distortion when evaluated on the ImageNet-PD subsets. Specifically, ssl:DINO-MPD achieves an average Top-1 accuracy of 60.70±0.04, while ssl:DINO-MPD IB further enhances this performance, achieving an even higher average Top-1 accuracy of 61.30±0.02. The integration of MPD and MPD IB into DINO demonstrates their adaptability and effectiveness in improving performance across self-supervised approaches, especially in scenarios involving perspective distortion.

\begin{table*}[h]
\caption{Linear evaluation performance comparison for knowledge distillation based self-supervised method DINO \cite{caron2021emerging}. Standard DINO is original method pretrained for 100 epochs with batch size of 512 and \textit{ssl: DINO-MPD} model is MPD integrated DINO, following same pretraining hyperparameters. Post-pretraining, both the models are linearly trained for 100 epochs with batch size of 256. Top-1 accuracy reported. Evaluated on original ImageNet validation set and ImageNet-PD subsets (\textit{black background}). The probability of MPD is set to 0.4 for self-supervised pre-training and 0.8 for linear evaluation. ViT-small transformer \cite{dosovitskiy2020image} is common backbone. 
}
\scriptsize
\centering
\label{tab:Hype3-DINO-MPD-LE}
\tiny
\begin{tabular}{c|c|cccc}
\hline
Model         & Original Validation Set & \begin{tabular}[c]{@{}c@{}}Perspectively Distorted\\ Top-view (PD-T)\end{tabular} & \begin{tabular}[c]{@{}c@{}}Perspectively Distorted\\ Bottom-view (PD-B)\end{tabular} & \begin{tabular}[c]{@{}c@{}}Perspectively Distorted\\ Left-view (PD-L)\end{tabular} & \begin{tabular}[c]{@{}c@{}}Perspectively Distorted\\ Right-view (PD-R)\end{tabular} \\ \hline
standard DINO & 74.00                   & 46.31                                                                             & 46.05                                                                                & 46.15                                                                              & 45.98                                                                               \\ \hline
ssl:DINO-MPD & 72.36±0.01              & 60.72±0.02                                                                        & 60.86±0.02                                                                           & 60.63±0.02                                                                         & 60.58±0.02                                                                          \\ \hline
\end{tabular}
\end{table*}
\begin{table*}[h]
\caption{Linear evaluation performance comparison for DINO-MPD IB: Standard DINO \cite{caron2021emerging} is original method pretrained for 100 epochs with batch size of 512 and ssl: \textit{DINO-MPD IB model} is MPD IB integrated DINO, following same pretraining hyperparameters. Post-pretraining, both the models are linearly trained for 100 epochs with batch size of 256. Top-1 accuracy reported. Evaluated on original ImageNet validation set and ImageNet-PD subsets (\textit{with integrated padding background}). The probability of MPD IB is set to 0.4 for self-supervised pre-training and 0.8 for linear evaluation. ViT-small transformer \cite{dosovitskiy2020image} is common backbone.  }
\scriptsize
\centering
\label{tab:Hype3-DINO-MPD-IB-LE}
\tiny
\begin{tabular}{c|c|cccc}
\hline
Model         & Original Validation Set & \begin{tabular}[c]{@{}c@{}}Perspectively Distorted\\ Top-view (PD-T)\end{tabular} & \begin{tabular}[c]{@{}c@{}}Perspectively Distorted\\ Bottom-view (PD-B)\end{tabular} & \begin{tabular}[c]{@{}c@{}}Perspectively Distorted\\ Left-view (PD-L)\end{tabular} & \begin{tabular}[c]{@{}c@{}}Perspectively Distorted\\ Right-view (PD-R)\end{tabular} \\ \hline
standard DINO & 74.00                   & 45.22                                                                             & 44.80                                                                                & 44.05                                                                              & 45.66                                                                               \\ \hline
ssl:DINO-MPD IB & 72.05±0.01              & 61.04±0.01                                                                        & 61.50±0.02                                                                           & 60.98±0.03                                                                         & 61.66±0.02                                                                           \\ \hline
\end{tabular}
\end{table*}
\begin{figure*}[h]
  \centering
  \includegraphics[width=1.0\columnwidth]{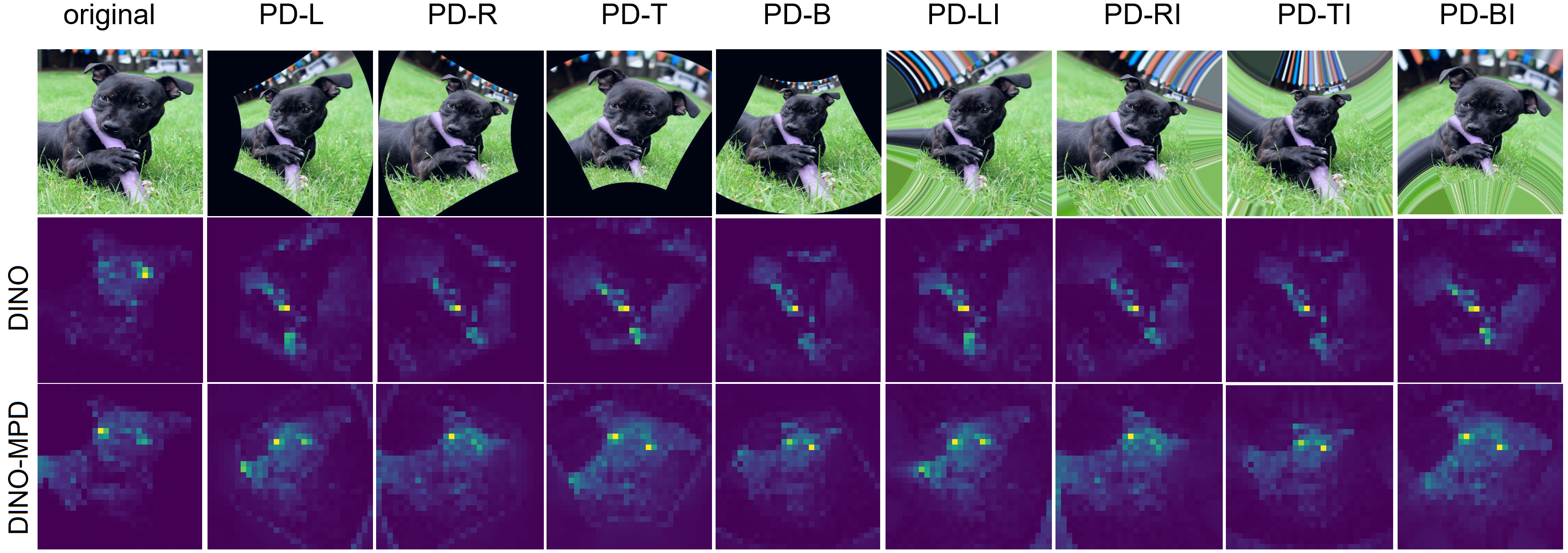}
  \caption{Activation maps of visual attention: Comparison of attention maps between DINO and MPD integrated DINO (\textit{ssl:DINO-MPD} model). First attention head is used to visualize the attention. Example taken from ImageNet-PD set with original image. Both the models are pretrained for 100 epochs using ViT small architecture. DINO-MPD demonstrates on learning perspective distortion in input image.} 
  \label{fig:dino_attn_example}
\end{figure*}
Furthermore, the activation maps from ssl:DINO-MPD, as visualized in Fig. \ref{fig:dino_attn_example}, qualitatively demonstrate its superior understanding of perspective distortion in input images. These maps reveal a more focused and relevant visual attention, indicating a deeper comprehension of the image structure and content, especially in perspectively distorted scenarios. The success of ssl:DINO-MPD in both quantitative performance and qualitative visual attention underscores the versatility of MPD. It shows that MPD can be effectively combined with various SSL methods, preserving their inherent strengths and augmenting them to handle better the challenges posed by perspective distortions. This enhances the claims made in our main paper about MPD's capabilities, providing a strong case for its widespread applicability in diverse self-supervised learning approaches.

\section{MPD vs. other augmentations}
Table \ref{tab:imagenetaugs} demonstrates the effectiveness of MPD against other popular augmentation methods on the ImageNet validation set, ImageNet-PD subsets, and ImageNet-E \cite{li2023imagenet}. Augmentation methods such as Mixup \cite{zhang2018mixup}, Cutout \cite{devries2017improved}, AugMix \cite{hendrycksaugmix}, and Pixmix \cite{hendrycks2022pixmix} showed slight improvements on the ImageNet-PD benchmarks. For instance, Mixup improved the Top-1 accuracy to 65.46\% on PD-T and 68.43\% on PD-R, but these methods generally degraded performance on ImageNet-E, with Mixup dropping to 82.38\%. This suggests that these augmentations do not adequately capture the characteristics of perspective distortions. In contrast, MPD-trained models, both in supervised and self-supervised settings, consistently outperformed all other methods on the perspective distortion benchmarks and ImageNet-E. The supervised MPD model achieved impressive Top-1 accuracies of 73.00\%, 72.31\%, 73.50\%, and 72.91\% on PD-T, PD-B, PD-L, and PD-R, respectively, and 85.92\% on ImageNet-E. The self-supervised MPD model further enhanced performance, achieving Top-1 accuracies of 73.23\%, 73.06\%, 73.55\%, and 73.54\% on PD-T, PD-B, PD-L, and PD-R, respectively, and 86.66\% on ImageNet-E. These results indicate that MPD effectively addresses perspective distortions, maintaining high accuracy across both the original and distorted datasets.
\begin{table}[h]
\centering
\caption{Comparisons of MPD with other popular augmentation methods. ResNet50: baseline model trained on ImageNet \cite{deng2009imagenet} and subsequent models incorporates mentioned augmentation methods. All the models are evaluated on original ImageNet validation set, ImageNet-PD subsets (black background), and Imagenet-E. Both MPD trained models (supervised and SimCLR \cite{chen2020simple} self-supervised) outperformed on perspective distortion benchmark ImageNet-PD and ImageNet-E while maintain performance on original ImageNet validation set.} 
\label{tab:imagenetaugs}
\begin{tabular}{c|c|cccc|c}
\hline
\multirow{3}{*}{Method}                               & \multirow{2}{*}{\begin{tabular}[c]{@{}c@{}}Original \\ Val set\end{tabular}} & \multicolumn{4}{c|}{ImageNet PD}                                                                                                                                                                                                                                                                                       & \multirow{2}{*}{ImageNet-E} \\ \cline{3-6}
                                                      &                                                                              & \multicolumn{1}{c|}{\begin{tabular}[c]{@{}c@{}}Top-view \\ (PD-T)\end{tabular}} & \multicolumn{1}{c|}{\begin{tabular}[c]{@{}c@{}}Bottom-view\\  (PD-B)\end{tabular}} & \multicolumn{1}{c|}{\begin{tabular}[c]{@{}c@{}}Left-view \\ (PD-L)\end{tabular}} & \begin{tabular}[c]{@{}c@{}}Right-view \\ (PD-R)\end{tabular} &                             \\ \cline{2-7} 
                                                      & Top1                                                                         & Top1/Top5                                                                       & Top1/Top5                                                                          & Top1/Top5                                                                        & Top1/Top5                                                    & Top1                        \\ \hline
ResNet50 \cite{pytorch_vision}                                                  & 76.13                                                                        & 63.37/83.61                                                                     & 61.15/81.86                                                                        & 65.20/85.13                                                                      & 65.84/85.64                                                  & 84.28                       \\
+Mixup \cite{zhang2018mixup}                                                & 77.46                                                                        & 65.46/85.15                                                                     & 66.79/85.91                                                                        & 68.02/86.98                                                                      & 68.43/87.22                                                  & 82.38                       \\
+Cutout \cite{devries2017improved}                                               & 77.08                                                                        & 64.27/84.20                                                                     & 62.04/82.86                                                                        & 65.45/85.40                                                                      & 65.61/85.46                                                  & 79.58                       \\
+AugMix \cite{hendrycksaugmix}                                               & 77.53                                                                        & 64.12/84.01                                                                     & 62.90/82.33                                                                        & 65.95/85.94                                                                      & 66.49/86.03                                                  & 80.97                       \\
+Pixmix \cite{hendrycks2022pixmix}                                               & 77.37                                                                        & 65.52/85.21                                                                     & 64.76/84.54                                                                        & 67.26/86.67                                                                      & 67.56/86.80                                                  & 82.38                       \\ \hline
+MPD                                                  & 76.34                                                                        & \textbf{73.00/90.69}                                                            & \textbf{72.31/90.81}                                                               & \textbf{73.50/91.33}                                                             & \textbf{72.91/91.29}                                         & \textbf{85.92}              \\
\begin{tabular}[c]{@{}c@{}}+MPD \\ (SSL)\end{tabular} & 76.77                                                                        & \textbf{73.23/91.39}                                                            & \textbf{73.06/91.54}                                                               & \textbf{73.55/91.53}                                                             & \textbf{73.54/91.65}                                         & \textbf{86.66}              \\ \hline
\end{tabular}
\end{table}

\section{Continuation (section \ref{subsec:fisheye}): MPD's generalizability}
This section extents the results for MPD's adaptability in Fisheye, Crowd counting, and Person re-identifiation.
\subsection{Crowd counting} \label{sec:supp_cc}
ResNet50 encoder is employed in P2P-Net \cite{song2021rethinking}, the integration of MPD and MPD-AutoCrowd continue to demonstrate remarkable improvements, shown in Table \ref{tab:MPD_auto_crowd_extended}. For instance, the standard P2P-Net with ResNet50 records an MAE of 299.5 and an MSE of 382.48 on the UCF-CC50 dataset. In contrast, the incorporation of MPD significantly enhances performance, yielding an MAE of 95.10 and an MSE of 132.47.
\begin{table}[h]
\caption{Ablation on encoder part of crowd-counting method P2P-Net with MPD-CC \& MPD-AutoCrowd}
\centering
\footnotesize
\label{tab:MPD_auto_crowd_extended}
\begin{tabular}{c|c|cc}
\hline
\multirow{2}{*}{\textbf{Method}}                          & \multirow{2}{*}{\textbf{Encoder}} & \multicolumn{2}{c}{\textbf{UCF\_CC-50}} \\ \cline{3-4} 
                                                          &                                   & \textbf{MAE}       & \textbf{MSE}       \\ \hline
P2P-Net                                                   & standard VGG16                    & 172.72             & 256.18             \\
MPD                                                      & standard VGG16                    & \textbf{101.30}    & \textbf{140.65}    \\
\begin{tabular}[c]{@{}c@{}}MPD-\\ AutoCrowd\end{tabular} & standard VGG16                    & \textbf{96.80}     & \textbf{139.50}    \\
P2P-Net                                                   & standard ResNet50                 & 299.5             & 382.48             \\
MPD                                                      & standard ResNet50                 & \textbf{95.10}     & \textbf{132.47}    \\
\begin{tabular}[c]{@{}c@{}}MPD-\\ AutoCrowd\end{tabular} & standard ResNet50                 & \textbf{105.4}     & \textbf{124.61}    \\ \hline
\end{tabular}
\end{table}
Moreover, MPD-AutoCrowd further improves these metrics, achieving an MAE of 105.4 and an MSE of 124.61. These results underscore the capability of MPD to effectively deal with the complexities of crowd scenes, even when coupled with a different encoder. This ablation study complements the findings presented in Table \ref{tab:hype5A} from the main paper, where MPD-CC and MPD-AutoCrowd outperform other crowd-counting methods across various benchmarks. The consistent improvement across different datasets and encoder configurations, especially with the challenging UCF-CC50 dataset, highlights MPD's robustness and adaptability. MPD-AutoCrowd's notable performance in handling varied crowd densities with high accuracy, particularly with an MAE of 96.80 and MSE of 139.50 on UCF-CC50, further validates the efficacy of MPD in specialized applications, e.g., crowd counting. 

\begin{figure}[htp]
  \centering
  \includegraphics[width=.9\textwidth]{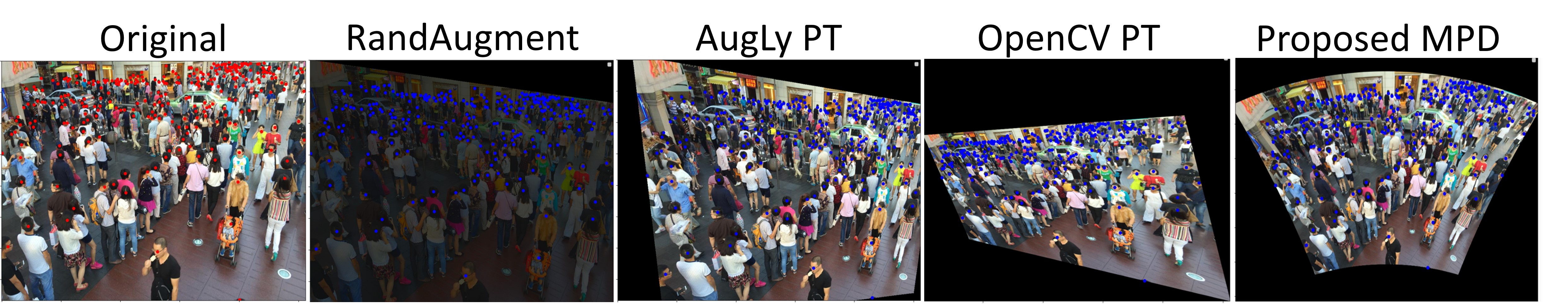}
  \caption{Applied augmentations on crowd counting}
  \label{fig:augs_cc}
\end{figure}

Further, We adapted three augmentation methods tailored for crowd counting and observed MPD outperforms them, as shown in Table \ref{tab:cc_data_augs}. Specifically, we compared MPD with RandAugment \cite{cubuk2020randaugment}, OpenCV PT \cite{zhang2020perspective}, and AugLy PT \cite{papakipos2022augly} across three datasets: SHHA, SHHB, and UCF\_CC\_50. MPD consistently achieved lower MAE and MSE values, indicating superior performance in handling crowd counting tasks. For instance, MPD-CC achieved a MAE of 51.93 and MSE of 84.3 on SHHA, significantly outperforming AugLy PT's 56.18 MAE and 107.67 MSE. Furthermore, the samples shown in Fig. \ref{fig:augs_cc} illustrates that perspective transform augmentations from \cite{papakipos2022augly} and \cite{zhang2020perspective} do not capture the non-linearity inherent in real-world perspective distortions, as they primarily rely on affine transformations or translations. This limitation underscores the effectiveness of MPD in capturing complex distortions, leading to its superior performance in crowd counting benchmarks.
\begin{table}[h]
\centering
\caption{Comparisons on augmentation methods in crowd counting. \textit{PT: Perspective Transform. Metrics: MAE and MSE errors}}
\label{tab:cc_data_augs}
\begin{tabular}{c|cc|cc|cc}
\hline
\multirow{2}{*}{Method}                    & \multicolumn{2}{c|}{SHHA}       & \multicolumn{2}{c|}{SHHB}     & \multicolumn{2}{c}{UCF\_CC\_50}  \\ \cline{2-7} 
                                                         & MAE            & MSE            & MAE           & MSE           & MAE            & MSE             \\ \hline
RandAugment \cite{cubuk2020randaugment}                  & 59.81          & 113.48         & 10.03         & 19.84         & 187.62         & 271.28          \\
OpenCV PT \cite{zhang2020perspective} & 57.73          & 109.82         & 7.78          & 13.41         & 181.05         & 263.52          \\
AugLy PT \cite{papakipos2022augly}    & 56.18          & 107.67         & 8.09          & 14.05         & 174.74         & 259.08          \\ \hline
MPD-CC                                                   & \textbf{51.93} & \textbf{84.3}  & \textbf{6.73} & \textbf{9.82} & \textbf{101.3} & \textbf{140.65} \\
MPD-AutoCrowd                                            & \textbf{50.81} & \textbf{85.01} & \textbf{6.61} & \textbf{9.58} & \textbf{96.8}  & \textbf{139.5}  \\ \hline
\end{tabular} 
\end{table}
These results affirm the main paper's assertion about the potential of MPD in complex visual recognition tasks and demonstrate its versatility in adapting to different encoder configurations while maintaining high performance.
\subsection{Transfer learning on Fisheye images}
\begin{figure}[h]
  \centering
  \includegraphics[width=0.7\columnwidth]{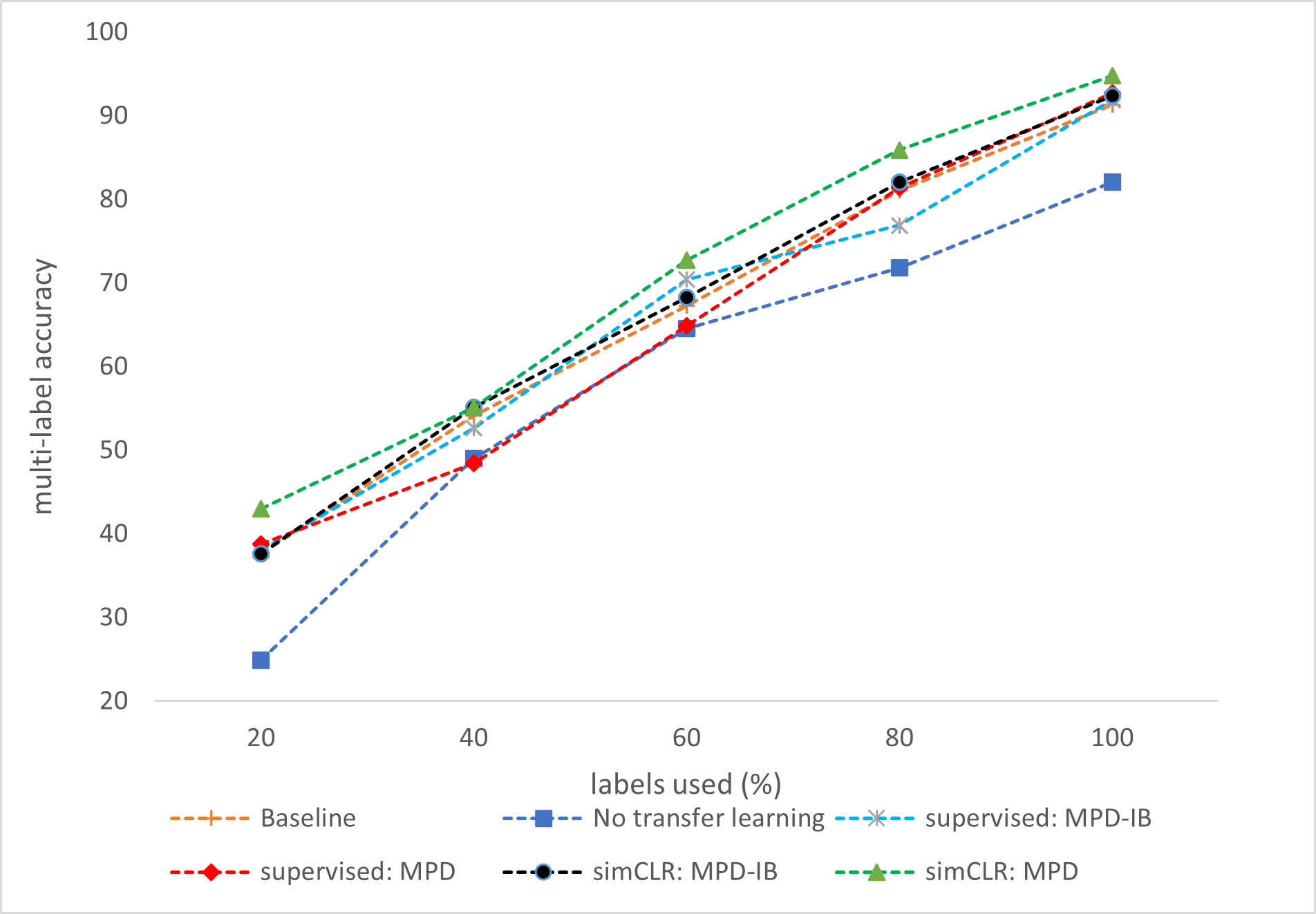}
  \caption{Label efficiency on Fisheye multi-label classification: Performance comparison across variants of MPD in supervised and self-supervised models with baseline (standard ResNet50) and no-transfer-learning model. All the models evaluated with 20\%, 40\%, 60\%, 80\%, and 100\% labels.} 
  \label{fig:voc_label_eff}
\end{figure}

Extended empirical findings, as illustrated in the accompanying graph ( Fig. \ref{fig:voc_label_eff}), consistently demonstrate that the \textit{ssl:MPD} and \textit{ssl:MPD IB} models excel in multi-label classification tasks on the VOC-360 fisheye dataset, showing a marked improvement in accuracy with an increasing percentage of labels used.

Complementing these results, Tables~\ref{tab:MPD_voc_prob_labels} and~\ref{tab:MPD_ib_voc_prob_labels} provide further insights into the models' performances across a spectrum of MPD probabilities. The data reveal an optimal probability threshold for the MPD transformation application, which maximizes label efficiency. Notably, a probability setting of 0.5 for MPD yields the highest accuracy when all labels are utilized, indicating a non-linear relationship between the transformation probability and classification accuracy.

\begin{table}[h]
\centering
\caption{Label efficiency on Fisheye multi-label classification across different probabilities of MPD. ResNet50 is encoder.}
\scriptsize
\label{tab:MPD_voc_prob_labels}
\begin{tabular}{c|ccccc}
\hline
\multirow{2}{*}{Probability} & \multicolumn{5}{c}{Labels (\%)}       \\ \cline{2-6} 
                             & 20    & 40    & 60    & 80    & 100   \\ \hline
0.2                          & 38.98 & 56.52 & 59.33 & 82.09 & 86.76 \\
0.3                          & 37.19 & 51.50 & 58.80 & 78.69 & 88.03 \\
0.4                          & 38.46 & 52.46 & 61.19 & 78.98 & 89.24 \\
0.5                          & 38.63 & 57.14 & 66.38 & 76.56 & 91.63 \\
0.8                          & 38.78 & 48.45 & 64.88 & 81.41 & 92.69 \\ \hline
\end{tabular}
\end{table}
\begin{table}[h]
\centering
\caption{Label efficiency on Fisheye multi-label classification across different probabilities of MPD IB. ResNet50 is encoder.}
\label{tab:MPD_ib_voc_prob_labels}
\begin{tabular}{c|lllll}
\hline
\multirow{2}{*}{Probability} & \multicolumn{5}{c}{Labels (\%)}                                                                                             \\ \cline{2-6} 
                             & \multicolumn{1}{c}{20} & \multicolumn{1}{c}{40} & \multicolumn{1}{c}{60} & \multicolumn{1}{c}{80} & \multicolumn{1}{c}{100} \\ \hline
0.2                          & 32.88                  & 51.61                  & 62.62                  & 81.24                  & 87.57                   \\
0.3                          & 34.01                  & 51.87                  & 61.35                  & 83.01                  & 89.23                   \\
0.4                          & 41.39                  & 48.34                  & 66.68                  & 66.37                  & 86.26                   \\
0.5                          & 38.07                  & 52.62                  & 70.44                  & 76.90                  & 91.87                   \\
0.8                          & 38.22                  & 48.66                  & 58.65                  & 70.23                  & 89.24                   \\ \hline
\end{tabular}
\end{table}
The results substantiate MPD's versatility and adaptability across different learning approaches and datasets, which is pivotal for real-world applications where perspective distortion is a significant challenge.
\subsection{Person Re-Identification}
Our ablation studies further substantiate MPD's efficacy within the CLIP-ReIdent method by demonstrating additional results on the ResNet50x16 backbone, suggested in the original work \cite{habel2022clip}. As shown in Table~\ref{tab:hype6}, incorporating MPD with a ResNet50x16 encoder significantly outperforms the baseline Clip-ReIdent. The configuration with \( c_{real} \) and \( c_{imag} \) components ranging from 0.4 to 0.6 achieves the highest mAP of 91.95\% without re-ranking and 97.50\% with re-ranking.
\begin{table*}[h]
\caption{Ablation on the person re-identification for MPD (Clip-ReIdent) with CNN backbone:  Explores multiple configurations of perceptiveness components with ResNet50x16 encoder. MPD probability set to 0.1, same as before. Min. (component) and Max. (component) defines the range of randomly incorporated perspective distortion. Component selection takes place between c-real and c-imaginary, with equal probability. Direction of each component also chosen with equal probability being positive or negative.}
\centering
\scriptsize
\label{tab:hype_6}
\begin{tabular}{ccccc}
\hline
\textbf{Method} & \textbf{Min. (component)} & \textbf{Max. (component)} & \textbf{\begin{tabular}[c]{@{}c@{}}mAP\\ (w/o re-ranking)\end{tabular}} & \textbf{\begin{tabular}[c]{@{}c@{}}mAP\\ (with re-ranking)\end{tabular}} \\ \hline
 Clip-ReIdent - ResNet50x16    & -         & -    & 88.50          & 94.90          \\ \hline
MPD (Clip-ReIdent) -  ResNet50x16 & 0.2         & 0.6  & 91.51 & 97.40 \\ 
MPD (Clip-ReIdent) -  ResNet50x16 & 0.2         & 0.8  & 90.60 & 96.38 \\ 
MPD (Clip-ReIdent) -  ResNet50x16 & 0.4         & 0.6  & \textbf{91.95} & \textbf{97.50} \\ 
\hline
\end{tabular}
\end{table*}
These results are in line with the transformer-based enhancements reported in the main paper and highlight MPD's broad applicability in addressing perspective distortions across diverse architectures.

\subsection{Object Detection}
\label{subsec:od}

\begin{figure}[ht]
  \centering
  \includegraphics[width=0.8\textwidth]{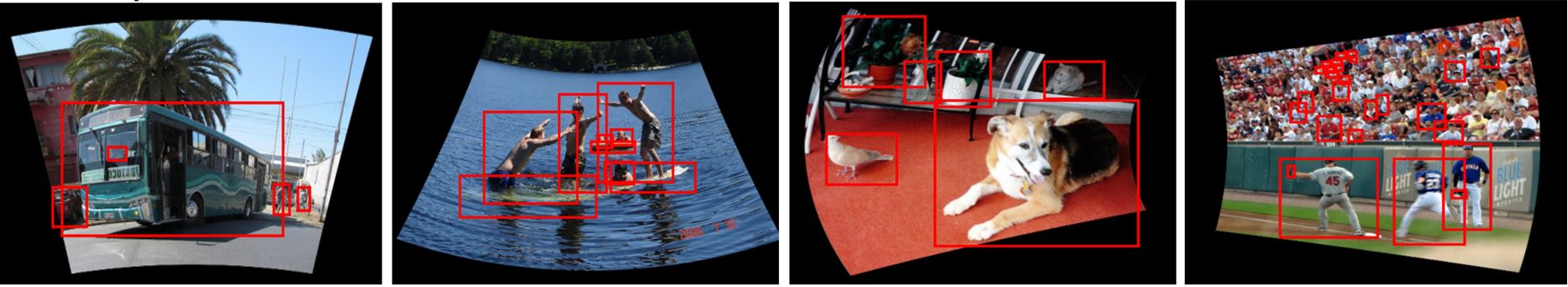}
  \vspace{-3mm}
  \caption{MPD transformed input \& bounding boxes during training}
  \label{fig:coco-pd-detailed}
\end{figure}

Table \ref{tab:coco_od_results_ablation} presents the object detection performance of MPD-OD on the COCO \cite{lin2014microsoft} dataset, with varying probabilities of applying MPD-OD during training. Figure \ref{fig:coco-pd-detailed} shows more MPD-OD transformed image examples. The results show that as the probability of applying MPD-OD increases from 0.1 to 0.5, there is a consistent improvement in detection performance for both FasterRCNN \cite{ren2015faster} and FCOS \cite{9010746} models. Specifically, FasterRCNN gradually increases IoU=0.50:0.95 from 37.60 to 40.00 and IoU=0.50 from 58.10 to 61.10. Similarly, FCOS shows an improvement from 39.10 to 40.20 in IoU=0.50:0.95 and from 58.50 to 60.30 in IoU=0.50. These trends indicate that higher probabilities of applying MPD-OD during training lead to better object detection performance, highlighting the effectiveness of MPD-OD in enhancing model robustness and accuracy. The consistent performance gains across different probabilities demonstrate the adaptability and efficacy of MPD-OD in handling diverse training scenarios and further strengthen the general usefulness of MPD. 

\begin{table}[h]
\centering
\scriptsize
\caption{Object detection performance of MPD-OD on COCO \cite{lin2014microsoft} dataset. Ablations on probability to apply MPD-OD on examples during training.  \textit{Metrics: IoU=0.50:0.95 / IoU=0.50}}
\label{tab:coco_od_results_ablation}
\setlength{\tabcolsep}{2pt}
\begin{tabular}{c|c|ccccc}
\hline
\multirow{2}{*}{Method} & \multirow{2}{*}{Original}       & \multicolumn{5}{c}{MPD-OD}                                                                                                                                                     \\
                        &                                 & P=0.1                           & P=0.2                           & P=0.3                           & P=0.4                           & P=0.5                                    \\ \hline
FasterRCNN\cite{ren2015faster} (NeurIPS'15) & 37.10/55.80                     & 37.60/58.10                     & 39.20/59.40                     & 39.50/60.60                     & 39.70/60.80                     & \textbf{40.00/61.10}                     \\ \hline
FCOS\cite{9010746} (ICCV'19)          & \multicolumn{1}{l|}{38.60/57.40} & \multicolumn{1}{l}{39.10/58.50} & \multicolumn{1}{l}{39.50/59.10} & \multicolumn{1}{l}{39.90/59.70} & \multicolumn{1}{l}{40.00/59.90} & \multicolumn{1}{l}{\textbf{40.20/60.30}} \\ \hline
\end{tabular}
\end{table}
